\documentclass{article}

\usepackage{iclr2027_conference,times}
\iclrfinalcopy

\usepackage[utf8]{inputenc}
\usepackage[T1]{fontenc}
\usepackage{booktabs}
\usepackage{array}
\usepackage{amsmath}
\usepackage{amssymb}
\usepackage{amsthm}
\usepackage{microtype}
\usepackage{xcolor}
\usepackage{colortbl}
\usepackage{graphicx}
\usepackage{multirow}
\usepackage{xspace}
\usepackage{caption}
\usepackage{enumitem}
\usepackage{adjustbox}
\usepackage{placeins}
\usepackage{tabularx}
\usepackage{longtable}
\usepackage[most]{tcolorbox}
\usepackage{wrapfig}
\usepackage{needspace}
\tcbuselibrary{listingsutf8}
\usepackage{hyperref}
\usepackage{url}
\hypersetup{hidelinks}

\theoremstyle{definition}
\newtheorem{definition}{Definition}

\definecolor{TealDark}{HTML}{1a6b6b}
\definecolor{TealMid}{HTML}{2a9090}
\definecolor{TealBg}{HTML}{f0f8f8}
\definecolor{TealRule}{HTML}{aad6d6}

\newcommand{\placeholder}[1]{%
  \tcbox[on line,
    colback=TealBg,colframe=TealMid,
    boxrule=0.4pt,arc=1.5pt,
    left=2pt,right=2pt,top=0.5pt,bottom=0.5pt,
    fontupper=\ttfamily\footnotesize\color{TealDark}
  ]{\{#1\}}%
}

\newcommand{\rolelabel}[1]{%
  \par\smallskip
  {\sffamily\footnotesize\bfseries\color{TealDark}[#1]}\par\nobreak\smallskip
}

\newcommand{\boxsep}{%
  \par\vspace{4pt}%
  {\color{TealRule}\rule{\linewidth}{0.4pt}}%
  \par\vspace{4pt}%
}

\newtcolorbox{promptbox}[1][]{%
  enhanced,
  breakable,
  colback=TealBg,
  colframe=TealDark,
  boxrule=0.8pt,
  arc=3pt,
  left=10pt, right=10pt, top=12pt, bottom=10pt,
  fonttitle=\sffamily\bfseries\small,
  before upper={\raggedright},
  coltitle=white,
  colbacktitle=TealDark,
  attach boxed title to top left={xshift=10pt, yshift=-\tcboxedtitleheight/2},
  boxed title style={colback=TealDark, sharp corners, rounded corners=northwest, boxrule=0pt},
  title={#1},
  drop fuzzy shadow=TealMid,
  before skip=16pt,
  after skip=14pt
}

\newtcblisting{schemabox}{%
  enhanced,
  breakable,
  colback=gray!4!white,
  colframe=gray!50,
  boxrule=0.8pt,
  arc=3pt,
  left=10pt, right=10pt, top=10pt, bottom=10pt,
  drop fuzzy shadow=gray!40,
  listing only,
  listing options={
    basicstyle=\ttfamily\footnotesize,
    breaklines=true,
    columns=fullflexible,
    keepspaces=true,
    showstringspaces=false
  },
  before skip=16pt,
  after skip=14pt
}

\graphicspath{{figures/}}

\DeclareMathOperator*{\argmax}{arg\,max}
\DeclareMathOperator*{\argmin}{arg\,min}
\newcommand{\calD}{\mathcal{D}}
\newcommand{\calS}{\mathcal{S}}
\newcommand{\calX}{\mathcal{X}}
\newcommand{\calY}{\mathcal{Y}}
\newcommand{\calT}{\mathcal{T}}
\newcommand{\calV}{\mathcal{V}}
\newcommand{\calQ}{\mathcal{Q}}

\newcommand{\Joint}{\textsc{Joint}\xspace}
\newcommand{\Decoupled}{\textsc{Decoupled}\xspace}
\newcommand{\Agent}{\textsc{Agent}\xspace}

\title{Joint, Decoupled, or Agent: When Each Multimodal Alignment Paradigm Wins}

\author{
\normalfont
\begin{tabular}{@{}c@{}}
\textbf{
Yi Fang$^{1}$ \quad
Gaurav Srivastava$^{1}$ \quad
Sirui Ding$^{2}$ \quad
Xuan Wang$^{1}$\thanks{Corresponding author: \texttt{xuanw@vt.edu}.}
} \\[0.8em]
\begin{tabular}[t]{@{}c@{}}
$^{1}$Department of Computer Science \\
Virginia Tech \\
Blacksburg, VA, USA \\
\texttt{\{yif,gks,xuanw\}@vt.edu}
\end{tabular}
\qquad
\begin{tabular}[t]{@{}c@{}}
$^{2}$Department of Biomedical Data Science \\
Stanford University \\
Stanford, CA, USA \\
\texttt{srding@stanford.edu}
\end{tabular}
\end{tabular}
}

\begin{document}

\addtocontents{toc}{\protect\setcounter{tocdepth}{-1}}

\maketitle
\lhead{}
\begin{abstract}
Jointly pretrained multimodal encoders are a common way to combine modalities, while alternatives integrate their information during downstream training or inference. When does each approach win? We compare three paradigms, instantiated with public checkpoints, across 16 paired datasets in five domains and multiple training-label budgets. \Joint\ trains a fusion head on frozen, jointly pretrained encoders. \Decoupled\ trains a fusion head on frozen, independently pretrained encoders. \Agent\ gives a frozen LLM task instructions, labelled examples, available sample text, and supervised unimodal readouts. No paradigm dominates the complete $144$ dataset--budget combinations: \Decoupled\ wins $54$, \Joint\ $44$, and \Agent\ $46$ by mean utility. ScienceQA, MELD, and MetMeme shift from \Agent\ to fusion as supervision increases. Between the fusion paradigms, \Joint\ leads on four of five QA datasets across all evaluated budgets, while higher unimodal $5$-NN accuracy is associated with a relative shift toward \Decoupled\ across eight classification datasets (Spearman $\rho=-0.67$). We test whether representation and task diagnostics predict these preferences on held-out datasets before fitting their candidate systems. A two-step cascade combines these diagnostics with task and budget metadata to select \Agent\ or fusion, then \Joint\ or \Decoupled. Under leave-one-group-out evaluation, it selects the winner in $71.4\%$ of cells versus $54.2\%$ for metadata-only routing, with mean relative regret $0.028$ versus $0.057$. Using geometry, supervised representation diagnostics, and text--label matching together yields higher accuracy and lower regret than using any one of these sources with metadata. These results connect paradigm preferences to supervision budget, task format, and unimodal $5$-NN accuracy. Target-task diagnostics thus support selecting among the three paradigms before candidate fitting.
\end{abstract}

\section{Introduction}

Multimodal systems are increasingly assembled from pretrained components. Vision--language, audio--language, biomedical, molecular, and protein--ligand settings all expose encoders or paired representation models that can be reused for downstream tasks \citep{pmlr-v139-radford21a,wu2024largescaleclap,zhang2023biomedclip,liu2023moleculestm,gorantla2025balm}. A central design choice is when and how to integrate the modalities. Given paired inputs and a downstream label, cross-modal information can enter through jointly pretrained representations (\Joint), supervised fusion of independently pretrained encoders (\Decoupled), or inference-time composition by an LLM reading unimodal outputs and available text (\Agent). We call \Joint\ and \Decoupled\ the \emph{fusion paradigms}. \Decoupled\ only learns task-specific fusion on independently pretrained encoders; \Joint\ initializes its frozen encoders from paired pretraining and then learns task-specific fusion. \Agent\ adds LLM inference to supervised readouts.

Choosing among these paradigms requires relating the target task to the available representations and supervision. We study when their preferences change and whether target-task diagnostics can predict those preferences before fitting the candidate systems.

We benchmark the three paradigms on $16$ paired datasets in five domains, varying each dataset's training-label budget. Each domain has a default encoder pair per fusion paradigm. \Agent\ uses the same unimodal encoders as \Decoupled. We match encoder architecture and scale where available, retaining the pretrained representations.

\begin{wrapfigure}{r}{0.42\textwidth}
  \centering
  \includegraphics[width=0.42\textwidth]{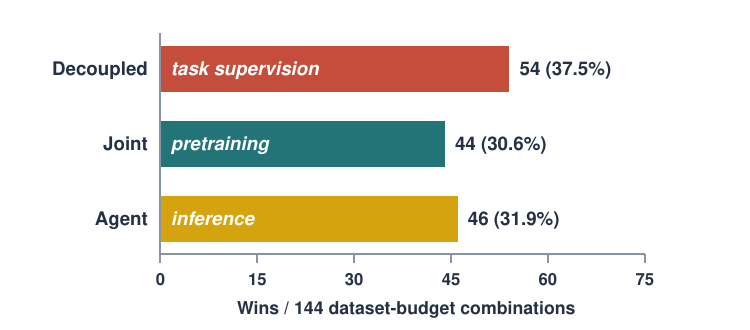}
  \caption{\textbf{Paradigm wins across $144$ dataset--budget combinations.} The system with the highest mean utility wins.}
  \label{fig:win_summary}
\end{wrapfigure}

Three patterns connect paradigm preference to supervision budget, task format, and unimodal $5$-NN accuracy. ScienceQA, MELD, and MetMeme shift from \Agent\ to fusion as supervision increases. AI2D and CUB retain their preferred paradigms across the evaluated budgets. Among fusion systems, candidate-matching QA usually favors jointly pretrained encoders. In classification, higher unimodal $5$-NN accuracy is associated with a relative preference for independently pretrained encoders, suggesting that task-relevant information already present in each modality can support effective downstream fusion (Section~\ref{sec:jd-conditions}).

Component comparisons show that a paradigm win can coexist with little benefit from integration. On AI2D, LLM-only exceeds the full \Agent; on CUB, the visual probe nearly matches \Decoupled. Readouts improve \Agent\ accuracy on ScienceQA and VQA-RAD (Section~\ref{sec:WithinParadigm-variants}).

We predict preferences among the three alignment paradigms from representation geometry, supervised representation diagnostics, text--label matching, and task and budget metadata. A two-step cascade first selects Agent versus fusion, then Joint versus Decoupled. On held-out datasets, the cascade improves selection accuracy and reduces relative regret over metadata-only routing and a task-and-budget lookup. Feature ablations assess their contributions (Section~\ref{sec:manifold-cascade}).

Our contributions are \textbf{(1)} a comparison of three multimodal alignment paradigms under common target-task and budget protocols, identifying preferences associated with supervision budget, task format, and unimodal $5$-NN accuracy, \textbf{(2)} component controls and implementation variants that assess the gains from integration and the stability of paradigm preferences, and \textbf{(3)} held-out-dataset evidence that representation and task diagnostics improve paradigm selection before candidate adaptation, supported by metadata baselines and diagnostic-group ablations.

\section{Method}

We compare three pretrained alignment paradigms on paired target observations, define their utility and selection protocol, and \mbox{predict the preferred paradigm with a two-step cascade.}

\subsection{Three Alignment Paradigms}
\label{sec:three-alignments}

Let $\calD_d=\{(x_i^{(a)},x_i^{(b)},y_i)\}_{i=1}^{n_d}$ be a paired dataset. At budget $K$, seed $s\in\{0,\ldots,4\}$ determines the labelled training set $\calS_{d,K}^{(s)}$ and head initialization. All paradigms share the full test split $\calQ_d$ and a disjoint development split $\calV_d$, fixed across budgets, heads, and seeds. Budget $K$ excludes development labels: fitting and selection use $|\calS_{d,K}^{(s)}|+|\calV_d|$ labels. For metric $m_d$ with direction $s_d\in\{+1,-1\}$, define

\begin{equation}
    U_{\mathcal R}(f) = s_d\,m_d\!\left((f(x),y)_{(x,y)\in\mathcal R}\right),
    \qquad U(p;d,K)=\frac{1}{5}\sum_{s=0}^{4}U_{\calQ_d}\!\left(f_{p,d,K}^{(s)}\right),
    \label{eq:utility}
\end{equation}
where $m_d$ is accuracy or MAE. Utilities are comparable only within a cell and exclude deployment cost. For the candidate set $\mathcal P=\{\Joint,\Decoupled,\Agent\}$, let $U^\star(d,K)=\max_{p\in\mathcal P}U(p;d,K)$. Routing accuracy and regret use this candidate set. LLM-only and unimodal probes assess component value (Section~\ref{sec:WithinParadigm-variants}).

\paragraph{Fusion Heads}
\Joint\ and \Decoupled\ both train a fusion head on a frozen encoder pair $(G_a, G_b)$. The head consists of a fusion operator $\psi_\omega$, drawn from a search space $\Psi$ of eight standard architectures (Appendix~\ref{app:fusion-operators}), followed by a task head $h_\theta$. For each $\psi\in\Psi$, parameters are fit on the labelled set (suppressing seed indices here),
\begin{equation}
    (\omega_\psi^\star,\theta_\psi^\star) \;=\; \argmin_{\omega,\theta}\sum_{(x,y)\in\calS_{d,K}} \ell_d\!\bigl(h_\theta\!\bigl(\psi_\omega(G_a(x^{(a)}),G_b(x^{(b)}))\bigr),\,y\bigr).
    \label{eq:fm-head}
\end{equation}
For classification and regression, select hyperparameters and checkpoints on $\calV_d$ within each head $\psi$ (Table~\ref{tab:training_hparams}). QA uses a fixed optimization configuration with development-based checkpoint selection (Appendix~\ref{app:qa-scoring}). For each seed, then select the head:
\begin{equation}
    \hat\psi_{p,d,K}^{(s)}=\argmax_{\psi\in\Psi}U_{\calV_d}\!\left(f_{p,\psi,d,K}^{(s)}\right),
    \qquad f_{p,d,K}^{(s)}=f_{p,\hat\psi_{p,d,K}^{(s)},d,K}^{(s)}.
    \label{eq:head-selection}
\end{equation}
The five selected models determine the mean utilities and routing targets. For QA, Eq.~\eqref{eq:fm-head} uses a shared scalar scorer for each image--candidate or molecule--candidate pair, with cross-entropy over candidate scores. For image QA, each candidate text concatenates the question and answer.

\paragraph{\Joint: Alignment at Pretraining}
\Joint\ sets $(G_a,G_b)=(F_J^a,F_J^b)$ in Eq.~\eqref{eq:fm-head}, \mbox{using towers jointly pretrained on paired multimodal data.}

\paragraph{\Decoupled: Supervised Fusion of Unaligned Encoders}
\Decoupled\ uses independently pretrained unimodal encoders $E_a:\calX_a\to\mathbb{R}^{r_a}$ and $E_b:\calX_b\to\mathbb{R}^{r_b}$, never aligned to each other, and sets $(G_a,G_b)=(E_a,E_b)$ in Eq.~\eqref{eq:fm-head}. Matching controls architecture family and scale (Definition~\ref{def:matching}), not pretraining histories. Supervised \mbox{fusion need not create a shared embedding space.}

\paragraph{\Agent: LLM-Based Inference-Time Composition}
\Agent\ combines a frozen instruction-tuned LLM with readouts $g_a,g_b$ using the same frozen $E_a,E_b$ as \Decoupled. Trainable readouts are fitted separately per seed on $\calS_{d,K}$ and selected on $\calV_d$. For image MCQ, the image readout is a trained bilinear candidate scorer, while the text readout uses unscaled text--option cosine similarity without a fitted head (Appendix~\ref{app:tool-schema}). Outputs are per-candidate scores for classification/QA and scalar predictions for regression. The LLM receives task instruction $\mathcal I_d$, up to ten labelled in-context examples $\mathcal E_{d,K}\subset\calS_{d,K}$, readout outputs $o_a,o_b$, and raw sample text $x^{(t)}$ when available. Non-text inputs reach it only through readouts. For KIBA and PDBbind, $x^{(t)}$ contains ligand SMILES. The LLM returns a task answer, parsed into a class, candidate index, or regression value:
\begin{equation}
    f_{\Agent,d,K}(x)=\mathrm{parse}\bigl(\pi_{\mathrm{LLM}}(\mathcal I_d,\mathcal E_{d,K},x^{(t)},o_a,o_b)\bigr).
\end{equation}
Inline-1 supplies both readouts in one greedy-decoding turn. Utility averages the same five labelled-set seeds as \Decoupled\ (Appendix~\ref{app:agent-prompt}). Readouts and demonstrations depend on the labelled set, making Agent performance budget-dependent even with a frozen LLM.

\paragraph{Prerequisites and Tie Band}
All paradigms use paired target inputs. \Decoupled\ needs paired labels for fusion and \Joint\ also needs a paired-pretrained checkpoint. \Agent\ fits its trainable readouts separately on these paired target observations. Our tie convention $\Agent\prec\Decoupled\prec\Joint$ prioritizes fewer cross-modal training requirements. For tolerance $\epsilon\ge0$, let

\begin{equation}
    \calT_\epsilon(d,K) \;=\; \bigl\{\,p : U^\star(d,K) - U(p;d,K) \le \epsilon_d\,\bigr\}, \qquad \epsilon_d = \epsilon\,|U^\star(d,K)|,
    \label{eq:tie-set}
\end{equation}
be the paradigms within tolerance. The earliest in the preference order is the target. $\epsilon=0$ selects a maximizer. The tolerance $\epsilon$ specifies an application-defined utility margin. Main-table raw counts instead use maximum utility, assigning exact fusion ties to \Decoupled.

\subsection{Feature Extraction}
\label{sec:geometry-summary}
The features combine representation geometry, supervised representation diagnostics (two ANOVA and three $5$-NN features), a text diagnostic, and task and budget metadata (Table~\ref{tab:feature-pool-summary}). Geometry describes the structure within each representation and the correspondence between modalities. ANOVA and $5$-NN assess how target labels organize these representations and how well local neighbours recover them, providing task-specific evidence about the inputs available to supervised fusion.

These features use frozen $E_a,E_b$ embeddings and labels from $\calS_{d,K}$, except mutual-kNN, which uses up to $8{,}000$ paired observations from the full training pool. Per-side geometry enters as minimum and maximum values. CKA, mutual-kNN, and concatenated kNN require paired observations. Feature definitions and inputs appear in Appendix~\ref{app:features} and Table~\ref{tab:feature-availability}.

The composite text diagnostic combines label-name matching and supervised text-neighbour prediction. For classification \mbox{with $C$ classes,}
\begin{equation}
\Delta_{\mathrm{zs}}(\calS)=\operatorname{norm}(a_{\mathrm{TL}})-\operatorname{norm}(a_{\mathrm{kNN}}^{\mathrm{text}}),\qquad
\operatorname{norm}(a)=\frac{a-1/C}{1-1/C}.
\end{equation}
Here $a_{\mathrm{TL}}$ is cosine-nearest-class-name accuracy using a fixed BGE-M3 text probe. $a_{\mathrm{kNN}}^{\mathrm{text}}$ is leave-one-out $5$-NN accuracy in the default \Decoupled\ text embedding. The label-name term uses up to $20$ sampled examples from $\calS_{d,K}$; the neighbour term uses the labelled set capped at $8{,}000$ rows, so the two accuracies can be evaluated on different subsets. Regression instead uses $-r$, the Pearson correlation of text-side $5$-NN predictions. QA uses zero because options vary by question. This diagnostic receives no size residualization. For classification, its design asks whether label semantics support prediction beyond local text-neighbour evidence. This motivates its use in the \Agent/Fusion decision, where the LLM interprets sample text and label names alongside readouts. Its predictive contribution is evaluated through the feature ablations. Appendix~\ref{app:zs-feature} gives implementation details.

\subsection{Two-Step Cascade}
\label{sec:hier-selection}

The hierarchy follows the paradigms\textquotesingle\ shared components: \Joint\ and \Decoupled\ use supervised fusion of frozen embeddings, while \Agent\ adds LLM composition of text and readouts. It first separates these integration procedures, then distinguishes the fusion encoders\textquotesingle\ pretraining source. Step~1 asks whether \Agent\ is competitive. Otherwise Step~2 chooses between \Joint\ and \Decoupled. Each step selects diagnostics and budget features alongside mandatory task indicators.

Step~1 learns \Agent/Fusion labels on all cells under Eq.~\eqref{eq:tie-set}. Step~2 learns \Joint/\Decoupled\ labels on Fusion cells (Appendix~\ref{app:step-labels}).

\paragraph{Feature Processing and Selection}
Supervised representation diagnostics are residualized against $\log_{10}|\calS_{d,K}|$, separately for regression and other tasks. Step~1 always includes regression and QA indicators; Step~2 always includes the QA indicator. The text diagnostic $\Delta_{\mathrm{zs}}$ is mandatory in Step~1. Each step selects a budget block containing all descriptors, all except budget fraction, or none. The symbolic-sequence indicator is a ranked candidate in both steps.

Residualization and Laplacian ranking are fitted once per outer fold using diagnostics from all outer-training datasets, including inner-validation groups, without paradigm-winner labels. Inner selection of feature width and budget block uses these fixed transformations; imputation, missingness indicators, standardization, and classifiers are refitted within each training partition. The outer-held-out dataset enters only at prediction time. Appendix~\ref{app:cascade-config} specifies the full procedure.

\paragraph{Classifier and Decision}
Each step uses L2-regularized logistic regression ($C=1$), with decision score $s_t(x)=w_t^\top\phi_t(x)+b_t$. Grouped cross-validation on the outer training set gives out-of-fold scores $o_{t,i}$. For the step\textquotesingle s training positive-label fraction $\pi_t$, the threshold is
\begin{equation}
\tau_t=Q_{1-\pi_t}\!\left(\{o_{t,i}\}\right),\qquad t\in\{1,2\}.
\end{equation}
Step~2 computes its prior and score quantile on Fusion-labelled cells. The quantile threshold approximately matches the positive rate on the pooled out-of-fold scores to the training prior, with deviations from finite samples and tied scores. The cascade predicts
\begin{equation}
\hat p(d,K;\epsilon)=
\begin{cases}
\Agent,&s_1\ge\tau_1,\\
\Joint,&s_1<\tau_1\ \text{and}\ s_2\ge\tau_2,\\
\Decoupled,&\text{otherwise.}
\end{cases}
\label{eq:framework}
\end{equation}
Final classifiers fit the outer training groups and use these independently determined thresholds.

\paragraph{Evaluation Methodology}
We evaluate $144$ dataset--budget combinations from $16$ datasets in $15$ groups: DrugBank-Desc and DrugBank-Pharma are always held out together, in both outer evaluation and inner selection. Each held-out cell reads one labelled-set feature seed per run. Results average five runs, while training cells use five-seed feature means. Targets use the mean candidate utilities and are re-derived at each $\epsilon$. We report cell-weighted winner accuracy and relative regret $(U^\star-U(\hat p))/|U^\star|$. The paired-bootstrap protocol uses $4{,}000$ bootstrap draws of the $15$ groups, keeping all budgets and seeds of \mbox{a selected group together (Appendix~\ref{app:eval-protocol}).}

\section{Experiments}
\label{sec:experiments}

We examine budget-dependent preferences, input contributions, and prediction of paradigm choice. Appendix~\ref{app:additional-results} expands these analyses.

\paragraph{Experimental Setup}
We evaluate $16$ paired datasets across five domains and three formats (classification, regression, QA). We vary training budget $K$ under the protocol in Section~\ref{sec:three-alignments}. Shots are per class or regression quantile bucket, so $20$-shot can exceed $10\%$ in label count (Table~\ref{tab:label-counts}). A cell is one dataset--budget combination. Main tables report $48$ cells at three budgets per dataset. Routing uses $144$ dataset--budget combinations over the full grids. Encoder pairs are in Appendix~\ref{app:datasets}. The default \Agent\ uses Qwen3-32B~\citep{yang2025qwen3,qwen2025qwen332bmodelcard}.

\subsection{Paradigm Preferences Depend on Task and Supervision}

\begin{figure}[t]
  \centering
  \includegraphics[width=0.94\textwidth]{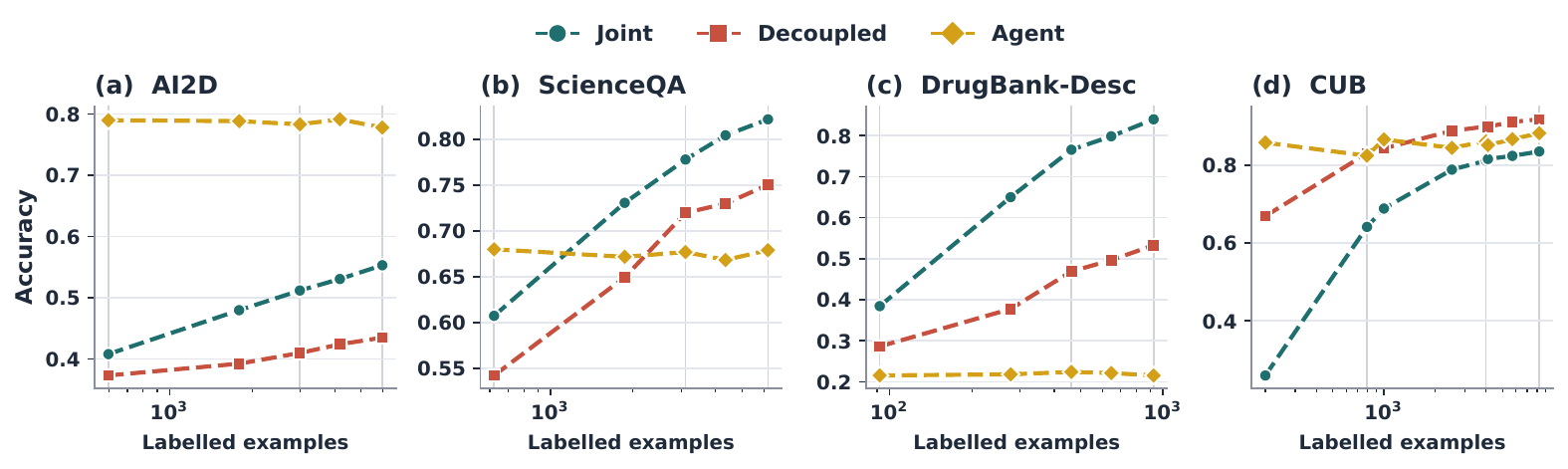}
  \caption{\textbf{Persistent and changing paradigm preferences across supervision budgets.} Utility versus labelled examples (log scale). Lines mark main-table budgets. More datasets: Figure~\ref{fig:budget_curves}.}
  \label{fig:budget_main}
\end{figure}

\paragraph{Paradigm preferences can persist or reverse as supervision increases.} Tables~\ref{tab:agent_fm_cls_reg} and~\ref{tab:agent_fm_qa} report accuracy and MAE for $48$ dataset--budget cells. Figure~\ref{fig:win_summary} summarizes wins over all $144$ cells in the full budget grids. Across full budget grids, AI2D favors \Agent, CUB favors \Decoupled, and DrugBank-Desc favors \Joint. ScienceQA instead shifts from \Agent\ to \Joint, while CrisisMMD returns to \Agent\ at full budget (Figure~\ref{fig:budget_main}). MELD and MetMeme also move from \Agent\ at $20$-shot to fusion with larger training sets. Budget effects can be non-monotonic.

\paragraph{Most paradigm preferences are consistent across seeds.} On the fixed test splits, the same paradigm wins strictly across all five training seeds in about $79\%$ of the main-table cells (Table~\ref{tab:uncertainty}).

\paragraph{Performance margins vary substantially relative to inference cost.}
Estimated \Agent\ inference FLOPs are approximately $360$--$3{,}000$ times those of fusion. For classification and QA, its mean advantage over the better fusion system ranges from marginal to substantial. Appendix~\ref{app:cost} reports the gain distribution, full-grid relative margins, and routing costs.

\begin{table}[t]
\centering
\caption{\textbf{Paradigm comparison on CLS \& REG datasets.} Accuracy ($\uparrow$) for CLS, MAE ($\downarrow$) for REG (IEMOCAP, KIBA, PDBbind). \Joint\ and \Decoupled: $5$-seed means of the seed-specific development-selected models. \Agent: means over the same $5$ seeds as \Decoupled. $20$-shot $=$ $20$ training labels per class, or per quantile bucket for regression (counts in Table~\ref{tab:label-counts}). \textbf{Bold} = winner per budget, ties going to \Decoupled. Seed variability: Appendix~\ref{app:perseed}. For the eight control datasets, full-budget \Agent\ shares the full test split and \mbox{five-seed aggregation of Table~\ref{tab:agent-controls}.}}
\label{tab:agent_fm_cls_reg}
\footnotesize
\setlength{\tabcolsep}{4pt}
\begin{adjustbox}{max width=\textwidth}
\begin{tabular}{l ccc ccc ccc}
\toprule
\textbf{Dataset}
  & \multicolumn{3}{c}{\textbf{20-shot}}
  & \multicolumn{3}{c}{\textbf{10\%}}
  & \multicolumn{3}{c}{\textbf{100\%}} \\
\cmidrule(lr){2-4}\cmidrule(lr){5-7}\cmidrule(lr){8-10}
  & \textbf{Joint} & \textbf{Decoupled} & \textbf{Agent}
  & \textbf{Joint} & \textbf{Decoupled} & \textbf{Agent}
  & \textbf{Joint} & \textbf{Decoupled} & \textbf{Agent} \\
\midrule
\rowcolor{gray!10}\multicolumn{10}{c}{\textbf{\textit{Classification (accuracy $\uparrow$)}}} \\
\midrule
CUB & 0.811 & \textbf{0.899} & 0.861 & 0.641 & \textbf{0.830} & 0.824 & 0.833 & \textbf{0.917} & 0.882 \\
CrisisMMD & 0.458 & 0.306 & \textbf{0.460} & \textbf{0.634} & 0.476 & 0.627 & 0.668 & 0.538 & \textbf{0.683} \\
Food101 & 0.767 & 0.695 & \textbf{0.872} & 0.804 & 0.733 & \textbf{0.880} & 0.862 & 0.808 & \textbf{0.906} \\
Fracture & \textbf{0.294} & 0.247 & 0.095 & \textbf{0.245} & 0.199 & 0.131 & 0.388 & \textbf{0.405} & 0.136 \\
MELD & 0.375 & 0.445 & \textbf{0.591} & 0.535 & \textbf{0.564} & 0.523 & 0.571 & \textbf{0.637} & 0.516 \\
Memotion & \textbf{0.420} & 0.349 & 0.166 & 0.454 & \textbf{0.454} & 0.354 & 0.447 & \textbf{0.451} & 0.219 \\
MetMeme & 0.252 & 0.208 & \textbf{0.290} & \textbf{0.319} & 0.266 & 0.296 & \textbf{0.409} & 0.323 & 0.307 \\
OxfordPet & 0.918 & \textbf{0.949} & 0.927 & 0.836 & \textbf{0.928} & 0.841 & 0.945 & \textbf{0.962} & 0.945 \\
\midrule
\rowcolor{gray!10}\multicolumn{10}{c}{\textbf{\textit{Regression (MAE $\downarrow$)}}} \\
\midrule
IEMOCAP & 0.725 & \textbf{0.680} & 0.730 & 0.648 & \textbf{0.606} & 0.717 & 0.556 & \textbf{0.517} & 0.584 \\
KIBA & 0.589 & \textbf{0.581} & 1.535 & 0.517 & \textbf{0.434} & 0.655 & 0.420 & \textbf{0.344} & 0.817 \\
PDBbind & \textbf{1.230} & 1.261 & 1.531 & 1.193 & \textbf{1.170} & 1.477 & \textbf{1.160} & 1.165 & 1.360 \\
\bottomrule
\end{tabular}
\end{adjustbox}
\end{table}

\begin{table}[t]
\centering
\caption{\textbf{Paradigm comparison on QA datasets} (accuracy $\uparrow$). Protocol and markers as \mbox{in Table~\ref{tab:agent_fm_cls_reg}.}}
\label{tab:agent_fm_qa}
\footnotesize
\setlength{\tabcolsep}{4pt}
\begin{adjustbox}{max width=\textwidth}
\begin{tabular}{l ccc ccc ccc}
\toprule
\textbf{Dataset}
  & \multicolumn{3}{c}{\textbf{10\%}}
  & \multicolumn{3}{c}{\textbf{50\%}}
  & \multicolumn{3}{c}{\textbf{100\%}} \\
\cmidrule(lr){2-4}\cmidrule(lr){5-7}\cmidrule(lr){8-10}
  & \textbf{Joint} & \textbf{Decoupled} & \textbf{Agent}
  & \textbf{Joint} & \textbf{Decoupled} & \textbf{Agent}
  & \textbf{Joint} & \textbf{Decoupled} & \textbf{Agent} \\
\midrule
AI2D & 0.408 & 0.373 & \textbf{0.790} & 0.512 & 0.410 & \textbf{0.783} & 0.551 & 0.432 & \textbf{0.778} \\
DrugBank-Desc & \textbf{0.384} & 0.286 & 0.216 & \textbf{0.766} & 0.469 & 0.224 & \textbf{0.840} & 0.533 & 0.216 \\
DrugBank-Pharma & \textbf{0.426} & 0.311 & 0.277 & \textbf{0.671} & 0.487 & 0.267 & \textbf{0.772} & 0.595 & 0.277 \\
ScienceQA & 0.607 & 0.542 & \textbf{0.680} & \textbf{0.778} & 0.720 & 0.677 & \textbf{0.819} & 0.750 & 0.679 \\
VQA-RAD & 0.295 & 0.295 & \textbf{0.325} & 0.289 & 0.314 & \textbf{0.341} & 0.300 & 0.318 & \textbf{0.375} \\
\bottomrule
\end{tabular}
\end{adjustbox}
\end{table}

\subsection{Input Interventions and Implementation Sensitivity}
\label{sec:WithinParadigm-variants}

Matched controls and implementation variants test component value and preference stability.

\subsubsection{Text and Readouts Have Task-Dependent Value}
\label{sec:agent-variants}

\begin{figure}[t]
\centering
\includegraphics[width=\textwidth]{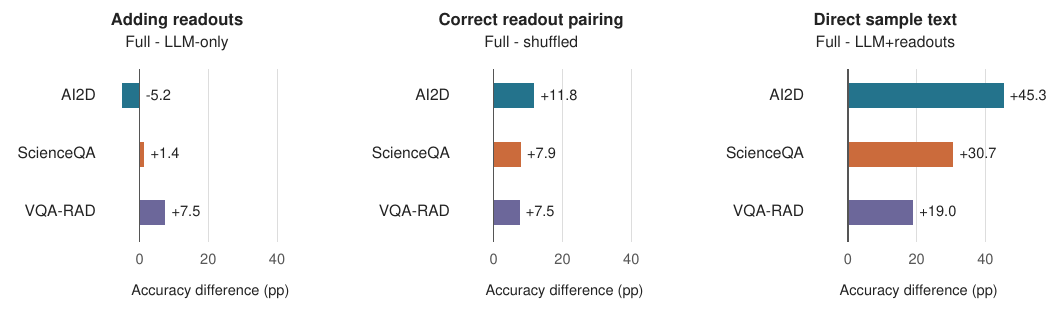}
\caption{\textbf{QA input controls.} Full-system minus control accuracy, in percentage points. Each contrast compares the full system with a separate input control on matched full-budget test examples. Table~\ref{tab:agent-controls} lists all controls and the input retained in each condition.}
\label{fig:qa-control-contrasts}
\end{figure}

\paragraph{Readout information does not always improve QA accuracy.}
Controls use the main-table full-budget samples and five-seed aggregation (Table~\ref{tab:agent-controls}). LLM-only exceeds all three paradigms only on AI2D ($0.830$ versus $0.778$ for \Agent). Readouts improve ScienceQA and VQA-RAD means by $1.4$ and $7.5$ points. Yet full \Agent\ exceeds shuffled readouts by $11.8$, $7.9$, and $7.5$ points on these three tasks (Figure~\ref{fig:qa-control-contrasts}). On AI2D, omitting readouts beats both informative and shuffled scores.

Food101 readout gains are $2.9/2.8/0.8$ points at $20$-shot/$10\%$/$100\%$, over LLM-only's $0.843/0.852/0.898$ (Table~\ref{tab:agent-score-avg}). \Agent\ wins throughout, with diminishing readout gains.

\paragraph{Readouts remain useful even without informative label names.}
CUB and OxfordPet retain $0.600$ and $0.632$ under anonymised labels, above LLM-only and shuffled controls.

\paragraph{Raw question and option text contribute to QA performance.} Deleting only the question body yields $0.602/0.545/0.205$ on AI2D/ScienceQA/VQA-RAD. Removing question and option contents lowers these to $0.325/0.372/0.185$.

\paragraph{Component controls show where integration adds value.}
Score averaging without the LLM trails \Agent\ by $9.6$--$24.9$ points on the reported Agent-winning classification settings (Appendix~\ref{app:agent-controls}). CUB\textquotesingle s visual probe nearly matches \Decoupled\ at full budget ($0.916/0.917$; Table~\ref{tab:text_tower_ablation}).

\subsubsection{Preferences Also Depend on the Candidate Implementations}

\paragraph{Fusion-head choice strongly affects MCQ accuracy.}
At full budget, the diagnostic best--worst head spread is $0.11$--$0.58$ on MCQ versus $0.006$--$0.13$ on classification (Figure~\ref{fig:fusion_map}). Restricting \Joint\ to concat reverses the PDBbind winner. These spreads show why head selection matters when comparing the fusion paradigms.

\paragraph{The value of domain specialization depends on the encoder configuration.} On Fracture, biomedical \Joint\ checkpoints outperform general CLIP by up to $9.7$ points, whereas neither Rad-DINO nor Phikon improves \Decoupled\ over DINOv2 (Figure~\ref{fig:domain_pretraining}). These variants change architecture as well as pretraining (Appendix~\ref{app:domain-ablation}).

\paragraph{Providing both readouts usually outperforms single-tool selection.}
Across the seven classification datasets with both readouts available to the Multi-2 controller, at three budgets, supplying both readouts in one turn (Inline-1) usually beats requesting one in multiple turns (Multi-2). On CUB, Multi-2 selects the text readout on $79\%$ of examples, and mean accuracy across budgets falls from about $0.86$ to $0.16$ (Figure~\ref{fig:agent_dispatch_main}).

\paragraph{Paradigm preferences are largely stable across capable LLMs, but a tiny controller performs poorly.}
Gemma-3-27B-it leads on CUB and OxfordPet across reported budgets, and Qwen3-32B on Food101. Across tasks and budgets evaluated with all four LLMs, the \Agent--fusion preference usually remains unchanged (Figure~\ref{fig:agent_llm_main}). SmolLM2-135M never outperforms the better fusion system in its evaluated classification and QA settings and has a mean parse-failure rate near $90\%$, indicating difficulty following the answer protocol (Appendix~\ref{app:agent-controls}).

\paragraph{In-context examples do not consistently improve performance.}
In-context examples yield less than one accuracy point on average but help MetMeme by $9.9$ points at $20$-shot and hurt OxfordPet by $5.7$ points at $10\%$. Across all eight classification datasets at three budgets, removing examples usually preserves the preferred paradigm, even when accuracy changes substantially (Figure~\ref{fig:agent_kshot_main}).

\subsection{Diagnostics of Paradigm Preference and Grouped Prediction}
\label{sec:manifold-cascade}
\begin{figure}[t]
\centering
\includegraphics[width=0.92\textwidth]{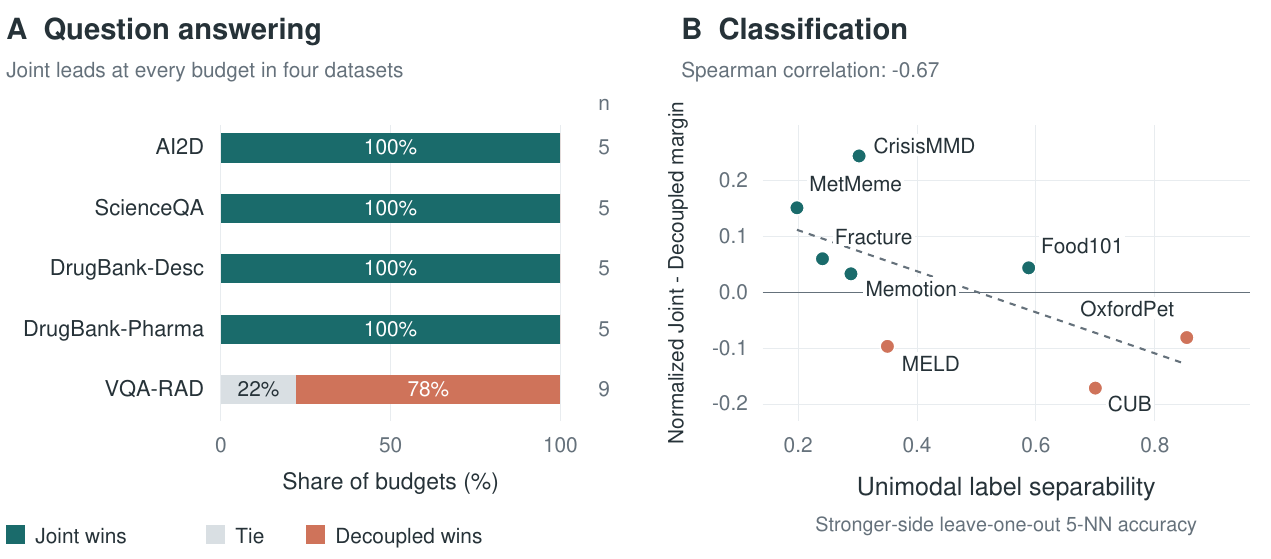}
\caption{\textbf{Diagnostics and fusion preference.} \textbf{A:} QA win shares over $n$ budgets. \textbf{B:} Unimodal $5$-NN accuracy versus normalized Joint--Decoupled margin across classification datasets. Dashed line: descriptive fit. Definitions and sensitivity checks: Appendix~\ref{app:jd-descriptive}.}
\label{fig:jd-conditions}
\end{figure}

\paragraph{Task format and unimodal $5$-NN accuracy relate to fusion preference.}
\label{sec:jd-conditions}
Figure~\ref{fig:jd-conditions} connects two diagnostics to paradigm preference. Across all budgets, \Joint\ exceeds \Decoupled\ on four of five QA datasets, whose fusion heads perform candidate matching. VQA-RAD favors \Decoupled\ or ties. Across eight classification datasets, higher unimodal $5$-NN accuracy is associated with a smaller \Joint-minus-\Decoupled\ margin (Spearman $\rho=-0.67$). Each point represents one dataset, averaging over matched budget cells. The horizontal axis uses the stronger modality\textquotesingle s five-seed mean $5$-NN accuracy; the vertical axis averages the Joint--Decoupled utility difference normalized by the best of all three paradigms. The association remains negative under dataset deletion and chance normalization (Appendix~\ref{app:jd-descriptive}).

Table~\ref{tab:router-comparison} evaluates prediction of the preferred member of $\{\Joint,\Decoupled,\Agent\}$ on $144$ dataset--budget combinations in $15$ held-out groups. Metadata-only routing measures the added value of diagnostics; Direct and the other classifiers test alternative predictors.

\begin{table}[!htbp]
\centering\footnotesize\setlength{\tabcolsep}{4pt}
\caption{\textbf{Winner prediction and relative regret.} Five-seed means over $144$ dataset--budget combinations with $15$ outer groups. The cascade uses prior-matched thresholds. Bold marks the highest accuracy and lowest regret at each tolerance.}
\label{tab:router-comparison}
\begin{adjustbox}{max width=\textwidth}\begin{tabular}{@{}c rr rr rr rr@{}}\toprule
& \multicolumn{2}{c}{Cascade} & \multicolumn{2}{c}{Direct three-way} & \multicolumn{2}{c}{TB-Majority} & \multicolumn{2}{c}{Metadata-only} \\
$\epsilon$ & Acc.$\uparrow$ & Regret$\downarrow$ & Acc.$\uparrow$ & Regret$\downarrow$ & Acc.$\uparrow$ & Regret$\downarrow$ & Acc.$\uparrow$ & Regret$\downarrow$ \\\midrule
0\% & \textbf{0.714} & \textbf{0.028} & 0.450 & 0.091 & 0.486 & 0.127 & 0.542 & 0.057 \\
0.2\% & \textbf{0.697} & \textbf{0.036} & 0.432 & 0.109 & 0.500 & 0.127 & 0.542 & 0.067 \\
0.5\% & \textbf{0.708} & \textbf{0.038} & 0.442 & 0.104 & 0.507 & 0.127 & 0.569 & 0.066 \\
1\% & \textbf{0.623} & \textbf{0.043} & 0.599 & 0.051 & 0.507 & 0.127 & 0.569 & 0.068 \\
2\% & \textbf{0.646} & 0.078 & 0.568 & 0.087 & 0.514 & 0.127 & 0.583 & \textbf{0.064} \\
\bottomrule\end{tabular}\end{adjustbox}\end{table}

\Needspace{5\baselineskip}
\paragraph{Diagnostics add predictive information beyond metadata.} Under the same cascade structure and tuning protocol, diagnostics improve accuracy from $0.542$ to $0.714$ and reduce regret from $0.057$ to $0.028$. Paired $95\%$ intervals for these differences are $[0.001,0.346]$ and $[-0.054,-0.002]$. The gains persist when any one evaluation group is omitted from the fixed predictions (Appendix~\ref{app:metadata-group-sensitivity}).

\paragraph{Diagnostic groups contribute jointly to routing.}
Adding any single diagnostic group to metadata yields accuracy of $0.535$--$0.549$, versus $0.714$ for Full (Table~\ref{tab:routing-feature-ablation}). Removing supervised representation diagnostics or the text diagnostic reduces accuracy to $0.544$ or $0.529$ and raises regret to $0.108$ or $0.105$. Removing geometry has a smaller effect, with accuracy $0.671$ and regret $0.041$. Adding geometry alone slightly increases accuracy over metadata-only ($0.549$ versus $0.542$) but raises regret ($0.113$ versus $0.057$), illustrating that accuracy gains can coexist with more costly errors. All three removals increase mean regret, and the reported paired regret-difference intervals favour Full. These results average five runs with separately sampled text-probe examples.

\paragraph{The cascade improves both stage decisions over Direct.}
At $\epsilon=0$, the cascade reaches $0.806$ Step~1 accuracy and $0.843$ Step~2 accuracy, with the latter evaluated on settings whose true winner is \Joint\ or \Decoupled. Direct reports $0.654$ and $0.627$ for the corresponding binary decisions on the same evaluation subsets.

\paragraph{Routing accuracy is stable under the tested feature-encoder change.}
Replacing the general-domain feature encoders with the base-tier configuration preserves routing accuracy: final accuracy changes from $71.4\%$ to $71.8\%$, with Step~1 at $0.804$ and Step~2 at $0.867$ (Table~\ref{tab:routing-stage-accuracy}).

\paragraph{Prediction performance varies across implementations.}
Direct uses the same feature information, preprocessing, grouped validation, and feature-selection framework as the cascade, but uses a single-stage three-class objective and argmax decisions (Appendix~\ref{app:direct-config}). At $\epsilon=0$, accuracy/regret is $0.450/0.091$ for Direct and $0.486/0.127$ for TB-Majority. Among six further classifiers, logistic regression ($C=0.1$) performs best at $0.571/0.086$ (Table~\ref{tab:routing-additional-classifiers}, Appendix~\ref{app:additional-routing-classifiers}). Paired intervals for the cascade comparisons with Direct and TB-Majority appear in Tables~\ref{tab:router-utility} and~\ref{tab:cascade-prediction}.

\paragraph{Higher selection accuracy does not always mean lower regret.} CUB has $27.5\%$ selection accuracy but regret of only $0.0284$. MELD reaches $64.0\%$ accuracy yet incurs $0.1189$ regret, above metadata-only\textquotesingle s $0.0414$ (Table~\ref{tab:router-per-dataset}). CUB errors overselect \Agent\ at larger budgets, whereas costly MELD errors select \Decoupled\ in k-shot settings. In the inspected CUB, KIBA, and MELD folds, feature selection omits explicit budget descriptors from both cascade stages. Their diagnostics still vary with the labelled subset, but this indirect budget signal coexists with errors in tracking budget-dependent preferences. This observation motivates testing mandatory budget features in future routing variants (Appendix~\ref{app:per-dataset-routing}).

\section{Related Work}

\textbf{Joint Pretraining, Supervised Fusion, and Post-hoc Alignment.} CLIP-style joint pretraining \citep{pmlr-v139-radford21a,Zhai_2023_ICCV,Cherti_2023_CVPR} and related biomedical, audio, and molecular models \citep{zhang2023biomedclip,wu2024largescaleclap,liu2023moleculestm}, together with paired protein--ligand adaptation \citep{gorantla2025balm} motivate \Joint. Supervised fusion of frozen representations \citep{baltrusaitis2018multimodal} motivates \Decoupled, while MultiBench and MULTIBENCH++ provide cross-domain fusion benchmarks \citep{NEURIPS_DATASETS_AND_BENCHMARKS2021_37693cfc,xue2025multibenchpp}. We compare preferences across three alignment paradigms and test their predictability before target-task adaptation. Our fusion heads include a query-based bridge inspired by frozen-encoder methods \citep{li2023blip2}.

Post-hoc alignment explicitly builds cross-modal compatibility through lightweight functions \citep{rosenfeld2022ape}, projectors, fusion modules, or token-level adaptation \citep{maniparambil2025frozen,zhang2025assessing,groger2025structure,kulasekara2025saffe,tupakula2025thin,mahmud2024maavt}, or classification-head prototypes \citep{mendez2026prototypes}.

\textbf{LLM Composition and Multimodal LLMs.} Tool-using LMs compose model outputs at inference \citep{yao2023react,schick2023toolformer,shen2023hugginggpt,yang2023mmreact,gao2023assistgpt}. Multimodal LLMs learn modality--language connections through pretraining or instruction tuning \citep{liu2023llava,bai2023qwenvl,bai2025qwen3vl,openai2023gpt4vision,geminiteam2025gemini,han2024onellm}, and MMMU evaluates deployed multimodal systems \citep{Yue_2024_CVPR}. Our \Agent\ composes unimodal readouts and available text through a frozen text LLM at inference time.

\textbf{Selection Before Target-Task Adaptation.} Algorithm selection uses instance features \citep{rice1976algorithm}. Transferability scores rank checkpoints from target representations and labels \citep{nguyen2020leep,you2021logme}. Multimodal information analysis studies redundancy, uniqueness, and synergy \citep{liang2023quantifying}. At a new target task, we predict the paradigm preference before fitting its candidate heads, using \mbox{a router trained on outcomes from other datasets.}

\textbf{Representation Geometry and the Modality Gap.} Diagnostics include manifold structure \citep{tenenbaum2000isomap,roweis2000lle}, uniformity \citep{wang2020understanding}, anisotropy \citep{ethayarajh2019contextual}, neural collapse \citep{papyan2020neuralcollapse}, and intrinsic dimension \citep{pope2021intrinsic}. Modality-gap studies examine shared-space separation, training interventions, information imbalance, and semantic compression \citep{liang2022mind,eslami2025mitigate,schrodi2025two,grassucci2025compression,grassucci2026medical}. Spectral similarity characterizes eigenvalue structure, while basis alignment concerns cross-modal correspondence \citep{sarkar2026spectral}. Our router uses paired relational features across embedding widths, per-side spectral diagnostics, and task diagnostics for paradigm selection.

\section{Conclusion}
Multimodal alignment selection involves two distinct questions: which paradigm performs best, and how much its integration contributes beyond the individual components. Our results show why both matter: a winning paradigm can derive much of its performance from one component, while changes in supervision can alter the preferred paradigm. The held-out routing results make representation and task diagnostics useful for narrowing this choice before candidate fitting. Selection errors also differ in utility cost, so winner accuracy and relative regret together characterize the value of that recommendation. This distinction motivates a utility-aware routing objective that weights mistakes by their observed performance gaps. Such an objective would prioritize avoiding costly errors while giving less weight to choices among nearly tied paradigms, a direction for evaluation on held-out datasets. Extending the candidate set to unimodal and LLM-only systems would connect component evaluation directly to routing. Further work can examine incomplete modality correspondences and multimodal instruction tuning \citep{liu2023llava,han2024onellm}, and use controlled pretraining comparisons to isolate the contribution of alignment.

\clearpage
\section*{AI usage statement}
Large language models, including Claude and Codex, were used to assist with writing, literature lookup, and consistency checks across the manuscript, tables, and figures. The LLMs evaluated as \Agent\ controllers are objects of study, distinct from these writing assistants. The authors take responsibility for the experimental evidence, \mbox{references, and interpretations in the final manuscript.}

\section*{Reproducibility statement}
All datasets and pretrained checkpoints are public (Appendix~\ref{app:datasets}). Upon publication, we will release code and records to regenerate the results (Appendix~\ref{app:reproducibility}). Appendices~\ref{app:training-details}, \ref{app:agent-prompt}, and~\ref{app:decision-rule} give training details, the \Agent\ prompt, and feature definitions.

\section*{Ethics statement}
This work benchmarks existing public datasets and pretrained models, and collects no new data from human subjects. Some datasets contain social-media content that may be offensive (CrisisMMD, Memotion, MetMeme), and two contain medical images (Fracture, VQA-RAD). We use every dataset only for the benchmark task its release \mbox{defines, under its \mbox{stated licence and access conditions.}}

\clearpage
\bibliographystyle{iclr2027_conference}
\bibliography{refs}

\begin{thebibliography}{72}
\providecommand{\natexlab}[1]{#1}
\providecommand{\url}[1]{\texttt{#1}}
\expandafter\ifx\csname urlstyle\endcsname\relax
  \providecommand{\doi}[1]{doi: #1}\else
  \providecommand{\doi}{doi: \begingroup \urlstyle{rm}\Url}\fi

\bibitem[Ahmad et~al.(2022)Ahmad, Simon, Chithrananda, Grand, and Ramsundar]{ahmad2022chemberta2}
Walid Ahmad, Elana Simon, Seyone Chithrananda, Gabriel Grand, and Bharath Ramsundar.
\newblock {ChemBERTa-2}: Towards chemical foundation models.
\newblock \emph{arXiv \mbox{preprint arXiv:2209.01712}},~2022.

\bibitem[Arevalo et~al.(2017)Arevalo, Solorio, Montes-y G{\'o}mez, and Gonz{\'a}lez]{arevalo2017gated}
John Arevalo, Thamar Solorio, Manuel Montes-y G{\'o}mez, and Fabio~A Gonz{\'a}lez.
\newblock Gated multimodal units for information fusion.
\newblock \emph{arXiv \mbox{preprint arXiv:1702.01992}},~2017.

\bibitem[Baevski et~al.(2020)Baevski, Zhou, Mohamed, and Auli]{baevski2020wav2vec2}
Alexei Baevski, Yuhao Zhou, Abdelrahman Mohamed, and Michael Auli.
\newblock wav2vec 2.0: A framework for self-supervised learning of speech representations.
\newblock In \emph{Advances in Neural Information \mbox{Processing Systems}},~volume~33,~pp.\  \mbox{12449--12460}. \mbox{Curran Associates, Inc.},~2020.

\bibitem[Bai et~al.(2023)Bai, Bai, Yang, Wang, Tan, Wang, Lin, Zhou, and Zhou]{bai2023qwenvl}
Jinze Bai, Shuai Bai, Shusheng Yang, Shijie Wang, Sinan Tan, Peng Wang, Junyang Lin, Chang Zhou, and Jingren Zhou.
\newblock {Qwen-VL}: A versatile vision-language model for understanding, localization, text reading, and beyond,~2023.
\newblock URL \url{https://arxiv.org/abs/2308.12966}.

\bibitem[Bai et~al.(2025)Bai, Cai, Chen, Chen, Chen, Cheng, Deng, Ding, Gao, Ge, Ge, Guo, Huang, Huang, Huang, Hui, Jiang, Li, Li, Li, Li, Lin, Lin, Liu, Liu, Liu, Liu, Liu, Liu, Lu, Luo, Lv, Men, Meng, Ren, Ren, Song, Sun, Tang, Tu, Wan, Wang, Wang, Wang, Wang, Xie, Xu, Xu, Xu, Yang, Yang, Yang, Yang, Yu, Zhang, Zhang, Zhang, Zheng, Zhong, Zhou, Zhou, Zhou, Zhu, and Zhu]{bai2025qwen3vl}
Shuai Bai, Yuxuan Cai, Ruizhe Chen, Keqin Chen, Xionghui Chen, Zesen Cheng, Lianghao Deng, Wei Ding, Chang Gao, Chunjiang Ge, Wenbin Ge, Zhifang Guo, Qidong Huang, Jie Huang, Fei Huang, Binyuan Hui, Shutong Jiang, Zhaohai Li, Mingsheng Li, Mei Li, Kaixin Li, Zicheng Lin, Junyang Lin, Xuejing Liu, Jiawei Liu, Chenglong Liu, Yang Liu, Dayiheng Liu, Shixuan Liu, Dunjie Lu, Ruilin Luo, Chenxu Lv, Rui Men, Lingchen Meng, Xuancheng Ren, Xingzhang Ren, Sibo Song, Yuchong Sun, Jun Tang, Jianhong Tu, Jianqiang Wan, Peng Wang, Pengfei Wang, Qiuyue Wang, Yuxuan Wang, Tianbao Xie, Yiheng Xu, Haiyang Xu, Jin Xu, Zhibo Yang, Mingkun Yang, Jianxin Yang, An~Yang, Bowen Yu, Fei Zhang, Hang Zhang, Xi~Zhang, Bo~Zheng, Humen Zhong, Jingren Zhou, Fan Zhou, Jing Zhou, Yuanzhi Zhu, and Ke~Zhu.
\newblock {Qwen3-VL} technical report,~2025.
\newblock URL \url{https://arxiv.org/abs/2511.21631}.

\bibitem[Baltru{\v{s}}aitis et~al.(2019)Baltru{\v{s}}aitis, Ahuja, and Morency]{baltrusaitis2018multimodal}
Tadas Baltru{\v{s}}aitis, Chaitanya Ahuja, and Louis-Philippe Morency.
\newblock Multimodal machine learning: A survey and taxonomy.
\newblock \emph{IEEE transactions on pattern analysis and \mbox{machine intelligence}},~41\penalty0 (2):\mbox{423--443},~2019.

\bibitem[Beltagy et~al.(2019)Beltagy, Lo, and Cohan]{beltagy2019scibert}
Iz~Beltagy, Kyle Lo, and Arman Cohan.
\newblock {SciBERT}: A pretrained language model for scientific text.
\newblock In \emph{Proceedings of the 2019 Conference on Empirical Methods in Natural Language Processing and the 9th International Joint Conference on Natural Language \mbox{Processing (EMNLP-IJCNLP)}},~pp.\  \mbox{3615--3620}. \mbox{Association for Computational Linguistics},~2019.
\newblock \mbox{\doi{10.18653/v1/D19-1371}}.

\bibitem[Chen et~al.(2022)Chen, Du, Zhu, Ma, Berg-Kirkpatrick, and Dubnov]{chen2022htsat}
Ke~Chen, Xingjian Du, Bilei Zhu, Zejun Ma, Taylor Berg-Kirkpatrick, and Shlomo Dubnov.
\newblock {HTS-AT}: A hierarchical token-semantic audio transformer for sound classification and detection.
\newblock In \emph{ICASSP 2022 - 2022 IEEE International Conference on Acoustics, Speech and Signal \mbox{Processing (ICASSP)}},~pp.\  \mbox{646--650}. IEEE,~2022.
\newblock \mbox{\doi{10.1109/ICASSP43922.2022.9746312}}.

\bibitem[Cherti et~al.(2023)Cherti, Beaumont, Wightman, Wortsman, Ilharco, Gordon, Schuhmann, Schmidt, and Jitsev]{Cherti_2023_CVPR}
Mehdi Cherti, Romain Beaumont, Ross Wightman, Mitchell Wortsman, Gabriel Ilharco, Cade Gordon, Christoph Schuhmann, Ludwig Schmidt, and Jenia Jitsev.
\newblock Reproducible scaling laws for contrastive language-image learning.
\newblock In \emph{Proceedings of the IEEE/CVF conference on computer vision and \mbox{pattern recognition}},~pp.\  \mbox{2818--2829},~2023.

\bibitem[Eslami \& de~Melo(2025)Eslami and de~Melo]{eslami2025mitigate}
Sedigheh Eslami and Gerard de~Melo.
\newblock Mitigate the gap: Improving cross-modal alignment in {CLIP}.
\newblock In \emph{International Conference on \mbox{Learning Representations}},~2025.
\newblock URL \url{https://proceedings.iclr.cc/paper_files/paper/2025/file/cc1de06a58ba1db43538a37e076e466d-Paper-Conference.pdf}.

\bibitem[Eslami et~al.(2023)Eslami, Meinel, and de~Melo]{eslami2023pubmedclip}
Sedigheh Eslami, Christoph Meinel, and Gerard de~Melo.
\newblock {PubMedCLIP}: How much does {CLIP} benefit visual question answering in the medical domain?
\newblock In \emph{Findings of the Association for Computational Linguistics: \mbox{EACL 2023}},~pp.\  \mbox{1181--1193}. \mbox{Association for Computational Linguistics},~2023.
\newblock \mbox{\doi{10.18653/v1/2023.findings-eacl.88}}.

\bibitem[Ethayarajh(2019)]{ethayarajh2019contextual}
Kawin Ethayarajh.
\newblock How contextual are contextualized word representations? comparing the geometry of {BERT}, {ELMo}, and {GPT-2} embeddings.
\newblock In \emph{Proceedings of the 2019 Conference on Empirical Methods in Natural Language Processing and the 9th International Joint Conference on Natural Language \mbox{Processing (EMNLP-IJCNLP)}},~pp.\  \mbox{55--65}. \mbox{Association for Computational Linguistics},~2019.

\bibitem[Filiot et~al.(2023)Filiot, Ghermi, Olivier, Jacob, Fidon, Mac~Kain, Saillard, and Schiratti]{filiot2023phikon}
Alexandre Filiot, Ridouane Ghermi, Antoine Olivier, Paul Jacob, Lucas Fidon, Alice Mac~Kain, Charlie Saillard, and Jean-Baptiste Schiratti.
\newblock Scaling self-supervised learning for histopathology with masked image modeling.
\newblock \emph{medRxiv},~2023.
\newblock \mbox{\doi{10.1101/2023.07.21.23292757}}.

\bibitem[Gao et~al.(2023)Gao, Ji, Zhou, Lin, Chen, Fan, and Shou]{gao2023assistgpt}
Difei Gao, Lei Ji, Luowei Zhou, Kevin~Qinghong Lin, Joya Chen, Zihan Fan, and Mike~Zheng Shou.
\newblock {AssistGPT}: A general multi-modal assistant that can plan, execute, inspect, and learn.
\newblock \emph{arXiv \mbox{preprint arXiv:2306.08640}},~2023.

\bibitem[{Gemini Team, Google}(2023)]{geminiteam2025gemini}
{Gemini Team, Google}.
\newblock Gemini: a family of highly capable multimodal models.
\newblock \emph{arXiv \mbox{preprint arXiv:2312.11805}},~2023.

\bibitem[{Gemma Team, Google DeepMind}(2025)]{gemmateam2025gemma3}
{Gemma Team, Google DeepMind}.
\newblock Gemma 3 technical report.
\newblock \emph{arXiv \mbox{preprint arXiv:2503.19786}},~2025.

\bibitem[Gong et~al.(2021)Gong, Chung, and Glass]{gong2021ast}
Yuan Gong, Yu-An Chung, and James Glass.
\newblock {AST}: Audio spectrogram transformer.
\newblock In \emph{Proc. \mbox{Interspeech 2021}},~pp.\  \mbox{571--575},~2021.
\newblock \mbox{\doi{10.21437/Interspeech.2021-698}}.

\bibitem[Gorantla et~al.(2025)Gorantla, Gema, Yang, Serrano-Morr{\'a}s, Suutari, Ju{\'a}rez-Jim{\'e}nez, and Mey]{gorantla2025balm}
Rohan Gorantla, Aryo~Pradipta Gema, Ian~Xi Yang, {\'A}lvaro Serrano-Morr{\'a}s, Benjamin Suutari, Jordi Ju{\'a}rez-Jim{\'e}nez, and Antonia S. J.~S. Mey.
\newblock Learning binding affinities via fine-tuning of protein and ligand language models.
\newblock \emph{Journal of Chemical Information \mbox{and Modeling}},~65\penalty0 (22):\mbox{12279--12291},~2025.
\newblock \mbox{\doi{10.1021/acs.jcim.5c02063}}.

\bibitem[Grassucci et~al.(2025{\natexlab{a}})Grassucci, Cicchetti, and Comminiello]{grassucci2026medical}
Eleonora Grassucci, Giordano Cicchetti, and Danilo Comminiello.
\newblock Closing the gap in multimodal medical representation alignment.
\newblock In \emph{2025 IEEE 35th International Workshop on Machine Learning for Signal \mbox{Processing (MLSP)}},~pp.\  \mbox{1--6}. IEEE,~2025{\natexlab{a}}.
\newblock \mbox{\doi{10.1109/MLSP62443.2025.11204242}}.

\bibitem[Grassucci et~al.(2025{\natexlab{b}})Grassucci, Cicchetti, Uncini, and Comminiello]{grassucci2025compression}
Eleonora Grassucci, Giordano Cicchetti, Aurelio Uncini, and Danilo Comminiello.
\newblock Semantic compression via multimodal representation learning.
\newblock \emph{arXiv \mbox{preprint arXiv:2509.24431}},~2025{\natexlab{b}}.
\newblock URL \url{https://arxiv.org/abs/2509.24431}.

\bibitem[Gr{\"o}ger et~al.(2025)Gr{\"o}ger, Wen, Le, and Brbi{\'c}]{groger2025structure}
Fabian Gr{\"o}ger, Shuo Wen, Huyen Le, and Maria Brbi{\'c}.
\newblock With limited data for multimodal alignment, let the {STRUCTURE} guide you.
\newblock In \emph{Advances in Neural Information Processing \mbox{Systems (NeurIPS)}},~2025.
\newblock arXiv:2506.16895.

\bibitem[Gu et~al.(2021)Gu, Tinn, Cheng, Lucas, Usuyama, Liu, Naumann, Gao, and Poon]{gu2021pubmedbert}
Yu~Gu, Robert Tinn, Hao Cheng, Michael Lucas, Naoto Usuyama, Xiaodong Liu, Tristan Naumann, Jianfeng Gao, and Hoifung Poon.
\newblock Domain-specific language model pretraining for biomedical natural language processing.
\newblock \emph{ACM Transactions on Computing \mbox{for Healthcare}},~3\penalty0 (1):\mbox{1--23},~2021.
\newblock \mbox{\doi{10.1145/3458754}}.

\bibitem[Han et~al.(2024)Han, Gong, Zhang, Wang, Zhang, Lin, Qiao, Gao, and Yue]{han2024onellm}
Jiaming Han, Kaixiong Gong, Yiyuan Zhang, Jiaqi Wang, Kaipeng Zhang, Dahua Lin, Yu~Qiao, Peng Gao, and Xiangyu Yue.
\newblock {OneLLM}: One framework to align all modalities with language.
\newblock In \emph{Proceedings of the IEEE/CVF Conference on Computer Vision and \mbox{Pattern Recognition}},~pp.\  \mbox{26584--26595},~2024.
\newblock URL \url{https://arxiv.org/abs/2312.03700}.

\bibitem[He et~al.(2022)He, Chen, Xie, Li, Doll{\'a}r, and Girshick]{He_2022_CVPR}
Kaiming He, Xinlei Chen, Saining Xie, Yanghao Li, Piotr Doll{\'a}r, and Ross Girshick.
\newblock Masked autoencoders are scalable vision learners.
\newblock In \emph{Proceedings of the IEEE/CVF conference on computer vision and \mbox{pattern recognition}},~pp.\  \mbox{16000--16009},~2022.

\bibitem[He et~al.(2005)He, Cai, and Niyogi]{he2005laplacian}
Xiaofei He, Deng Cai, and Partha Niyogi.
\newblock Laplacian score for feature selection.
\newblock In \emph{Advances in Neural Information \mbox{Processing Systems}},~volume~18,~2005.

\bibitem[Huang et~al.(2023)Huang, Bianchi, Yuksekgonul, Montine, and Zou]{huang2023plip}
Zhi Huang, Federico Bianchi, Mert Yuksekgonul, Thomas~J. Montine, and James Zou.
\newblock A visual--language foundation model for pathology image analysis using medical twitter.
\newblock \emph{\mbox{Nature Medicine}},~29\penalty0 (9):\mbox{2307--2316},~2023.
\newblock \mbox{\doi{10.1038/s41591-023-02504-3}}.

\bibitem[Kim et~al.(2017)Kim, On, Lim, Kim, Ha, and Zhang]{kim2017hadamard}
Jin-Hwa Kim, Kyoung-Woon On, Woosang Lim, Jeonghee Kim, Jung-Woo Ha, and Byoung-Tak Zhang.
\newblock Hadamard product for low-rank bilinear pooling.
\newblock In \emph{International Conference on Learning \mbox{Representations (ICLR)}},~2017.

\bibitem[Kornblith et~al.(2019)Kornblith, Norouzi, Lee, and Hinton]{kornblith2019similarity}
Simon Kornblith, Mohammad Norouzi, Honglak Lee, and Geoffrey Hinton.
\newblock Similarity of neural network representations revisited.
\newblock In \emph{International Conference on \mbox{Machine Learning}},~volume~97,~pp.\  \mbox{3519--3529}. PMLR,~2019.

\bibitem[Kulasekara et~al.(2025)Kulasekara, Ingl{\'e}s-Romero, Imbern{\'o}n, and Abell{\'a}n]{kulasekara2025saffe}
Maithri Kulasekara, Juan~F. Ingl{\'e}s-Romero, Baldomero Imbern{\'o}n, and Jos{\'e}~L. Abell{\'a}n.
\newblock Saffe: Multimodal model composition with semantic-alignment fusion of frozen encoders.
\newblock \emph{The Journal \mbox{of Supercomputing}},~81\penalty0 (10):\mbox{1114},~2025.
\newblock \mbox{\doi{10.1007/s11227-025-07473-7}}.

\bibitem[Li et~al.(2023)Li, Li, Savarese, and Hoi]{li2023blip2}
Junnan Li, Dongxu Li, Silvio Savarese, and Steven Hoi.
\newblock {BLIP-2}: Bootstrapping language-image pre-training with frozen image encoders and large language models.
\newblock In \emph{International Conference on \mbox{Machine Learning}},~pp.\  \mbox{19730--19742},~2023.

\bibitem[Liang et~al.(2021)Liang, Lyu, Fan, Wu, Cheng, Wu, Chen, Wu, Lee, Zhu, Salakhutdinov, and Morency]{NEURIPS_DATASETS_AND_BENCHMARKS2021_37693cfc}
Paul~Pu Liang, Yiwei Lyu, Xiang Fan, Zetian Wu, Yun Cheng, Jason Wu, Leslie Chen, Peter Wu, Michelle~A. Lee, Yuke Zhu, Ruslan Salakhutdinov, and Louis-Philippe Morency.
\newblock {MultiBench}: Multiscale benchmarks for multimodal representation learning.
\newblock In \emph{Advances in Neural Information Processing Systems Datasets and Benchmarks Track \mbox{(Round 1)}},~2021.

\bibitem[Liang et~al.(2023)Liang, Cheng, Fan, Ling, Nie, Chen, Deng, Allen, Auerbach, Mahmood, Salakhutdinov, and Morency]{liang2023quantifying}
Paul~Pu Liang, Yun Cheng, Xiang Fan, Chun~Kai Ling, Suzanne Nie, Richard Chen, Zihao Deng, Nicholas Allen, Randy Auerbach, Faisal Mahmood, Ruslan Salakhutdinov, and Louis-Philippe Morency.
\newblock Quantifying \& modeling multimodal interactions: An information decomposition framework.
\newblock In \emph{Advances in Neural Information Processing \mbox{Systems (NeurIPS)}},~volume~36,~2023.

\bibitem[Liang et~al.(2022)Liang, Zhang, Kwon, Yeung, and Zou]{liang2022mind}
Victor~Weixin Liang, Yuhui Zhang, Yongchan Kwon, Serena Yeung, and James~Y Zou.
\newblock Mind the gap: Understanding the modality gap in multi-modal contrastive representation learning.
\newblock \emph{Advances in Neural Information \mbox{Processing Systems}},~35:\mbox{17612--17625},~2022.

\bibitem[Lin et~al.(2023)Lin, Akin, Rao, Hie, Zhu, Lu, Smetanin, Verkuil, Kabeli, Shmueli, dos Santos~Costa, Fazel-Zarandi, Sercu, Candido, and Rives]{lin2023esm2}
Zeming Lin, Halil Akin, Roshan Rao, Brian Hie, Zhongkai Zhu, Wenting Lu, Nikita Smetanin, Robert Verkuil, Ori Kabeli, Yaniv Shmueli, Allan dos Santos~Costa, Maryam Fazel-Zarandi, Tom Sercu, Salvatore Candido, and Alexander Rives.
\newblock Evolutionary-scale prediction of atomic-level protein structure with a language model.
\newblock \emph{Science},~379\penalty0 (6637):\mbox{1123--1130},~2023.
\newblock \mbox{\doi{10.1126/science.ade2574}}.

\bibitem[Liu et~al.(2023{\natexlab{a}})Liu, Li, Wu, and Lee]{liu2023llava}
Haotian Liu, Chunyuan Li, Qingyang Wu, and Yong~Jae Lee.
\newblock Visual instruction tuning.
\newblock \emph{Advances in neural information \mbox{processing systems}},~36:\mbox{34892--34916},~2023{\natexlab{a}}.

\bibitem[Liu et~al.(2022)Liu, Wang, Liu, Lasenby, Guo, and Tang]{liu2022graphmvp}
Shengchao Liu, Hanchen Wang, Weiyang Liu, Joan Lasenby, Hongyu Guo, and Jian Tang.
\newblock Pre-training molecular graph representation with {3D} geometry.
\newblock In \emph{International Conference on Learning \mbox{Representations (ICLR)}},~2022.

\bibitem[Liu et~al.(2023{\natexlab{b}})Liu, Nie, Wang, Lu, Qiao, Liu, Tang, Xiao, and Anandkumar]{liu2023moleculestm}
Shengchao Liu, Weili Nie, Chengpeng Wang, Jiarui Lu, Zhuoran Qiao, Ling Liu, Jian Tang, Chaowei Xiao, and Anima Anandkumar.
\newblock Multi-modal molecule structure--text model for text-based retrieval and editing.
\newblock \emph{Nature \mbox{Machine Intelligence}},~5\penalty0 (12):\mbox{1447--1457},~2023{\natexlab{b}}.

\bibitem[Mahmud et~al.(2024)Mahmud, Mo, Tian, and Marculescu]{mahmud2024maavt}
Tanvir Mahmud, Shentong Mo, Yapeng Tian, and Diana Marculescu.
\newblock {MA-AVT}: Modality alignment for parameter-efficient audio-visual transformers.
\newblock In \emph{Proceedings of the IEEE/CVF Conference on Computer Vision and Pattern Recognition \mbox{Workshops (CVPRW)}},~2024.
\newblock arXiv:2406.04930.

\bibitem[Maniparambil et~al.(2025)Maniparambil, Akshulakov, Djilali, Narayan, Singh, and O'Connor]{maniparambil2025frozen}
Mayug Maniparambil, Raiymbek Akshulakov, Yasser Abdelaziz~Dahou Djilali, Sanath Narayan, Ankit Singh, and Noel~E. O'Connor.
\newblock Harnessing frozen unimodal encoders for flexible multimodal alignment.
\newblock In \emph{Proceedings of the IEEE/CVF Conference on Computer Vision and Pattern \mbox{Recognition (CVPR)}},~pp.\  \mbox{29847--29857},~2025.
\newblock arXiv:2409.19425.

\bibitem[M{\'e}ndez et~al.(2026)M{\'e}ndez, Confalonieri, and D{\'i}az~Rodr{\'i}guez]{mendez2026prototypes}
David M{\'e}ndez, Roberto Confalonieri, and Natalia D{\'i}az~Rodr{\'i}guez.
\newblock Supervised classification heads as semantic prototypes: Unlocking vision-language alignment via weight recycling.
\newblock \emph{arXiv \mbox{preprint arXiv:2605.22484}},~2026.
\newblock URL \url{https://arxiv.org/abs/2605.22484}.

\bibitem[Nguyen et~al.(2020)Nguyen, Hassner, Seeger, and Archambeau]{nguyen2020leep}
Cuong~V. Nguyen, Tal Hassner, Matthias~W. Seeger, and C{\'e}dric Archambeau.
\newblock {LEEP}: A new measure to evaluate transferability of learned representations.
\newblock In \emph{International Conference on Machine \mbox{Learning (ICML)}},~volume 119 of \emph{PMLR},~pp.\  \mbox{7294--7305},~2020.

\bibitem[{OpenAI}(2023)]{openai2023gpt4vision}
{OpenAI}.
\newblock {GPT-4V(ision)} system card.
\newblock \url{https://cdn.openai.com/papers/GPTV_System_Card.pdf},~2023.

\bibitem[{OpenAI}(2025)]{openai2025gptoss}
{OpenAI}.
\newblock gpt-oss-120b \& gpt-oss-20b model card.
\newblock \emph{arXiv \mbox{preprint arXiv:2508.10925}},~2025.

\bibitem[Oquab et~al.(2024)Oquab, Darcet, Moutakanni, Vo, Szafraniec, Khalidov, Fernandez, Haziza, Massa, El-Nouby, Assran, Ballas, Galuba, Howes, Huang, Li, Misra, Rabbat, Sharma, Synnaeve, Xu, J{\'e}gou, Mairal, Labatut, Joulin, and Bojanowski]{oquab2024dinov2}
Maxime Oquab, Timoth{\'e}e Darcet, Th{\'e}o Moutakanni, Huy~V. Vo, Marc Szafraniec, Vasil Khalidov, Pierre Fernandez, Daniel Haziza, Francisco Massa, Alaaeldin El-Nouby, Mido Assran, Nicolas Ballas, Wojciech Galuba, Russell Howes, Po-Yao Huang, Shang-Wen Li, Ishan Misra, Michael Rabbat, Vasu Sharma, Gabriel Synnaeve, Hu~Xu, Herv{\'e} J{\'e}gou, Julien Mairal, Patrick Labatut, Armand Joulin, and Piotr Bojanowski.
\newblock {DINOv2}: Learning robust visual features without supervision.
\newblock \emph{Transactions on Machine Learning \mbox{Research (TMLR)}},~2024.

\bibitem[Papyan et~al.(2020)Papyan, Han, and Donoho]{papyan2020neuralcollapse}
Vardan Papyan, XY~Han, and David~L Donoho.
\newblock Prevalence of neural collapse during the terminal phase of deep learning training.
\newblock \emph{Proceedings of the National Academy \mbox{of Sciences}},~117\penalty0 (40):\mbox{24652--24663},~2020.

\bibitem[Perez et~al.(2018)Perez, Strub, de~Vries, Dumoulin, and Courville]{Perez_Strub_deVries_Dumoulin_Courville_2018}
Ethan Perez, Florian Strub, Harm de~Vries, Vincent Dumoulin, and Aaron Courville.
\newblock {FiLM}: Visual reasoning with a general conditioning layer.
\newblock In \emph{Proceedings of the AAAI conference on \mbox{artificial intelligence}},~volume~32,~2018.

\bibitem[P{\'e}rez-Garc{\'i}a et~al.(2025)P{\'e}rez-Garc{\'i}a, Sharma, Bond-Taylor, Bouzid, Salvatelli, Ilse, Bannur, Castro, Schwaighofer, Lungren, Wetscherek, Codella, Hyland, Alvarez-Valle, and Oktay]{perezgarcia2025raddino}
Fernando P{\'e}rez-Garc{\'i}a, Harshita Sharma, Sam Bond-Taylor, Kenza Bouzid, Valentina Salvatelli, Maximilian Ilse, Shruthi Bannur, Daniel~C. Castro, Anton Schwaighofer, Matthew~P. Lungren, Maria~Teodora Wetscherek, Noel Codella, Stephanie~L. Hyland, Javier Alvarez-Valle, and Ozan Oktay.
\newblock Exploring scalable medical image encoders beyond text supervision.
\newblock \emph{Nature \mbox{Machine Intelligence}},~7\penalty0 (1):\mbox{119--130},~2025.
\newblock \mbox{\doi{10.1038/s42256-024-00965-w}}.

\bibitem[Pope et~al.(2021)Pope, Zhu, Abdelkader, Goldblum, and Goldstein]{pope2021intrinsic}
Phillip Pope, Chen Zhu, Ahmed Abdelkader, Micah Goldblum, and Tom Goldstein.
\newblock The intrinsic dimension of images and its impact on learning.
\newblock In \emph{International Conference on Learning \mbox{Representations (ICLR)}},~2021.

\bibitem[{Qwen Team}(2025)]{qwen2025qwen332bmodelcard}
{Qwen Team}.
\newblock {Qwen3-32B}.
\newblock Hugging Face model card, \url{https://huggingface.co/Qwen/Qwen3-32B},~2025.

\bibitem[Radford et~al.(2021)Radford, Kim, Hallacy, Ramesh, Goh, Agarwal, Sastry, Askell, Mishkin, Clark, Krueger, and Sutskever]{pmlr-v139-radford21a}
Alec Radford, Jong~Wook Kim, Chris Hallacy, Aditya Ramesh, Gabriel Goh, Sandhini Agarwal, Girish Sastry, Amanda Askell, Pamela Mishkin, Jack Clark, Gretchen Krueger, and Ilya Sutskever.
\newblock Learning transferable visual models from natural language supervision.
\newblock In \emph{International conference on \mbox{machine learning}},~pp.\  \mbox{8748--8763}. PMLR,~2021.

\bibitem[Reed et~al.(2016)Reed, Akata, Lee, and Schiele]{reed2016learning}
Scott Reed, Zeynep Akata, Honglak Lee, and Bernt Schiele.
\newblock Learning deep representations of fine-grained visual descriptions.
\newblock In \emph{Proceedings of the IEEE Conference on Computer Vision and Pattern Recognition},~2016.

\bibitem[Rice(1976)]{rice1976algorithm}
John~R. Rice.
\newblock The algorithm selection problem.
\newblock \emph{Advances \mbox{in Computers}},~15:\mbox{65--118},~1976.

\bibitem[Rosenfeld et~al.(2022)Rosenfeld, Nakkiran, Pouransari, Tuzel, and Faghri]{rosenfeld2022ape}
Elan Rosenfeld, Preetum Nakkiran, Hadi Pouransari, Oncel Tuzel, and Fartash Faghri.
\newblock {APE}: Aligning pretrained encoders to quickly learn aligned multimodal representations.
\newblock \emph{arXiv \mbox{preprint arXiv:2210.03927}},~2022.
\newblock URL \url{https://arxiv.org/abs/2210.03927}.

\bibitem[Roweis \& Saul(2000)Roweis and Saul]{roweis2000lle}
Sam~T. Roweis and Lawrence~K. Saul.
\newblock Nonlinear dimensionality reduction by locally linear embedding.
\newblock \emph{Science},~290\penalty0 (5500):\mbox{2323--2326},~2000.
\newblock \mbox{\doi{10.1126/science.290.5500.2323}}.

\bibitem[Sarkar(2026)]{sarkar2026spectral}
Krisanu Sarkar.
\newblock On the spectral geometry of cross-modal representations: A functional map diagnostic for multimodal alignment.
\newblock \emph{arXiv \mbox{preprint arXiv:2604.08579}},~2026.
\newblock URL \url{https://arxiv.org/abs/2604.08579}.

\bibitem[Schick et~al.(2023)Schick, Dwivedi-Yu, Dess{\`\i}, Raileanu, Lomeli, Hambro, Zettlemoyer, Cancedda, and Scialom]{schick2023toolformer}
Timo Schick, Jane Dwivedi-Yu, Roberto Dess{\`\i}, Roberta Raileanu, Maria Lomeli, Eric Hambro, Luke Zettlemoyer, Nicola Cancedda, and Thomas Scialom.
\newblock Toolformer: Language models can teach themselves to use tools.
\newblock \emph{Advances in neural information \mbox{processing systems}},~36:\mbox{68539--68551},~2023.

\bibitem[Schrodi et~al.(2025)Schrodi, Hoffmann, Argus, Fischer, and Brox]{schrodi2025two}
Simon Schrodi, David~T. Hoffmann, Max Argus, Volker Fischer, and Thomas Brox.
\newblock Two effects, one trigger: On the modality gap, object bias, and information imbalance in contrastive vision-language models.
\newblock In \emph{International Conference on \mbox{Learning Representations}},~2025.
\newblock URL \url{https://arxiv.org/abs/2404.07983}.

\bibitem[Shen et~al.(2023)Shen, Song, Tan, Li, Lu, and Zhuang]{shen2023hugginggpt}
Yongliang Shen, Kaitao Song, Xu~Tan, Dongsheng Li, Weiming Lu, and Yueting Zhuang.
\newblock {HuggingGPT}: Solving {AI} tasks with {ChatGPT} and its friends in {Hugging Face}.
\newblock \emph{Advances in Neural Information \mbox{Processing Systems}},~36:\mbox{38154--38180},~2023.

\bibitem[Tenenbaum et~al.(2000)Tenenbaum, de~Silva, and Langford]{tenenbaum2000isomap}
Joshua~B Tenenbaum, Vin de~Silva, and John~C Langford.
\newblock A global geometric framework for nonlinear dimensionality reduction.
\newblock \emph{science},~290\penalty0 (5500):\mbox{2319--2323},~2000.

\bibitem[Tsai et~al.(2019)Tsai, Bai, Liang, Kolter, Morency, and Salakhutdinov]{tsai-etal-2019-multimodal}
Yao-Hung~Hubert Tsai, Shaojie Bai, Paul~Pu Liang, J~Zico Kolter, Louis-Philippe Morency, and Ruslan Salakhutdinov.
\newblock Multimodal transformer for unaligned multimodal language sequences.
\newblock In \emph{Proceedings of the 57th annual meeting of the association for \mbox{computational linguistics}},~pp.\  \mbox{6558--6569},~2019.

\bibitem[Tupakula(2025)]{tupakula2025thin}
Mallikarjuna Tupakula.
\newblock Thin bridges for drug text alignment: Lightweight contrastive learning for target specific drug retrieval.
\newblock \emph{arXiv \mbox{preprint arXiv:2510.03309}},~2025.

\bibitem[Wang \& Isola(2020)Wang and Isola]{wang2020understanding}
Tongzhou Wang and Phillip Isola.
\newblock Understanding contrastive representation learning through alignment and uniformity on the hypersphere.
\newblock In \emph{International conference on \mbox{machine learning}},~pp.\  \mbox{9929--9939}. PMLR,~2020.

\bibitem[Wu et~al.(2023)Wu, Chen, Zhang, Hui, Berg-Kirkpatrick, and Dubnov]{wu2024largescaleclap}
Yusong Wu, Ke~Chen, Tianyu Zhang, Yuchen Hui, Taylor Berg-Kirkpatrick, and Shlomo Dubnov.
\newblock Large-scale contrastive language-audio pretraining with feature fusion and keyword-to-caption augmentation.
\newblock In \emph{ICASSP 2023-2023 IEEE International Conference on Acoustics, Speech and Signal \mbox{Processing (ICASSP)}},~pp.\  \mbox{1--5}. IEEE,~2023.

\bibitem[Xue et~al.(2026)Xue, Zhang, Xue, Liu, Wang, and Han]{xue2025multibenchpp}
Leyan Xue, Changqing Zhang, Kecheng Xue, Xiaohong Liu, Guangyu Wang, and Zongbo Han.
\newblock {MULTIBENCH++}: A unified and comprehensive multimodal fusion benchmarking across specialized domains.
\newblock In \emph{Proceedings of the AAAI Conference on \mbox{Artificial Intelligence}},~volume~40,~pp.\  \mbox{27450--27458},~2026.
\newblock \mbox{\doi{10.1609/aaai.v40i32.39963}}.

\bibitem[Yang et~al.(2025)Yang, Li, Yang, Zhang, Hui, Zheng, Yu, Gao, Huang, Lv, Zheng, Liu, Zhou, Huang, Hu, Ge, Wei, Lin, Tang, Yang, Tu, Zhang, Yang, Yang, Zhou, Zhou, Lin, Dang, Bao, Yang, Yu, Deng, Li, Xue, Li, Zhang, Wang, Zhu, Men, Gao, Liu, Luo, Li, Tang, Yin, Ren, Wang, Zhang, Ren, Fan, Su, Zhang, Zhang, Wan, Liu, Wang, Cui, Zhang, Zhou, and Qiu]{yang2025qwen3}
An~Yang, Anfeng Li, Baosong Yang, Beichen Zhang, Binyuan Hui, Bo~Zheng, Bowen Yu, Chang Gao, Chengen Huang, Chenxu Lv, Chujie Zheng, Dayiheng Liu, Fan Zhou, Fei Huang, Feng Hu, Hao Ge, Haoran Wei, Huan Lin, Jialong Tang, Jian Yang, Jianhong Tu, Jianwei Zhang, Jianxin Yang, Jiaxi Yang, Jing Zhou, Jingren Zhou, Junyang Lin, Kai Dang, Keqin Bao, Kexin Yang, Le~Yu, Lianghao Deng, Mei Li, Mingfeng Xue, Mingze Li, Pei Zhang, Peng Wang, Qin Zhu, Rui Men, Ruize Gao, Shixuan Liu, Shuang Luo, Tianhao Li, Tianyi Tang, Wenbiao Yin, Xingzhang Ren, Xinyu Wang, Xinyu Zhang, Xuancheng Ren, Yang Fan, Yang Su, Yichang Zhang, Yinger Zhang, Yu~Wan, Yuqiong Liu, Zekun Wang, Zeyu Cui, Zhenru Zhang, Zhipeng Zhou, and Zihan Qiu.
\newblock {Qwen3} technical report.
\newblock \emph{arXiv \mbox{preprint arXiv:2505.09388}},~2025.

\bibitem[Yang et~al.(2023)Yang, Li, Wang, Lin, Azarnasab, Ahmed, Liu, Liu, Zeng, and Wang]{yang2023mmreact}
Zhengyuan Yang, Linjie Li, Jianfeng Wang, Kevin Lin, Ehsan Azarnasab, Faisal Ahmed, Zicheng Liu, Ce~Liu, Michael Zeng, and Lijuan Wang.
\newblock {MM-REACT}: Prompting {ChatGPT} for multimodal reasoning and action.
\newblock \emph{arXiv \mbox{preprint arXiv:2303.11381}},~2023.

\bibitem[Yao et~al.(2023)Yao, Zhao, Yu, Du, Shafran, Narasimhan, and Cao]{yao2023react}
Shunyu Yao, Jeffrey Zhao, Dian Yu, Nan Du, Izhak Shafran, Karthik Narasimhan, and Yuan Cao.
\newblock {ReAct}: Synergizing reasoning and acting in language models.
\newblock In \emph{International Conference on Learning \mbox{Representations (ICLR)}},~2023.

\bibitem[You et~al.(2021)You, Liu, Wang, and Long]{you2021logme}
Kaichao You, Yong Liu, Jianmin Wang, and Mingsheng Long.
\newblock {LogME}: Practical assessment of pre-trained models for transfer learning.
\newblock In \emph{International Conference on Machine \mbox{Learning (ICML)}},~volume 139 of \emph{PMLR},~pp.\  \mbox{12133--12143},~2021.

\bibitem[Yue et~al.(2024)Yue, Ni, Zhang, Zheng, Liu, Zhang, Stevens, Jiang, Ren, Sun, Wei, Yu, Yuan, Sun, Yin, Zheng, Yang, Liu, Huang, Sun, Su, and Chen]{Yue_2024_CVPR}
Xiang Yue, Yuansheng Ni, Kai Zhang, Tianyu Zheng, Ruoqi Liu, Ge~Zhang, Samuel Stevens, Dongfu Jiang, Weiming Ren, Yuxuan Sun, Cong Wei, Botao Yu, Ruibin Yuan, Renliang Sun, Ming Yin, Boyuan Zheng, Zhenzhu Yang, Yibo Liu, Wenhao Huang, Huan Sun, Yu~Su, and Wenhu Chen.
\newblock {MMMU}: A massive multi-discipline multimodal understanding and reasoning benchmark for expert {AGI}.
\newblock In \emph{Proceedings of the IEEE/CVF conference on computer vision and \mbox{pattern recognition}},~pp.\  \mbox{9556--9567},~2024.

\bibitem[Zhai et~al.(2023)Zhai, Mustafa, Kolesnikov, and Beyer]{Zhai_2023_ICCV}
Xiaohua Zhai, Basil Mustafa, Alexander Kolesnikov, and Lucas Beyer.
\newblock Sigmoid loss for language image pre-training.
\newblock In \emph{Proceedings of the IEEE/CVF international conference on \mbox{computer vision}},~pp.\  \mbox{11975--11986},~2023.

\bibitem[Zhang et~al.(2025{\natexlab{a}})Zhang, Yang, and Agrawal]{zhang2025assessing}
Le~Zhang, Qian Yang, and Aishwarya Agrawal.
\newblock Assessing and learning alignment of unimodal vision and language models.
\newblock In \emph{Proceedings of the IEEE/CVF Conference on Computer Vision and Pattern \mbox{Recognition (CVPR)}},~pp.\  \mbox{14604--14614},~2025{\natexlab{a}}.

\bibitem[Zhang et~al.(2025{\natexlab{b}})Zhang, Xu, Usuyama, Xu, Bagga, Tinn, Preston, Rao, Wei, Valluri, Wong, Tupini, Wang, Mazzola, Shukla, Liden, Gao, Crabtree, Piening, Bifulco, Lungren, Naumann, Wang, and Poon]{zhang2023biomedclip}
Sheng Zhang, Yanbo Xu, Naoto Usuyama, Hanwen Xu, Jaspreet Bagga, Robert Tinn, Sam Preston, Rajesh Rao, Mu~Wei, Naveen Valluri, Cliff Wong, Andrea Tupini, Yu~Wang, Matt Mazzola, Swadheen Shukla, Lars Liden, Jianfeng Gao, Angela Crabtree, Brian Piening, Carlo Bifulco, Matthew~P. Lungren, Tristan Naumann, Sheng Wang, and Hoifung Poon.
\newblock A multimodal biomedical foundation model trained from fifteen million image--text pairs.
\newblock \emph{\mbox{NEJM AI}},~2\penalty0 (1),~2025{\natexlab{b}}.
\newblock \mbox{\doi{10.1056/AIoa2400640}}.

\end{thebibliography}

\appendix

\addtocontents{toc}{\protect\setcounter{tocdepth}{2}}
\renewcommand{\contentsname}{Appendix Contents}
\tableofcontents
\clearpage

\renewcommand{\topfraction}{0.95}
\renewcommand{\textfraction}{0.05}
\renewcommand{\floatpagefraction}{0.85}
\setcounter{topnumber}{3}
\setcounter{totalnumber}{4}
\section{Additional Results and Ablations}
\label{app:additional-results}

This appendix expands Section~\ref{sec:experiments}. Budget curves and seed stability appear in Sections~\ref{app:figures} and~\ref{app:perseed}. Fusion-head, scaling, and domain-pretraining ablations are in Sections~\ref{app:fusion-ablation}, \ref{app:scaling-ablation}, and~\ref{app:domain-pretraining-ablation}. Sections~\ref{app:text-tower-ablation}, \ref{app:agent-ablations}, and~\ref{app:agent-controls} examine individual modalities and Agent components. Routing comparisons, utility differences, and task-level failures are in Sections~\ref{app:baselines}, \ref{app:utility-routing}, and~\ref{app:per-dataset-routing}. Section~\ref{app:regression-parsing} reports output-format failures.

\subsection{Win Map and Label-Budget Curves}
\label{app:figures}

\begin{figure}[t]
    \centering
    \includegraphics[width=0.95\textwidth]{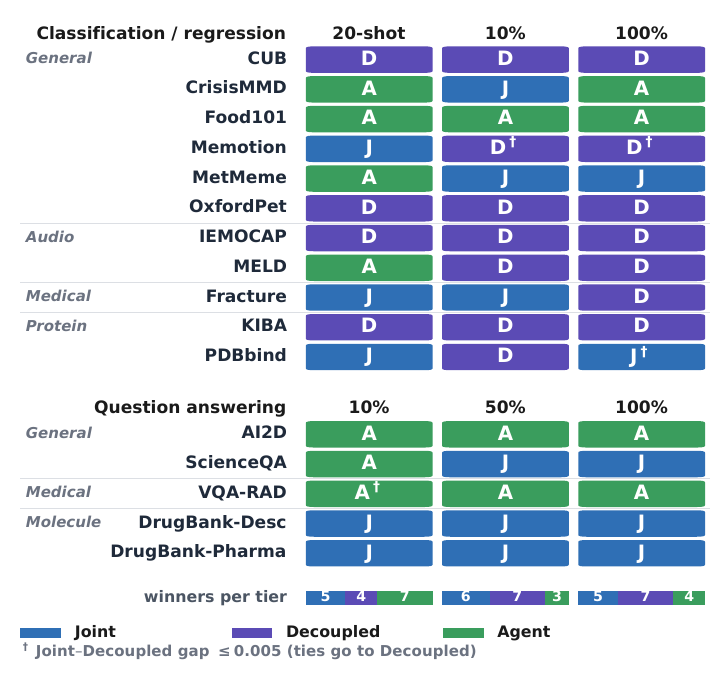}
    \caption{\textbf{Win map of the $48$ main-table cells}, coloured and lettered by the winning paradigm. $\dagger$ marks the four cells with a \Joint--\Decoupled gap of at most $0.005$. Rows are grouped by task type, which use different budgets, and ordered by domain. \mbox{The bottom strip counts winners per budget tier.}}
    \label{fig:win_heatmap}
\end{figure}

\begin{figure}[t]
    \centering
    \includegraphics[width=\textwidth]{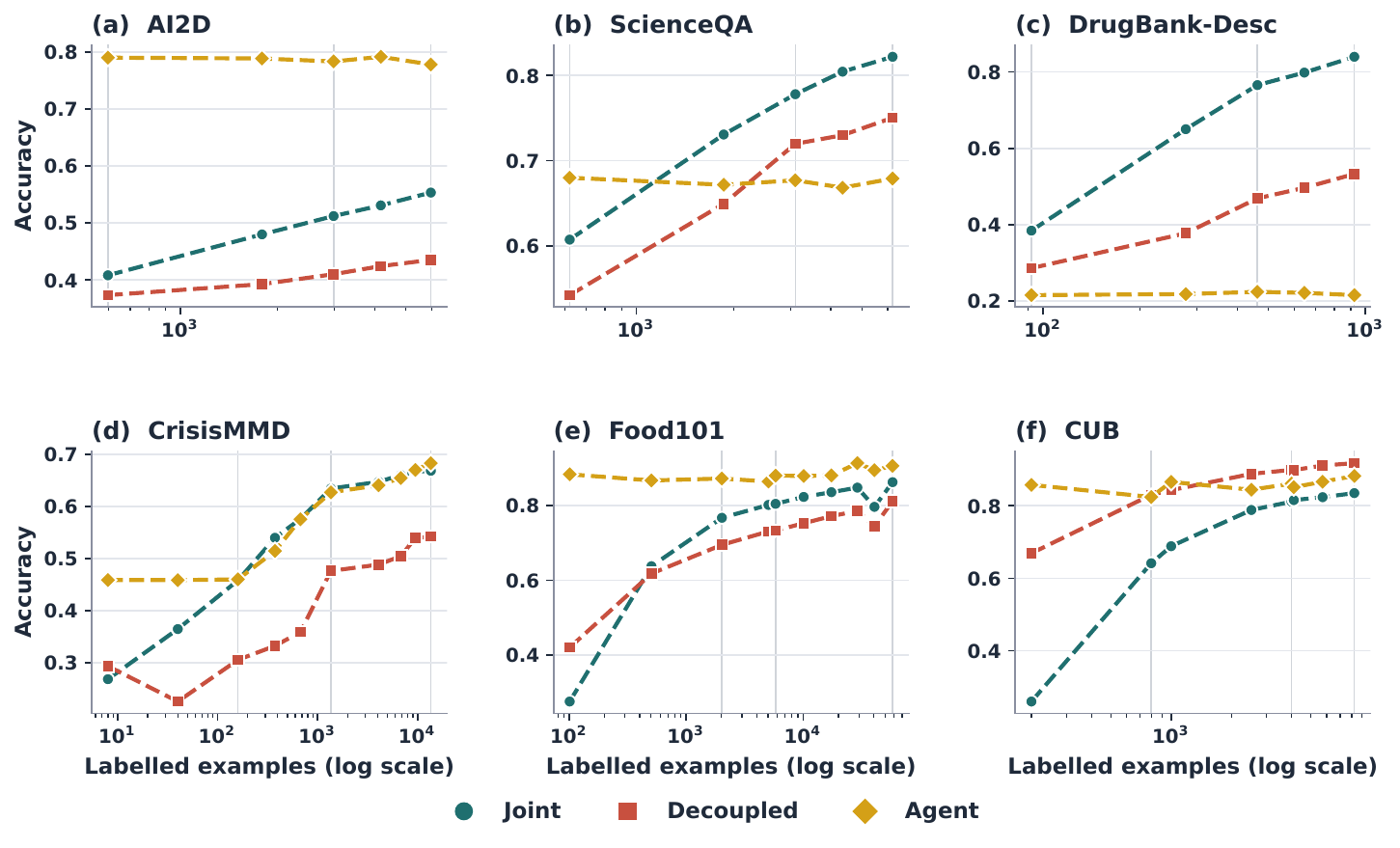}
    \caption{\textbf{Learning curves for six datasets} over each dataset's full budget grid (log scale). Vertical lines mark the budgets of the main tables. Teal = \Joint, coral = \Decoupled, ochre = \Agent. The curves distinguish persistent preferences from budget-dependent reversals. The budget dependence of \Agent\ includes \mbox{changes in its fitted readouts and available demonstrations.}}
    \label{fig:budget_curves}
\end{figure}

Figure~\ref{fig:win_heatmap} shows winners at the three representative budgets in the main tables. Figure~\ref{fig:budget_curves} traces utility over the full budget grids of six datasets.

\subsection{Per-Seed Variance and Winner Stability of the Main Tables}
\label{app:perseed}

Every \Joint\ and \Decoupled\ entry in Tables~\ref{tab:agent_fm_cls_reg} and~\ref{tab:agent_fm_qa} is a mean over $5$ seed-specific development-selected models, the seed setting the labelled set below $100\%$ and the head initialisation at every budget. The seed-resolved results used below contain five observations for each configuration, including \Agent. Table~\ref{tab:perseed} lists the per-seed standard deviations for the CLS/REG cells, the mean utility gap, and its descriptive status. The better of \Joint\ and \Decoupled\ is decided by the sign of the mean gap, with ties going to \Decoupled. A cell is flagged a \emph{near-tie} if $|\mathrm{gap}| \le 0.005$, \emph{noisy} if the mean winner holds but $|\mathrm{gap}|$ is below the combined standard deviation $\sqrt{s_J^2 + s_D^2}$, and \emph{clear} otherwise, as descriptive seed-spread categories rather than significance tests: $23$ cells are clear, $7$ noisy, and $3$ near-ties. The near-ties are Memotion at $10\%$ and $100\%$ and PDBbind at $100\%$.

\begin{table}[t]
\centering
\footnotesize
\setlength{\tabcolsep}{4pt}
\renewcommand{\arraystretch}{1.08}
\caption{\textbf{Per-seed variance of Table~\ref{tab:agent_fm_cls_reg}.} Score columns report mean $\pm$ std over $5$ seeds: accuracy for CLS and positive MAE for REG. Gap is $U_J-U_D$, with utility equal to accuracy or $-$MAE. A positive gap favours \Joint. \textbf{Bold} marks the better of \Joint\ and \Decoupled, ties going to \Decoupled. Combined std $= \sqrt{s_J^2 + s_D^2}$. A cell is a near-tie if $|\mathrm{gap}| \le 0.005$, noisy if $|\mathrm{gap}|$ \mbox{is below the combined std, and clear otherwise.}}
\label{tab:perseed}
\begin{tabular}{@{}l l c c r c l@{}}
\toprule
\textbf{Dataset} & \textbf{Budget} & \textbf{\Joint} & \textbf{\Decoupled} & \textbf{Gap} & \textbf{Comb.\ std} & \textbf{Status} \\
\midrule
\rowcolor{gray!10}\multicolumn{7}{c}{\textbf{\textit{Classification (utility $=$ accuracy)}}} \\
\midrule
\multirow{3}{*}{CUB} & 20-shot & $0.811\pm0.004$ & $\mathbf{0.899}\pm0.002$ & $-0.088$ & $0.005$ & clear \\
 & 10\% & $0.641\pm0.013$ & $\mathbf{0.830}\pm0.009$ & $-0.189$ & $0.016$ & clear \\
 & 100\% & $0.833\pm0.004$ & $\mathbf{0.917}\pm0.001$ & $-0.084$ & $0.004$ & clear \\
\addlinespace[2pt]
\multirow{3}{*}{CrisisMMD} & 20-shot & $\mathbf{0.458}\pm0.032$ & $0.306\pm0.077$ & $+0.153$ & $0.083$ & clear \\
 & 10\% & $\mathbf{0.634}\pm0.008$ & $0.476\pm0.011$ & $+0.158$ & $0.013$ & clear \\
 & 100\% & $\mathbf{0.668}\pm0.007$ & $0.538\pm0.029$ & $+0.130$ & $0.030$ & clear \\
\addlinespace[2pt]
\multirow{3}{*}{Food101} & 20-shot & $\mathbf{0.767}\pm0.003$ & $0.695\pm0.003$ & $+0.072$ & $0.004$ & clear \\
 & 10\% & $\mathbf{0.804}\pm0.001$ & $0.733\pm0.002$ & $+0.072$ & $0.003$ & clear \\
 & 100\% & $\mathbf{0.862}\pm0.000$ & $0.808\pm0.000$ & $+0.054$ & $0.000$ & clear \\
\addlinespace[2pt]
\multirow{3}{*}{Fracture} & 20-shot & $\mathbf{0.294}\pm0.028$ & $0.247\pm0.032$ & $+0.047$ & $0.043$ & clear \\
 & 10\% & $\mathbf{0.245}\pm0.063$ & $0.199\pm0.022$ & $+0.046$ & $0.066$ & noisy \\
 & 100\% & $0.388\pm0.017$ & $\mathbf{0.405}\pm0.019$ & $-0.017$ & $0.025$ & noisy \\
\addlinespace[2pt]
\multirow{3}{*}{MELD} & 20-shot & $0.375\pm0.104$ & $\mathbf{0.445}\pm0.042$ & $-0.070$ & $0.112$ & noisy \\
 & 10\% & $0.535\pm0.007$ & $\mathbf{0.564}\pm0.005$ & $-0.029$ & $0.009$ & clear \\
 & 100\% & $0.571\pm0.004$ & $\mathbf{0.637}\pm0.010$ & $-0.066$ & $0.011$ & clear \\
\addlinespace[2pt]
\multirow{3}{*}{Memotion} & 20-shot & $\mathbf{0.420}\pm0.067$ & $0.349\pm0.106$ & $+0.071$ & $0.126$ & noisy \\
 & 10\% & $0.454\pm0.000$ & $\mathbf{0.454}\pm0.000$ & $+0.000$ & $0.000$ & near-tie$^\dagger$ \\
 & 100\% & $0.447\pm0.002$ & $\mathbf{0.451}\pm0.001$ & $-0.004$ & $0.002$ & near-tie$^\dagger$ \\
\addlinespace[2pt]
\multirow{3}{*}{MetMeme} & 20-shot & $\mathbf{0.252}\pm0.008$ & $0.208\pm0.016$ & $+0.044$ & $0.018$ & clear \\
 & 10\% & $\mathbf{0.319}\pm0.006$ & $0.266\pm0.022$ & $+0.053$ & $0.023$ & clear \\
 & 100\% & $\mathbf{0.409}\pm0.010$ & $0.323\pm0.003$ & $+0.086$ & $0.011$ & clear \\
\addlinespace[2pt]
\multirow{3}{*}{OxfordPet} & 20-shot & $0.918\pm0.006$ & $\mathbf{0.949}\pm0.003$ & $-0.031$ & $0.006$ & clear \\
 & 10\% & $0.836\pm0.020$ & $\mathbf{0.928}\pm0.007$ & $-0.093$ & $0.021$ & clear \\
 & 100\% & $0.945\pm0.001$ & $\mathbf{0.962}\pm0.001$ & $-0.017$ & $0.001$ & clear \\
\midrule
\rowcolor{gray!10}\multicolumn{7}{c}{\textbf{\textit{Regression (scores: MAE $\downarrow$, gap: $U_J-U_D$)}}} \\
\midrule
\multirow{3}{*}{IEMOCAP} & 20-shot & $0.725\pm0.015$ & $\mathbf{0.680}\pm0.017$ & $-0.045$ & $0.022$ & clear \\
 & 10\% & $0.648\pm0.005$ & $\mathbf{0.606}\pm0.013$ & $-0.041$ & $0.014$ & clear \\
 & 100\% & $0.556\pm0.007$ & $\mathbf{0.517}\pm0.010$ & $-0.039$ & $0.012$ & clear \\
\addlinespace[2pt]
\multirow{3}{*}{KIBA} & 20-shot & $0.589\pm0.009$ & $\mathbf{0.581}\pm0.008$ & $-0.008$ & $0.012$ & noisy \\
 & 10\% & $0.517\pm0.004$ & $\mathbf{0.434}\pm0.006$ & $-0.083$ & $0.007$ & clear \\
 & 100\% & $0.420\pm0.004$ & $\mathbf{0.344}\pm0.004$ & $-0.076$ & $0.006$ & clear \\
\addlinespace[2pt]
\multirow{3}{*}{PDBbind} & 20-shot & $\mathbf{1.230}\pm0.023$ & $1.261\pm0.028$ & $+0.031$ & $0.036$ & noisy \\
 & 10\% & $1.193\pm0.023$ & $\mathbf{1.170}\pm0.008$ & $-0.023$ & $0.024$ & noisy \\
 & 100\% & $\mathbf{1.160}\pm0.008$ & $1.165\pm0.003$ & $+0.005$ & $0.008$ & near-tie$^\dagger$ \\
\bottomrule
\end{tabular}
\end{table}

Table~\ref{tab:uncertainty} reports empirical winner counts from the five corresponding seed scores of all three configurations. Accuracy is maximized and MAE is minimized within each seed. A sole winner receives one win. Configurations tied for best share one win equally, so every row sums to five. The reported scores have three decimal places: equal values are ties at the reported precision, not evidence of statistical equivalence. Across $240$ comparisons, fractional counts are $81.5$, $87.0$, and $71.5$ for \Joint, \Decoupled, and \Agent, respectively. Dividing by five gives seed-averaged counts of $16.3$, $17.4$, and $14.3$ cells, whose sum is $48$. These differ from the $16/18/14$ mean-utility winner counts because the order of averaging and winner selection differs and this table shares tie credit rather than applying the preference rule. The same configuration is the sole winner in all five seeds in $38$ cells. The other ten show a change of winner or a best-score tie. There are seven tied-best seed comparisons. For example, CrisisMMD at $20$-shot gives $2/0/3$ wins for \Joint/\Decoupled/\Agent, while Memotion at $20$-shot gives $2.5/2.5/0$. They describe variability across the five runs on fixed test examples. No test examples are resampled in this analysis, so these counts do not quantify test-sampling uncertainty \mbox{or a probability of winning on future examples.}

\begin{table}[t]
\centering
\footnotesize
\setlength{\tabcolsep}{9pt}
\renewcommand{\arraystretch}{1.02}
\caption{\textbf{Five-seed winner stability for the $48$ main-table cells.} Each row counts wins by \Joint, \Decoupled, and \Agent\ over five corresponding seeds. A sole winner receives one win. Tied best configurations share one win equally, so every row sums to five. Higher accuracy wins for classification/QA and lower MAE for regression. Counts use the supplied three-decimal seed scores, so ties mean equality at that precision, not statistical equivalence. \mbox{Bold marks \mbox{the largest count in a row.}}}
\label{tab:uncertainty}
\begin{tabular}{@{}l l r r r@{}}
\toprule
\textbf{Dataset} & \textbf{Budget} & \textbf{Joint wins} & \textbf{Decoupled wins} & \textbf{Agent wins} \\
\midrule
\rowcolor{gray!10}\multicolumn{5}{l}{\textit{Classification}} \\
\multirow{3}{*}{CUB} & 20-shot & 0 & \textbf{5} & 0 \\
 & 10\% & 0 & 2 & \textbf{3} \\
 & 100\% & 0 & \textbf{5} & 0 \\
\addlinespace[2pt]
\multirow{3}{*}{CrisisMMD} & 20-shot & 2 & 0 & \textbf{3} \\
 & 10\% & \textbf{3.5} & 0 & 1.5 \\
 & 100\% & 1 & 0 & \textbf{4} \\
\addlinespace[2pt]
\multirow{3}{*}{Food101} & 20-shot & 0 & 0 & \textbf{5} \\
 & 10\% & 0 & 0 & \textbf{5} \\
 & 100\% & 0 & 0 & \textbf{5} \\
\addlinespace[2pt]
\multirow{3}{*}{Fracture} & 20-shot & \textbf{5} & 0 & 0 \\
 & 10\% & \textbf{5} & 0 & 0 \\
 & 100\% & 1 & \textbf{4} & 0 \\
\addlinespace[2pt]
\multirow{3}{*}{MELD} & 20-shot & 0 & 0 & \textbf{5} \\
 & 10\% & 0 & \textbf{5} & 0 \\
 & 100\% & 0 & \textbf{5} & 0 \\
\addlinespace[2pt]
\multirow{3}{*}{Memotion} & 20-shot & \textbf{2.5} & \textbf{2.5} & 0 \\
 & 10\% & \textbf{2.5} & \textbf{2.5} & 0 \\
 & 100\% & 0 & \textbf{5} & 0 \\
\addlinespace[2pt]
\multirow{3}{*}{MetMeme} & 20-shot & 0 & 0 & \textbf{5} \\
 & 10\% & \textbf{5} & 0 & 0 \\
 & 100\% & \textbf{5} & 0 & 0 \\
\addlinespace[2pt]
\multirow{3}{*}{OxfordPet} & 20-shot & 0 & \textbf{5} & 0 \\
 & 10\% & 0 & \textbf{5} & 0 \\
 & 100\% & 0 & \textbf{5} & 0 \\
\midrule
\rowcolor{gray!10}\multicolumn{5}{l}{\textit{Regression}} \\
\multirow{3}{*}{IEMOCAP} & 20-shot & 0 & \textbf{5} & 0 \\
 & 10\% & 0 & \textbf{5} & 0 \\
 & 100\% & 0 & \textbf{5} & 0 \\
\addlinespace[2pt]
\multirow{3}{*}{KIBA} & 20-shot & 1 & \textbf{4} & 0 \\
 & 10\% & 0 & \textbf{5} & 0 \\
 & 100\% & 0 & \textbf{5} & 0 \\
\addlinespace[2pt]
\multirow{3}{*}{PDBbind} & 20-shot & \textbf{4} & 1 & 0 \\
 & 10\% & 0 & \textbf{5} & 0 \\
 & 100\% & \textbf{4} & 1 & 0 \\
\midrule
\rowcolor{gray!10}\multicolumn{5}{l}{\textit{Question answering}} \\
\multirow{3}{*}{AI2D} & 10\% & 0 & 0 & \textbf{5} \\
 & 50\% & 0 & 0 & \textbf{5} \\
 & 100\% & 0 & 0 & \textbf{5} \\
\addlinespace[2pt]
\multirow{3}{*}{DrugBank-Desc} & 10\% & \textbf{5} & 0 & 0 \\
 & 50\% & \textbf{5} & 0 & 0 \\
 & 100\% & \textbf{5} & 0 & 0 \\
\addlinespace[2pt]
\multirow{3}{*}{DrugBank-Pharma} & 10\% & \textbf{5} & 0 & 0 \\
 & 50\% & \textbf{5} & 0 & 0 \\
 & 100\% & \textbf{5} & 0 & 0 \\
\addlinespace[2pt]
\multirow{3}{*}{ScienceQA} & 10\% & 0 & 0 & \textbf{5} \\
 & 50\% & \textbf{5} & 0 & 0 \\
 & 100\% & \textbf{5} & 0 & 0 \\
\addlinespace[2pt]
\multirow{3}{*}{VQA-RAD} & 10\% & 0 & 0 & \textbf{5} \\
 & 50\% & 0 & 0 & \textbf{5} \\
 & 100\% & 0 & 0 & \textbf{5} \\
\bottomrule
\end{tabular}
\end{table}

\subsection{Fusion Strategy Ablation}
\label{app:fusion-ablation}

Tables~\ref{tab:fusion_decoupled} and~\ref{tab:fusion_joint} compare fixed fusion heads at $100\%$ supervision on the default encoder pairs. The main tables instead average seed-specific development-selected heads, so their means can differ from every fixed-head entry. Restricting \Joint\ to a single concat head instead would change the winner on one of the sixteen datasets at $100\%$, PDBbind ($1.160 \to 1.187$ MAE), from \Joint\ to \Decoupled.

\begin{table}[t]
\centering
\caption{\textbf{Fusion-head grid for \Decoupled at the $100\%$ budget.} $5$-seed mean test score of each head on the default encoder pair. \textbf{Bold} marks the best per-head mean test score: highest accuracy or lowest MAE. Main-table means can combine different heads across seeds. Accuracy ($\uparrow$) for CLS and QA, MAE ($\downarrow$) for REG.}
\label{tab:fusion_decoupled}
\footnotesize
\setlength{\tabcolsep}{4pt}
\begin{adjustbox}{max width=\textwidth}
\begin{tabular}{l cccccccc}
\toprule
\textbf{Dataset} & \textbf{add} & \textbf{attention} & \textbf{bilinear} & \textbf{concat} & \textbf{film} & \textbf{gated} & \textbf{logit\_sum} & \textbf{mlp} \\
\midrule
\rowcolor{gray!10}\multicolumn{9}{c}{\textbf{\textit{Classification (accuracy $\uparrow$)}}} \\
\midrule
CUB & 0.909 & 0.911 & 0.876 & 0.917 & 0.906 & 0.909 & \textbf{0.917} & 0.893 \\
CrisisMMD & 0.529 & \textbf{0.542} & 0.510 & 0.494 & 0.524 & 0.534 & 0.497 & 0.473 \\
Food101 & 0.803 & 0.804 & 0.777 & 0.808 & 0.796 & 0.810 & 0.808 & \textbf{0.811} \\
Fracture & 0.379 & \textbf{0.407} & 0.402 & 0.303 & 0.387 & 0.399 & 0.304 & 0.373 \\
MELD & 0.627 & 0.616 & 0.629 & 0.515 & \textbf{0.639} & 0.636 & 0.513 & 0.620 \\
Memotion & 0.399 & 0.405 & 0.387 & \textbf{0.454} & 0.388 & 0.397 & 0.445 & 0.396 \\
MetMeme & 0.322 & 0.284 & 0.300 & 0.317 & 0.316 & \textbf{0.324} & 0.317 & 0.296 \\
OxfordPet & \textbf{0.964} & 0.960 & 0.958 & 0.958 & 0.963 & 0.963 & 0.958 & 0.961 \\
\midrule
\rowcolor{gray!10}\multicolumn{9}{c}{\textbf{\textit{QA (accuracy $\uparrow$)}}} \\
\midrule
AI2D & 0.428 & \textbf{0.435} & 0.417 & 0.409 & 0.424 & 0.422 & 0.328 & 0.410 \\
DrugBank-Desc & 0.467 & 0.479 & 0.469 & 0.512 & 0.481 & 0.517 & 0.252 & \textbf{0.533} \\
DrugBank-Pharma & 0.580 & 0.527 & \textbf{0.596} & 0.590 & 0.527 & 0.560 & 0.257 & 0.594 \\
ScienceQA & 0.706 & 0.701 & \textbf{0.750} & 0.693 & 0.750 & 0.730 & 0.460 & 0.643 \\
VQA-RAD & 0.302 & 0.305 & 0.308 & 0.305 & \textbf{0.321} & 0.310 & 0.262 & 0.320 \\
\midrule
\rowcolor{gray!10}\multicolumn{9}{c}{\textbf{\textit{Regression (MAE $\downarrow$)}}} \\
\midrule
IEMOCAP & 0.548 & \textbf{0.515} & 0.522 & 0.551 & 0.519 & 0.530 & 0.545 & 0.526 \\
KIBA & 0.476 & 0.346 & 0.422 & 0.490 & 0.419 & 0.349 & 0.482 & \textbf{0.342} \\
PDBbind & 1.175 & 1.189 & 1.184 & \textbf{1.162} & 1.204 & 1.172 & 1.163 & 1.200 \\
\bottomrule
\end{tabular}
\end{adjustbox}
\end{table}

\begin{table}[t]
\centering
\caption{\textbf{Fusion-head grid for \Joint at the $100\%$ budget.} $5$-seed mean test score of each head on the default encoder pair. \textbf{Bold} marks the best per-head mean test score: highest accuracy or lowest MAE. Main-table means can combine different heads across seeds. Accuracy ($\uparrow$) for CLS and QA, MAE ($\downarrow$) for REG.}
\label{tab:fusion_joint}
\footnotesize
\setlength{\tabcolsep}{4pt}
\begin{adjustbox}{max width=\textwidth}
\begin{tabular}{l cccccccc}
\toprule
\textbf{Dataset} & \textbf{add} & \textbf{attention} & \textbf{bilinear} & \textbf{concat} & \textbf{film} & \textbf{gated} & \textbf{logit\_sum} & \textbf{mlp} \\
\midrule
\rowcolor{gray!10}\multicolumn{9}{c}{\textbf{\textit{Classification (accuracy $\uparrow$)}}} \\
\midrule
CUB & 0.823 & \textbf{0.835} & 0.801 & 0.823 & 0.831 & 0.831 & 0.823 & 0.801 \\
CrisisMMD & 0.665 & 0.633 & \textbf{0.668} & 0.639 & 0.668 & 0.665 & 0.639 & 0.649 \\
Food101 & 0.853 & 0.840 & 0.832 & 0.862 & 0.850 & 0.854 & \textbf{0.862} & 0.855 \\
Fracture & 0.376 & 0.334 & 0.386 & 0.322 & 0.370 & 0.380 & 0.323 & \textbf{0.389} \\
MELD & 0.561 & 0.558 & 0.564 & 0.551 & \textbf{0.572} & 0.568 & 0.551 & 0.569 \\
Memotion & 0.419 & 0.433 & 0.415 & 0.449 & 0.413 & 0.418 & \textbf{0.450} & 0.424 \\
MetMeme & \textbf{0.410} & 0.346 & 0.390 & 0.408 & 0.393 & 0.408 & 0.408 & 0.391 \\
OxfordPet & 0.936 & 0.945 & 0.939 & 0.912 & \textbf{0.946} & 0.944 & 0.912 & 0.934 \\
\midrule
\rowcolor{gray!10}\multicolumn{9}{c}{\textbf{\textit{QA (accuracy $\uparrow$)}}} \\
\midrule
AI2D & 0.481 & 0.459 & 0.469 & 0.549 & 0.513 & 0.499 & 0.383 & \textbf{0.553} \\
DrugBank-Desc & 0.703 & 0.714 & 0.805 & 0.786 & 0.772 & 0.688 & 0.260 & \textbf{0.840} \\
DrugBank-Pharma & 0.723 & 0.671 & 0.669 & \textbf{0.772} & 0.687 & 0.737 & 0.216 & 0.758 \\
ScienceQA & 0.775 & 0.782 & 0.804 & 0.815 & 0.807 & 0.787 & 0.527 & \textbf{0.822} \\
VQA-RAD & 0.282 & \textbf{0.302} & 0.278 & 0.293 & 0.282 & 0.265 & 0.262 & 0.298 \\
\midrule
\rowcolor{gray!10}\multicolumn{9}{c}{\textbf{\textit{Regression (MAE $\downarrow$)}}} \\
\midrule
IEMOCAP & 0.589 & 0.559 & 0.582 & 0.593 & 0.577 & 0.557 & 0.588 & \textbf{0.555} \\
KIBA & 0.517 & 0.469 & 0.494 & 0.522 & 0.492 & \textbf{0.420} & 0.521 & 0.490 \\
PDBbind & 1.180 & \textbf{1.158} & 1.163 & 1.187 & 1.194 & 1.192 & 1.179 & 1.185 \\
\bottomrule
\end{tabular}
\end{adjustbox}
\end{table}

\begin{figure}[t]
    \centering
    \includegraphics[width=0.92\textwidth]{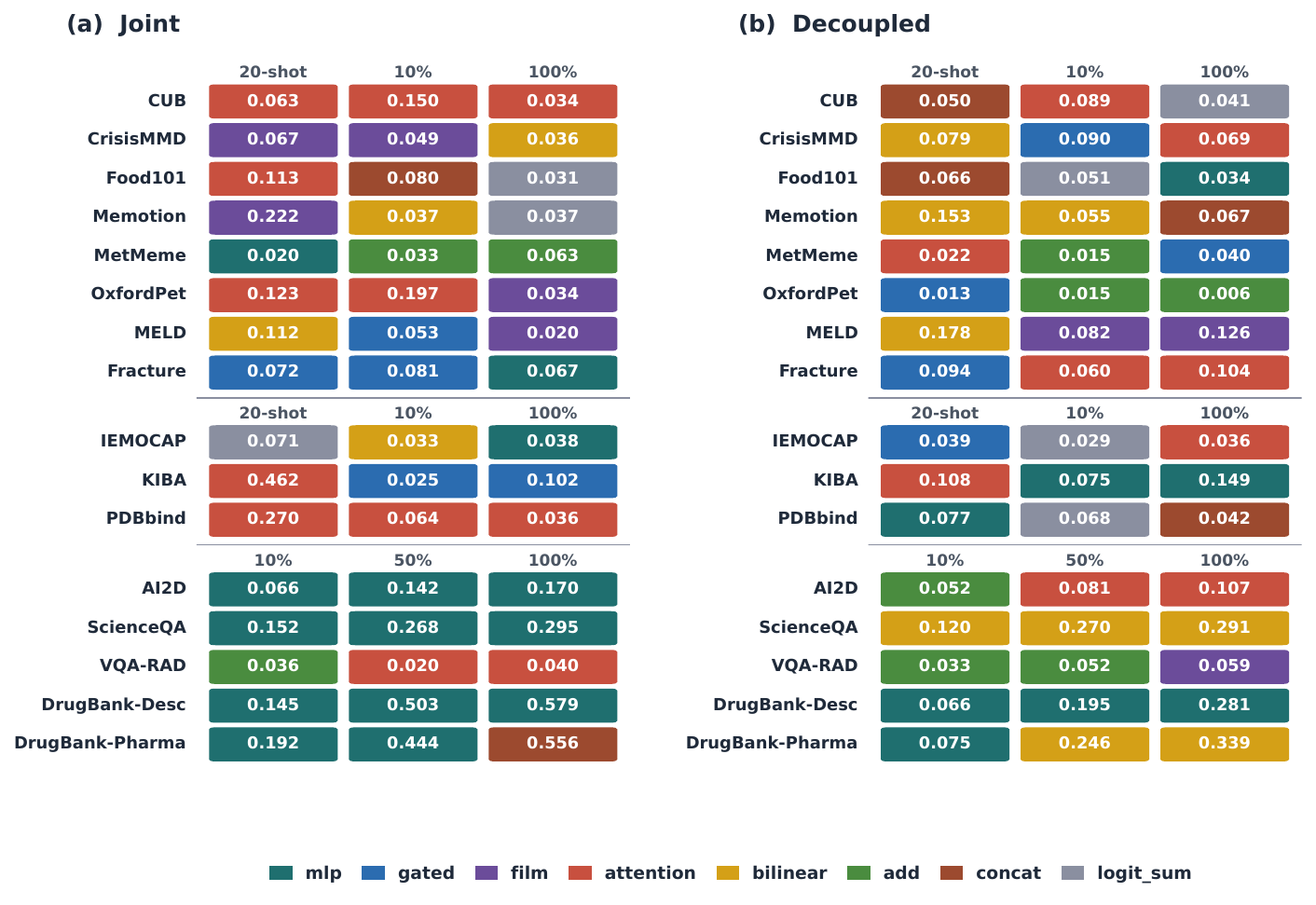}
    \caption{\textbf{Diagnostic test-best fusion head per cell}, coloured by the head with the best five-seed mean on the default pair and labelled with the spread between the best and the worst of the eight heads (task metric units). The spread is widest on the MCQ tasks and at the smallest budgets, and at most $0.13$ \mbox{on classification at full budget.}}
    \label{fig:fusion_map}
\end{figure}

\begin{figure}[t]
    \centering
    \includegraphics[width=0.85\textwidth]{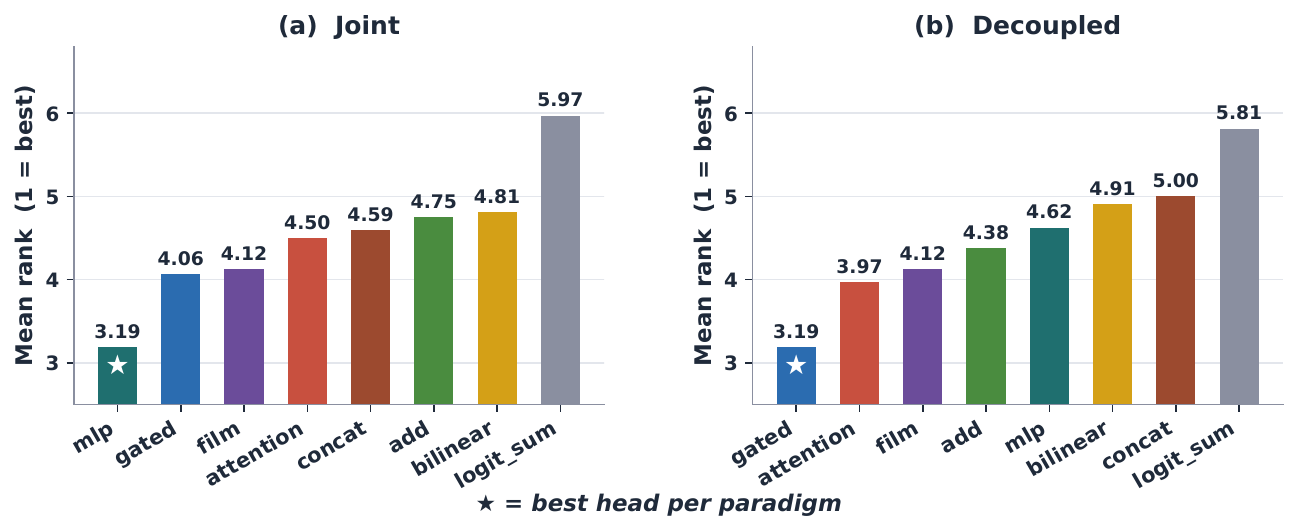}
    \caption{\textbf{Diagnostic test-performance rank of fusion heads} across $16$ datasets at $100\%$ \mbox{supervision. $\bigstar$ = paradigm best.}}
    \label{fig:fusion_rank}
\end{figure}

No single head dominates. For \Decoupled\ (Table~\ref{tab:fusion_decoupled}), \texttt{attention} is best on CrisisMMD, Fracture, AI2D and IEMOCAP, \texttt{mlp} on Food101, DrugBank-Desc and KIBA, \texttt{film} on MELD and VQA-RAD, \texttt{concat} on Memotion and PDBbind, \texttt{bilinear} on DrugBank-Pharma and, tied with \texttt{film}, on ScienceQA, \texttt{gated} on MetMeme and \texttt{add} on OxfordPet, and \texttt{concat} and \texttt{logit\_sum} tie on CUB, while \texttt{logit\_sum}, which only adds readout logits, is weakest on the molecular QA tasks ($0.252$ on DrugBank-Desc). The preferred heads also differ between paradigms: on ScienceQA, \Decoupled\ reaches $0.750$ with either \texttt{bilinear} or \texttt{film}, whereas \Joint\ reaches $0.822$ with \texttt{mlp} (Table~\ref{tab:fusion_joint}). Across the three reported budgets the best head changes on $15$ of the $16$ dataset--paradigm pairs of the classification datasets, but the spread between the best and the worst head tells where that choice matters: at $100\%$ it is $0.006$ to $0.13$ on classification and $0.11$ to $0.58$ on the MCQ tasks, and at $20$-shot it reaches $0.22$ on Memotion. Figures~\ref{fig:fusion_map} and~\ref{fig:fusion_rank} show the \mbox{head choice across budgets and the mean ranks.}

\subsection{Backbone Scaling Ablation}
\label{app:scaling-ablation}

\begin{figure}[t]
    \centering
    \includegraphics[width=\textwidth]{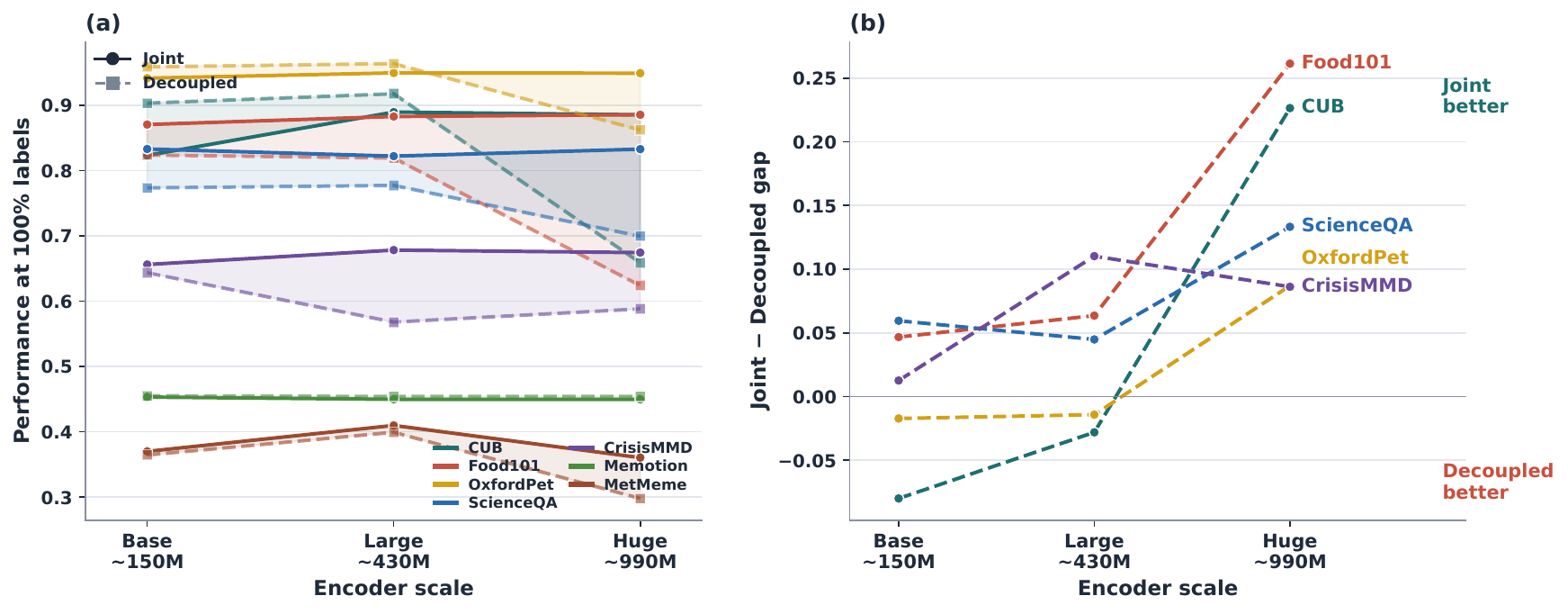}
    \caption{\textbf{Backbone scaling ablation.} Left: per-dataset scaling curves at full budget for the CLIP/SigLIP/EVA02 family. Solid lines = \Joint, dashed lines = \Decoupled. Right: the \Joint\ minus \Decoupled\ gap across scale tiers (base/large/huge). Datasets above the zero line favour \Joint\ and datasets below favour \Decoupled. The gap keeps its sign from base to large on both datasets. At the huge tier the only matched unaligned tower is MAE-ViT-H/14 (Table~\ref{tab:scaling}).}
    \label{fig:fm_scaling}
\end{figure}

Figure~\ref{fig:fm_scaling} compares base ($\sim$150M parameters), large ($\sim$430M), and huge ($\sim$990M) towers within the CLIP/SigLIP/EVA02 family. Each tier reports the pair with the highest five-seed test mean, forming a test-oracle envelope rather than the development-selected comparison used in the main benchmark. On CUB, \Decoupled\ wins at the base and large tiers at $100\%$ supervision (Table~\ref{tab:scaling}). On Food101, \Joint\ wins at every tier and its lead grows from $0.046$ at the base tier to $0.064$ at the large tier, because \Joint\ improves ($0.870$, $0.883$) while \Decoupled\ stays flat ($0.824$, $0.819$). The huge tier is a different comparison. No DINOv2 checkpoint exists at that size, so the only unaligned tower matched to OpenCLIP-ViT-H/14 is MAE-ViT-H/14, and \Decoupled\ falls to $0.659$ on CUB and $0.624$ on Food101. The huge-tier gap therefore reflects a joint change in scale and encoder family.

\begin{table}[t]
\centering
\footnotesize
\setlength{\tabcolsep}{4pt}
\caption{\textbf{Backbone scaling at the $100\%$ budget}, best encoder pair within each tier. \textbf{Bold} marks the better of \Joint\ and \Decoupled\ per dataset and tier. On CUB \Decoupled\ wins at the base and large tiers, and on Food101 \Joint\ wins at every tier. The huge tier has no DINOv2 checkpoint, and its only matched unaligned tower is MAE-H/14. $^{a}$On Food101 the best large-tier \Decoupled\ pair is \mbox{CLIP-L vision + GPT-2.}}
\label{tab:scaling}
\adjustbox{max width=\textwidth}{%
\begin{tabular}{@{}l l cc l cc@{}}
\toprule
 & \multicolumn{3}{c}{\textbf{\Joint}} & \multicolumn{3}{c}{\textbf{\Decoupled}} \\
\cmidrule(lr){2-4}\cmidrule(lr){5-7}
\textbf{Scale tier} & \textbf{Best pair} & \textbf{CUB} & \textbf{Food101} & \textbf{Best pair} & \textbf{CUB} & \textbf{Food101} \\
\midrule
Base ($\sim$150M) & SigLIP-B/16 & 0.823 & \textbf{0.870} & DINOv2-B + BERT & \textbf{0.903} & 0.824 \\
Large ($\sim$430M) & EVA02-L/14 & 0.889 & \textbf{0.883} & DINOv2-L + GPT-2$^{a}$ & \textbf{0.917} & 0.819 \\
Huge ($\sim$990M) & OpenCLIP-H/14 & \textbf{0.885} & \textbf{0.885} & MAE-H + GPT-2-medium & 0.659 & 0.624 \\
\bottomrule
\end{tabular}}
\end{table}

\subsection{Domain Pretraining Ablation}
\label{app:domain-pretraining-ablation}

Figure~\ref{fig:domain_pretraining} compares general and domain-specific encoders on medical and audio data (setup in Appendix~\ref{app:domain-ablation}). For \Joint\ on Fracture at $100\%$, the general CLIP-ViT-B/16 reaches $0.322$ and all three biomedical FMs are higher (PLIP $0.367$, BiomedCLIP $0.388$, PubMedCLIP $0.419$), a gain of up to $9.7$ points. For \Decoupled\ the picture is the opposite: the general DINOv2-B/14 tower ($0.405$) stays ahead of the radiology and pathology encoders (Rad-DINO $0.316$, Phikon $0.343$) at almost every budget. On MELD, the general-audio CLAP ($0.571$) is well ahead of a music-domain CLAP ($0.481$), a wrong-domain control whose head collapses to the majority class at every percentage budget, and \Decoupled\ leads \Joint\ at every percentage budget with any of the three audio encoders. Changing the encoder configuration affects both absolute accuracy and the system ranking: on Fracture at $100\%$ the default BiomedCLIP trails \Decoupled\ by $0.017$ and PubMedCLIP would lead it by $0.014$. PubMedCLIP also changes the visual patch size and text architecture relative to the default \Decoupled\ pair (Appendix~\ref{app:domain-ablation}), so this reversal reflects the full configuration change. At the two smaller Fracture budgets \Joint\ wins with either backbone.

\begin{figure}[t]
  \centering
  \includegraphics[width=\textwidth]{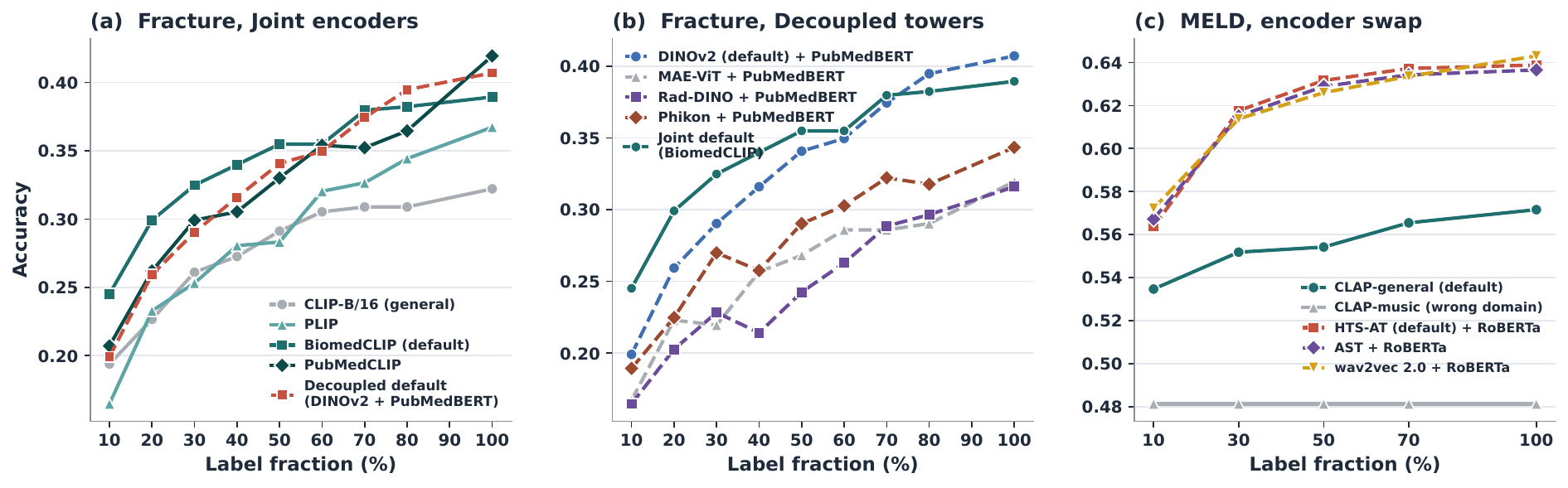}
  \caption{\textbf{General and domain-specific encoder configurations.} (a) Fracture, \Joint\ with the general CLIP-B/16 and three biomedical paired encoders: PubMedCLIP exceeds the default \Decoupled\ pair at the largest budgets, while also differing in visual patch size and text architecture. (b) Fracture, \Decoupled\ with domain-specific vision towers: none beats the general DINOv2. (c) MELD, general and music CLAP for \Joint\ and three audio encoders for \Decoupled, which leads at every percentage budget.}
  \label{fig:domain_pretraining}
\end{figure}

\subsection{Single-Modality Probes and the \Decoupled\ Pair}
\label{app:text-tower-ablation}

Table~\ref{tab:text_tower_ablation} places the default \Decoupled\ pair (DINOv2-L/14 + GPT-2) and the base-tier pair (DINOv2-B/14 + BERT-base) beside linear probes of the single frozen encoders and beside the matched \Joint\ FM (CLIP-ViT-L/14). Swapping the default pair for the base-tier pair changes \Decoupled\ by at most $0.016$ on CUB, OxfordPet and Food101 at $100\%$, and by $0.090$ on CrisisMMD ($0.628$ vs.\ $0.538$). The swap changes both towers, and on CrisisMMD the BERT probe ($0.588$) is also stronger than the GPT-2 probe ($0.471$).

\paragraph{Modality value and probe budget.}
On CUB, the DINOv2-L probe and default \Decoupled\ both reach $0.899$ at $20$-shot. At $10\%$ they reach $0.853$ and $0.830$, and at full budget $0.916$ and $0.917$. Adding text leaves accuracy unchanged at $20$-shot, lowers it at $10\%$, and improves it by only $0.1$ percentage points at full budget. Here $10\%$ supplies only $795$ training labels, whereas $20$-shot supplies $4{,}000$ ($20$ per class) and full budget $8{,}244$ (Table~\ref{tab:label-counts}), so both scores are lower at $10\%$ than at $20$-shot.

\paragraph{Why a single caption can be difficult to classify.}
CUB captions describe visible appearance, such as plumage colour, bill shape, and markings on individual body parts. The caption collection protocol asks annotators not to name the species and does not reveal the species label to them \citep{reed2016learning}. Our setup uses only the first of ten captions for each image. A single description can omit distinguishing details, while the attributes it mentions may be shared by several of the $200$ fine-grained bird categories. Identifying the species from that text alone can therefore be difficult, even when the description accurately matches the image. This ambiguity is a plausible contributor to the weak text-only results and the limited gain over the strong visual probe at larger budgets. Text representation and probe training may also contribute to this pattern.

\begin{table}[t]
\centering
\caption{\textbf{Single-modality probes, \Decoupled\ pairs and the matched \Joint\ FM in the general domain.} Higher is better ($\uparrow$). \textbf{Single Modality} = linear-probe accuracy of one frozen encoder. \textbf{Decoupled} = the default pair DINOv2-L/14 + GPT-2 and the base-tier pair DINOv2-B/14 + BERT-base, each at its best fusion head. \textbf{Joint} = CLIP ViT-L/14. All entries are $5$-seed means. \textbf{Bold} = best within each budget per row. CUB $10\%$ uses $795$ training labels, versus $4{,}000$ at $20$-shot. Its visual-probe accuracy is $0.853$ (Appendix~\ref{app:text-tower-ablation}).}
\label{tab:text_tower_ablation}
\footnotesize
\setlength{\tabcolsep}{4pt}
\begin{adjustbox}{max width=\textwidth}
\begin{tabular}{l cccccc cccccc cccccc}
\toprule
\textbf{Dataset}
  & \multicolumn{6}{c}{\textbf{20-shot}}
  & \multicolumn{6}{c}{\textbf{10\%}}
  & \multicolumn{6}{c}{\textbf{100\%}} \\
\cmidrule(lr){2-7}\cmidrule(lr){8-13}\cmidrule(lr){14-19}
  & \multicolumn{3}{c}{\textit{Single Mod.}} & \multicolumn{2}{c}{\textit{Decoupled}} & \textit{Joint}  & \multicolumn{3}{c}{\textit{Single Mod.}} & \multicolumn{2}{c}{\textit{Decoupled}} & \textit{Joint}  & \multicolumn{3}{c}{\textit{Single Mod.}} & \multicolumn{2}{c}{\textit{Decoupled}} & \textit{Joint} \\
\cmidrule(lr){2-4}\cmidrule(lr){5-6}\cmidrule(lr){7-7}\cmidrule(lr){8-10}\cmidrule(lr){11-12}\cmidrule(lr){13-13}\cmidrule(lr){14-16}\cmidrule(lr){17-18}\cmidrule(lr){19-19}
  & DINOv2-L & BERT & GPT-2 & L+GPT-2 & B+BERT & CLIP-L  & DINOv2-L & BERT & GPT-2 & L+GPT-2 & B+BERT & CLIP-L  & DINOv2-L & BERT & GPT-2 & L+GPT-2 & B+BERT & CLIP-L \\
\midrule
CUB & \textbf{0.899} & 0.020 & 0.015 & \textbf{0.899} & 0.884 & 0.811 & \textbf{0.853} & 0.006 & 0.006 & 0.830 & 0.809 & 0.641 & 0.916 & 0.032 & 0.020 & \textbf{0.917} & 0.903 & 0.833 \\
OxfordPet & 0.935 & 0.106 & 0.066 & \textbf{0.949} & 0.941 & 0.918 & 0.912 & 0.066 & 0.040 & \textbf{0.928} & 0.920 & 0.836 & 0.958 & 0.154 & 0.065 & \textbf{0.962} & 0.959 & 0.945 \\
Food101 & 0.691 & 0.137 & 0.082 & 0.695 & 0.674 & \textbf{0.767} & 0.727 & 0.205 & 0.087 & 0.733 & 0.727 & \textbf{0.804} & 0.792 & 0.450 & 0.143 & 0.808 & 0.824 & \textbf{0.862} \\
Memotion & 0.179 & \textbf{0.426} & 0.249 & 0.349 & 0.318 & 0.420 & 0.443 & \textbf{0.454} & \textbf{0.454} & \textbf{0.454} & 0.454 & \textbf{0.454} & 0.189 & \textbf{0.454} & \textbf{0.454} & 0.451 & 0.454 & 0.447 \\
CrisisMMD & 0.227 & 0.317 & 0.091 & 0.306 & 0.362 & \textbf{0.458} & 0.473 & 0.187 & 0.376 & 0.476 & 0.556 & \textbf{0.634} & 0.489 & 0.588 & 0.471 & 0.538 & 0.628 & \textbf{0.668} \\
\bottomrule
\end{tabular}
\end{adjustbox}
\end{table}

\paragraph{Text-only comparison on Memotion.}
On Memotion at $20$-shot, the BERT probe reaches $0.426$, above default \Decoupled\ at $0.349$. BERT is an alternative text encoder, whereas the default pair uses GPT-2. This comparison shows a competitive single-modality configuration. The comparison varies both modality access and the text encoder.

\subsection{Agent Ablations}
\label{app:agent-ablations}

Three ablations vary one component of the \Agent\ at a time, with Qwen3-32B, \mbox{Inline-1 and $N_{\mathrm{ICL}}{=}10$ as defaults.}

\begin{figure}[t]
    \centering
    \includegraphics[width=0.95\textwidth]{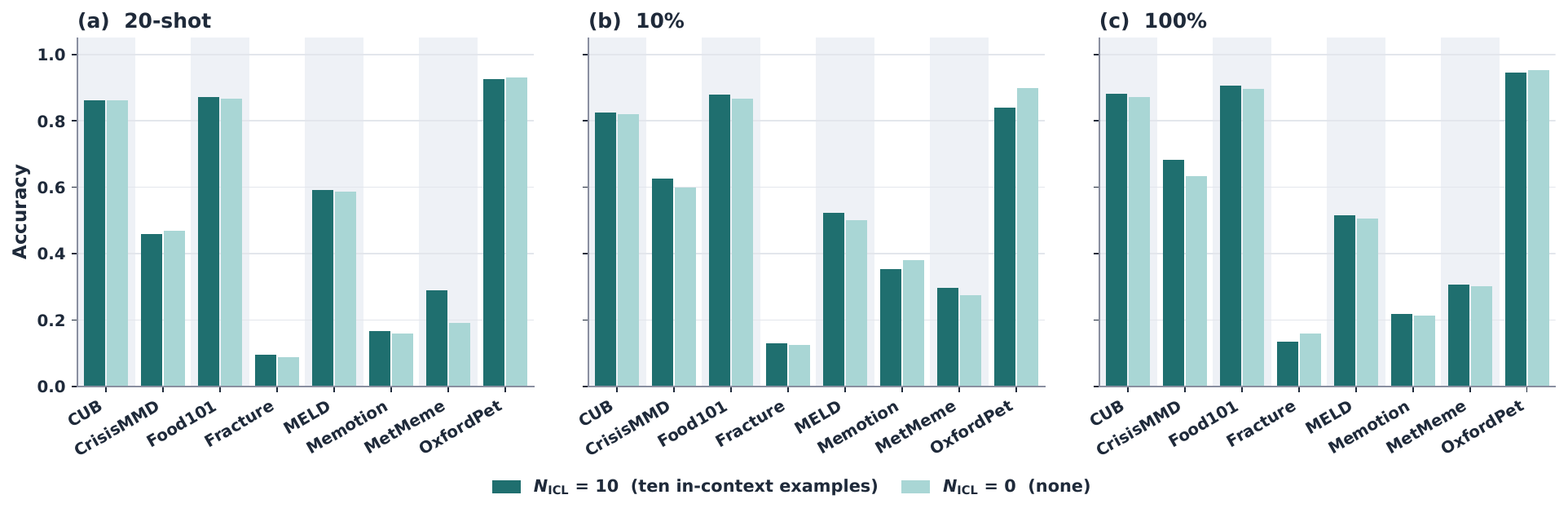}
    \caption{\textbf{In-context examples ($N_{\mathrm{ICL}}{=}10$ vs.\ $0$).} Per-dataset \Agent\ accuracy at the three label budgets. In-context examples help most on MetMeme and CrisisMMD and cost up to $0.057$ on OxfordPet. Numerical values \mbox{in Table~\ref{tab:kshot_cls}.}}
    \label{fig:agent_kshot_main}
\end{figure}

\paragraph{In-context examples (Figure~\ref{fig:agent_kshot_main}).}
Table~\ref{tab:kshot_cls} compares $N_{\mathrm{ICL}}{=}10$ in-context examples with none on the eight classification datasets. The mean gain is below one point. The largest gains are on MetMeme at $20$-shot ($+0.099$) and CrisisMMD at $100\%$ ($+0.049$), and the examples hurt on OxfordPet ($-0.057$ at $10\%$). On MetMeme at $20$-shot, removing demonstrations lowers \Agent\ from $0.290$ to $0.191$, below \Joint\ at $0.252$. On OxfordPet at $10\%$, removal raises \Agent\ from $0.841$ to $0.898$, while \Decoupled\ remains ahead at $0.928$. The two examples distinguish a reversal of paradigm preference from an accuracy change that preserves the winner.

\begin{table}[t]
\centering
\caption{\textbf{In-context example ablation} (LLM = Qwen3-32B, Inline-1). $N_{\mathrm{ICL}}{=}10$ is the main configuration and $N_{\mathrm{ICL}}{=}0$ removes demonstrations but retains the supervised readouts. $\Delta = \mathrm{Accuracy}_{10} - \mathrm{Accuracy}_{0}$. CLS datasets only. \textbf{Bold} \mbox{marks the largest gain and the largest loss.}}
\label{tab:kshot_cls}
\footnotesize
\setlength{\tabcolsep}{4pt}
\begin{adjustbox}{max width=\textwidth}
\begin{tabular}{l ccc ccc ccc}
\toprule
\textbf{Dataset}
  & \multicolumn{3}{c}{\textbf{20-shot}}
  & \multicolumn{3}{c}{\textbf{10\%}}
  & \multicolumn{3}{c}{\textbf{100\%}} \\
\cmidrule(lr){2-4}\cmidrule(lr){5-7}\cmidrule(lr){8-10}
  & $N_{\mathrm{ICL}}{=}10$ & $N_{\mathrm{ICL}}{=}0$ & $\Delta$
  & $N_{\mathrm{ICL}}{=}10$ & $N_{\mathrm{ICL}}{=}0$ & $\Delta$
  & $N_{\mathrm{ICL}}{=}10$ & $N_{\mathrm{ICL}}{=}0$ & $\Delta$ \\
\midrule
CUB       & 0.861 & 0.862 & $-$0.001          & 0.824 & 0.821 & $+$0.003          & 0.882 & 0.872 & $+$0.010 \\
CrisisMMD & 0.460 & 0.469 & $-$0.009          & 0.627 & 0.598 & $+$0.029          & 0.683 & 0.634 & $+$0.049 \\
Food101   & 0.872 & 0.866 & $+$0.006          & 0.880 & 0.868 & $+$0.012          & 0.906 & 0.897 & $+$0.009 \\
Fracture  & 0.095 & 0.088 & $+$0.007          & 0.131 & 0.124 & $+$0.007          & 0.136 & 0.159 & $-$0.023 \\
MELD      & 0.591 & 0.586 & $+$0.005          & 0.523 & 0.502 & $+$0.021          & 0.516 & 0.506 & $+$0.010 \\
Memotion  & 0.166 & 0.159 & $+$0.007          & 0.354 & 0.380 & $-$0.026          & 0.219 & 0.213 & $+$0.006 \\
MetMeme   & 0.290 & 0.191 & \textbf{$+$0.099} & 0.296 & 0.276 & $+$0.020 & 0.307 & 0.302 & $+$0.005 \\
OxfordPet & 0.927 & 0.932 & $-$0.005          & 0.841 & 0.898 & $\mathbf{-}$0.057 & 0.945 & 0.954 & $-$0.009 \\
\bottomrule
\end{tabular}
\end{adjustbox}
\end{table}

\begin{figure}[t]
    \centering
    \includegraphics[width=0.95\textwidth]{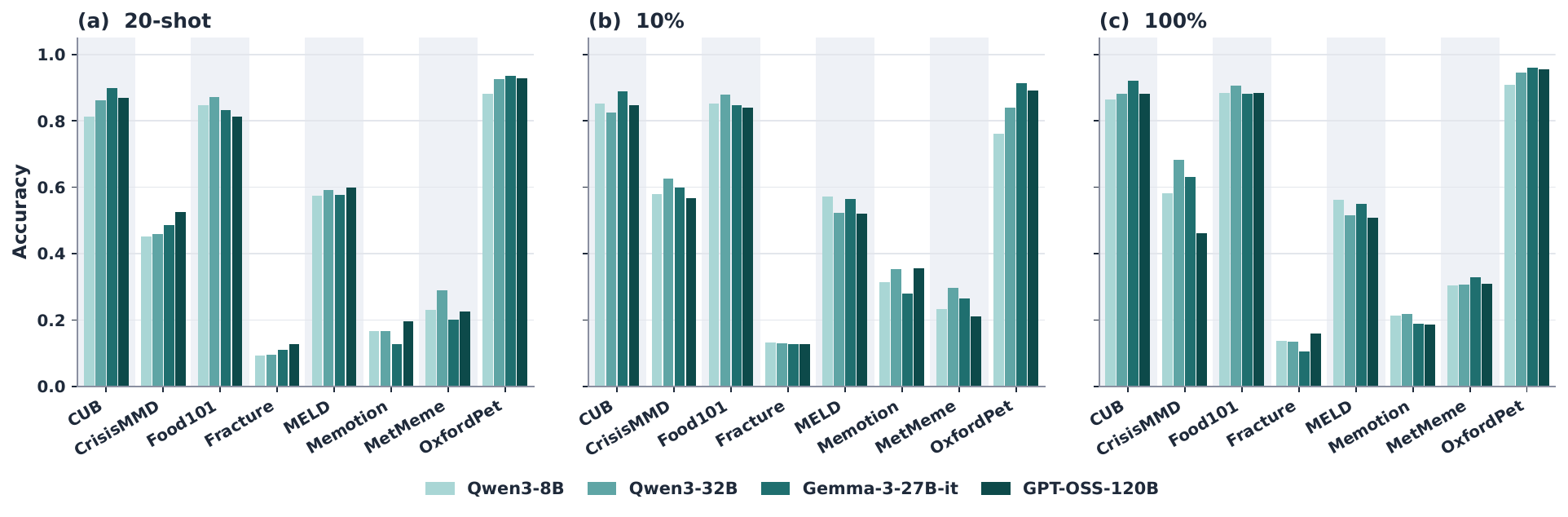}
    \caption{\textbf{LLM backbone for the deployed Agent.} Per-dataset Agent utility under Qwen3-8B, Qwen3-32B (deployed), Gemma-3-27B-it, and GPT-OSS-120B at the reported label budgets. At $20$-shot Qwen3-8B is below Qwen3-32B on $7$ of the $8$ classification datasets. \mbox{Numerical values in Tables~\ref{tab:agent_llm_cls_reg} and~\ref{tab:agent_llm_qa}.}}
    \label{fig:agent_llm_main}
\end{figure}

\paragraph{LLM backbone choice (Figure~\ref{fig:agent_llm_main}).}
Tables~\ref{tab:agent_llm_cls_reg} and~\ref{tab:agent_llm_qa} compare Qwen3-8B, Qwen3-32B, Gemma-3-27B-it \citep{gemmateam2025gemma3}, and GPT-OSS-120B \citep{openai2025gptoss} under Inline-1. No backbone dominates: Gemma-3-27B-it is best on CUB and OxfordPet at every budget and Qwen3-32B on Food101. Measured against the better fusion paradigm of Tables~\ref{tab:agent_fm_cls_reg} and~\ref{tab:agent_fm_qa}, the \Agent-versus-fusion preference usually agrees across all four backbones on their shared tasks and budgets. All four win AI2D at $10\%$ and $100\%$, Food101 at every budget, MELD at $20$-shot, and VQA-RAD at $10\%$, and none wins any regression, DrugBank, Fracture, Memotion, or OxfordPet cell. Some preferences change with the backbone. Accuracy differences are usually a few points but reach $0.22$ on CrisisMMD and $0.15$ on OxfordPet.

\begin{table}[t]
\centering
\caption{\textbf{Agent LLM comparison on CLS \& REG datasets} (Inline-1). Higher is better ($\uparrow$) for accuracy. Lower is better ($\downarrow$) for MAE on REG (IEMOCAP, KIBA, PDBbind). \textbf{Bold} = best backbone within each budget.}
\label{tab:agent_llm_cls_reg}
\footnotesize
\setlength{\tabcolsep}{4pt}
\begin{adjustbox}{max width=\textwidth}
\begin{tabular}{l cccc cccc cccc}
\toprule
\textbf{Dataset}
  & \multicolumn{4}{c}{\textbf{20-shot}}
  & \multicolumn{4}{c}{\textbf{10\%}}
  & \multicolumn{4}{c}{\textbf{100\%}} \\
\cmidrule(lr){2-5}\cmidrule(lr){6-9}\cmidrule(lr){10-13}
  & \textbf{Qwen3-8B} & \textbf{Qwen3-32B} & \textbf{Gemma-27B} & \textbf{GPT-120B}
  & \textbf{Qwen3-8B} & \textbf{Qwen3-32B} & \textbf{Gemma-27B} & \textbf{GPT-120B}
  & \textbf{Qwen3-8B} & \textbf{Qwen3-32B} & \textbf{Gemma-27B} & \textbf{GPT-120B} \\
\midrule
\rowcolor{gray!10}\multicolumn{13}{c}{\textbf{\textit{Classification (accuracy $\uparrow$)}}} \\
\midrule
CUB       & 0.814 & 0.861 & \textbf{0.900} & 0.869 & 0.853 & 0.824 & \textbf{0.888} & 0.848 & 0.865 & 0.882 & \textbf{0.921} & 0.882 \\
CrisisMMD & 0.452 & 0.460 & 0.485 & \textbf{0.526} & 0.580 & \textbf{0.627} & 0.600 & 0.567 & 0.581 & \textbf{0.683} & 0.632 & 0.462 \\
Food101   & 0.848 & \textbf{0.872} & 0.832 & 0.814 & 0.852 & \textbf{0.880} & 0.847 & 0.839 & 0.884 & \textbf{0.906} & 0.881 & 0.883 \\
Fracture  & 0.093 & 0.095 & 0.111 & \textbf{0.128} & \textbf{0.133} & 0.131 & 0.128 & 0.128 & 0.137 & 0.136 & 0.106 & \textbf{0.159} \\
MELD      & 0.574 & 0.591 & 0.576 & \textbf{0.598} & \textbf{0.571} & 0.523 & 0.564 & 0.521 & \textbf{0.562} & 0.516 & 0.550 & 0.509 \\
Memotion  & 0.166 & 0.166 & 0.127 & \textbf{0.197} & 0.314 & 0.354 & 0.280 & \textbf{0.357} & 0.213 & \textbf{0.219} & 0.189 & 0.186 \\
MetMeme   & 0.231 & \textbf{0.290} & 0.200 & 0.225 & 0.234 & \textbf{0.296} & 0.266 & 0.211 & 0.305 & 0.307 & \textbf{0.330} & 0.309 \\
OxfordPet & 0.882 & 0.927 & \textbf{0.936} & 0.928 & 0.761 & 0.841 & \textbf{0.913} & 0.892 & 0.908 & 0.945 & \textbf{0.960} & 0.955 \\
\midrule
\rowcolor{gray!10}\multicolumn{13}{c}{\textbf{\textit{Regression (MAE $\downarrow$)}}} \\
\midrule
IEMOCAP & \textbf{0.725} & 0.730 & 0.732 & 0.729 & \textbf{0.714} & 0.717 & 0.729 & 0.716 & 0.592 & \textbf{0.584} & 0.587 & \textbf{0.584} \\
KIBA    & 1.674 & \textbf{1.535} & 1.711 & 1.631 & 0.717 & 0.655 & \textbf{0.552} & 0.554 & 0.961 & 0.817 & 0.524 & \textbf{0.520} \\
PDBbind & 1.442 & 1.531 & \textbf{1.347} & 1.371 & 1.383 & 1.477 & 1.328 & \textbf{1.309} & 1.296 & 1.360 & \textbf{1.249} & 1.254 \\
\bottomrule
\end{tabular}
\end{adjustbox}
\end{table}

\begin{table}[t]
\centering
\caption{\textbf{Agent LLM comparison on QA datasets} (Inline-1, accuracy $\uparrow$). \textbf{Bold} = \mbox{best backbone within each budget.}}
\label{tab:agent_llm_qa}
\footnotesize
\setlength{\tabcolsep}{4pt}
\begin{adjustbox}{max width=\textwidth}
\begin{tabular}{l cccc cccc}
\toprule
\textbf{Dataset}
  & \multicolumn{4}{c}{\textbf{10\%}}
  & \multicolumn{4}{c}{\textbf{100\%}} \\
\cmidrule(lr){2-5}\cmidrule(lr){6-9}
  & \textbf{Qwen3-8B} & \textbf{Qwen3-32B} & \textbf{Gemma-27B} & \textbf{GPT-120B}
  & \textbf{Qwen3-8B} & \textbf{Qwen3-32B} & \textbf{Gemma-27B} & \textbf{GPT-120B} \\
\midrule
AI2D            & 0.727 & 0.790 & 0.746 & \textbf{0.794} & 0.732 & 0.778 & 0.740 & \textbf{0.791} \\
DrugBank-Desc   & 0.224 & 0.216 & \textbf{0.233} & 0.198 & 0.190 & 0.216 & \textbf{0.224} & 0.216 \\
DrugBank-Pharma & 0.238 & 0.277 & \textbf{0.287} & 0.238 & 0.238 & 0.277 & \textbf{0.287} & 0.208 \\
ScienceQA       & 0.623 & 0.680 & 0.604 & \textbf{0.684} & 0.634 & 0.679 & 0.589 & \textbf{0.685} \\
VQA-RAD         & 0.319 & 0.325 & \textbf{0.333} & 0.324 & 0.306 & \textbf{0.375} & 0.317 & 0.326 \\
\bottomrule
\end{tabular}
\end{adjustbox}
\end{table}

\begin{figure}[t]
    \centering
    \includegraphics[width=0.95\textwidth]{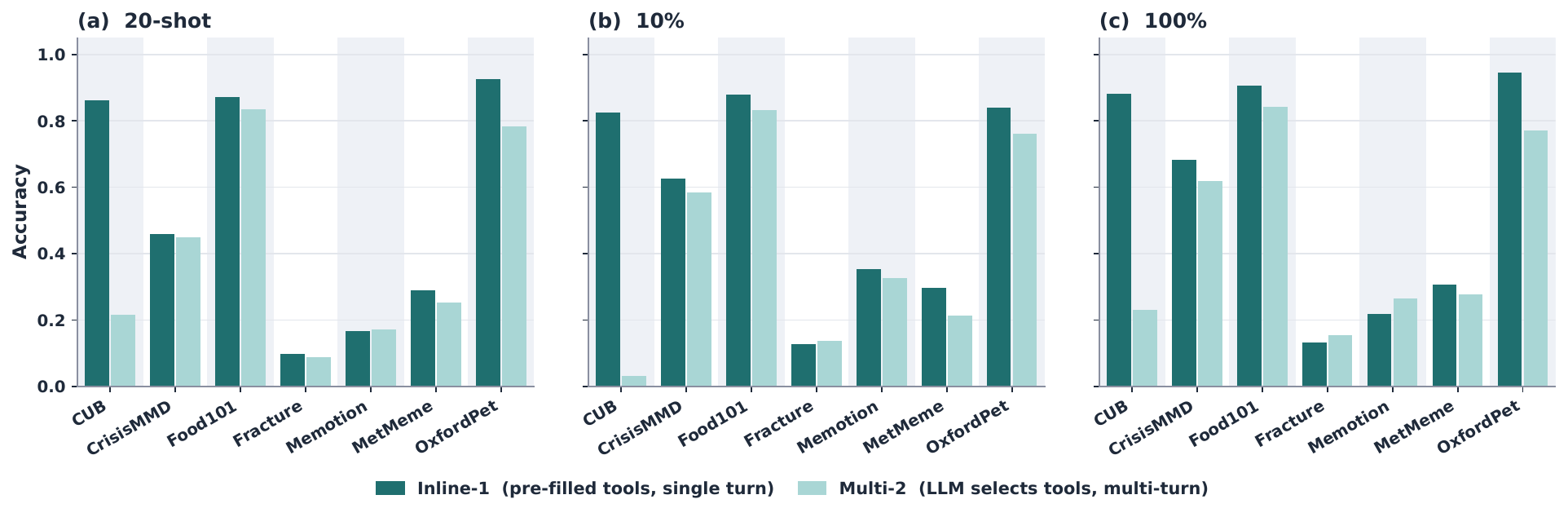}
    \caption{\textbf{Tool-dispatch mode (Inline-1 vs.\ Multi-2).} Per-dataset accuracy under the two dispatch modes on the seven classification datasets where both readouts were available to the Multi-2 controller. Multi-2 is usually less accurate, most visibly on CUB and OxfordPet, because the controller often calls the uninformative readout. Inline-1, deployed \mbox{throughout, uses one LLM turn instead of two.}}
    \label{fig:agent_dispatch_main}
\end{figure}

\paragraph{Tool-dispatch mode (Figure~\ref{fig:agent_dispatch_main}).}
\emph{Inline-1} pre-computes both readouts and provides them in a single LLM turn. \emph{Multi-2} shows the LLM the two tool schemas, lets it call one tool, and takes its answer in a second turn. The additional turn changes the interaction protocol, with computation and latency also depending on token counts and serving conditions. We compare the two on the seven classification datasets for which both readouts were available to the Multi-2 controller. Multi-2 is usually less accurate across the three reported budgets, with a mean gap of $-0.136$ ($-0.043$ excluding CUB). The controller makes close to one tool call per sample and often picks the uninformative readout: on CUB it calls the text readout on $79\%$ of samples, and accuracy falls from $0.86$ to $0.16$ averaged over budgets. OxfordPet loses $0.13$, Food101, MetMeme, and CrisisMMD lose $0.04$ to $0.05$, and Fracture and Memotion change by less than $0.01$. Inline-1 \mbox{is the configuration used throughout the main benchmark.}

\subsection{Agent Prior-Knowledge Controls}
\label{app:agent-controls}

To test whether \Agent's accuracy reflects the multimodal readouts or the frozen LLM's parametric prior, we report five controls under the deployed Qwen3-32B Inline-1 setting at the $100\%$ training budget. \textbf{LLM-only} removes both readouts. \textbf{Shuffled-tool} preserves the readout format but supplies another sample's readouts (a seeded permutation of the sample-to-readout map). \textbf{Anonymised labels} replaces every class name or MCQ option the LLM sees, in prompts and readouts, by a semantically empty token (\texttt{class\_i}, \texttt{option\_j}). \textbf{No demonstrations (classification) / No question (QA)} denotes task-specific interventions. For classification, No demonstrations removes the in-context example texts and labels while retaining the current sample text, class names and descriptions, and both readout outputs. On QA tasks, the control deletes only the question body, retaining answer options, instructions, demonstrations, and all other prompt content. All six conditions use the complete test split and the same five-seed aggregation as the corresponding full-budget entries in Tables~\ref{tab:agent_fm_cls_reg} and~\ref{tab:agent_fm_qa}. Anonymisation retains the question and other sample text.

\begin{table}[t]
\centering
\footnotesize
\setlength{\tabcolsep}{4pt}
\caption{\textbf{Agent controls} (accuracy). All six conditions use Qwen3-32B Inline-1 at full training budget, each dataset's complete test split (Table~\ref{tab:dataset_full}), and the same five-seed aggregation. \textbf{Deployed} reproduces the corresponding main-table \Agent\ results. The other columns apply the indicated input interventions. LLM-only omits both readouts. Anonymisation replaces label/option names while retaining the question or other raw sample text. For classification, No demonstrations removes the in-context example texts and labels while retaining the current sample text, class names and descriptions, and both readout outputs. These classification entries reproduce the $100\%$-budget $N_{\mathrm{ICL}}{=}0$ results in Table~\ref{tab:kshot_cls}. For QA, No question deletes only the question body. Answer options, instructions, demonstrations, and other prompt content remain unchanged. For classification, LLM+readouts removes the current sample text, class names and descriptions, and in-context example texts and labels. Readout scores and nonsemantic candidate identifiers remain available to the LLM. The contrast measures joint removal of direct sample text and label semantics. For QA, LLM+readouts retains the controller and readout outputs but removes direct raw sample text, including the question and answer-option contents. Candidate identifiers remain available for returning an answer. This is distinct from removing only the question \mbox{body and from probability averaging without an LLM.}}
\label{tab:agent-controls}
\adjustbox{max width=\textwidth}{%
\begin{tabular}{@{}l l c c c c c c@{}}
\toprule
\textbf{Task} & \textbf{Dataset} & \textbf{Deployed} & \textbf{LLM-only} & \textbf{Shuffled-tool} & \textbf{Anon.\ labels} & \shortstack{\textbf{No demos (CLS)}\\\textbf{No question (QA)}} & \textbf{LLM+readouts} \\
\midrule
\multirow{5}{*}{CLS} & CrisisMMD & 0.683 & 0.580 & 0.492 & 0.532 & 0.634 & 0.325 \\
 & CUB       & 0.882 & 0.033 & 0.025 & 0.600 & 0.872 & 0.458 \\
 & Fracture  & 0.136 & 0.096 & 0.081 & 0.125 & 0.159 & 0.082 \\
 & Memotion  & 0.219 & 0.238 & 0.202 & 0.245 & 0.213 & 0.202 \\
 & OxfordPet & 0.945 & 0.203 & 0.224 & 0.632 & 0.954 & 0.195 \\
\midrule
\multirow{3}{*}{QA} & AI2D      & 0.778 & \textbf{0.830} & 0.660 & 0.460 & 0.602 & 0.325 \\
 & ScienceQA & 0.679 & 0.665 & 0.600 & 0.765 & 0.545 & 0.372 \\
 & VQA-RAD   & \textbf{0.375} & 0.300 & 0.300 & 0.235 & 0.205 & 0.185 \\
\bottomrule
\end{tabular}}
\end{table}

Relative to LLM-only, the full system changes mean accuracy by $-0.052$ on AI2D, $+0.014$ on ScienceQA, and $+0.075$ on VQA-RAD. It exceeds shuffled-readout performance by $0.118$, $0.079$, and $0.075$, respectively. Thus, sensitivity to readout content can coexist with either positive or negative gains from supplying it. These are five-seed mean contrasts on matched test examples, with uncertainty reporting described in Appendix~\ref{app:reporting-scope}.

On CUB and OxfordPet, anonymised labels retain $0.600$ and $0.632$, compared with $0.033$ and $0.203$ for LLM-only and $0.025$ and $0.224$ for shuffled readouts. This indicates useful readout evidence even when label names are unavailable. For MCQ, anonymising options removes their semantic content but retains the question. The No demonstrations / No question column keeps the LLM and readouts. Classification removes only the in-context examples, retaining current sample text and class semantics. QA removes only the question body, retaining all other prompt content. The classification entries reproduce the $100\%$-budget $N_{\mathrm{ICL}}{=}0$ results in Table~\ref{tab:kshot_cls}. Accuracy is $0.634$, $0.872$, $0.159$, $0.213$, and $0.954$ on CrisisMMD, CUB, Fracture, Memotion, and OxfordPet, and $0.602$, $0.545$, and $0.205$ on AI2D, ScienceQA, and VQA-RAD. The full system exceeds this control by $17.6$, $13.4$, and $17.0$ percentage points on the three QA datasets. Removing demonstrations lowers accuracy on CrisisMMD, CUB, and Memotion, but raises it on Fracture ($0.136$ to $0.159$) and OxfordPet ($0.945$ to $0.954$). Appendix~\ref{app:reporting-scope} specifies the inputs retained by each intervention.

For classification, LLM+readouts removes the current sample text, class names and descriptions, and in-context example texts and labels. Readout scores and nonsemantic candidate identifiers remain available to the LLM. The contrast measures the joint removal of direct sample text and label semantics. For QA, the LLM-plus-readouts control retains the controller and readout outputs while removing direct raw sample text, including the question and answer-option contents. Candidate identifiers support answer selection. On the same test examples, this column reports $0.325$, $0.458$, $0.082$, $0.202$, $0.195$, $0.325$, $0.372$, and $0.185$ in table order. On AI2D, ScienceQA, and VQA-RAD, these accuracies are respectively $27.7$, $17.3$, and $2.0$ percentage points below the question-body-ablation column. The two columns distinguish question removal from removal of all direct text content.

\paragraph{Score averaging.}
Table~\ref{tab:agent-score-avg} compares two interventions: its upper panel removes the LLM on the Agent-winning settings of CrisisMMD, Food101, MELD, and MetMeme, while its lower panel removes the readouts on Food101 at three budgets. Averaging the class probabilities of the two readouts reaches roughly the stronger readout, which is near or below \Decoupled, and stays $9.6$ to $24.9$ points below \Agent\ in each reported setting. The text readout is below \Agent\ on all four datasets, whereas the LLM reads the raw text of the sample. The full system combines raw-text access and LLM-mediated use of readouts, and outperforms probability averaging in these comparisons. For Food101, averaging reaches $0.695$, $0.737$, and $0.810$ at $20$-shot, $10\%$, and full budget, versus $0.872$, $0.880$, and $0.906$ for \Agent. The listed single-modality probes in Table~\ref{tab:text_tower_ablation} also remain below \Agent. LLM-only reaches $0.843$, $0.852$, and $0.898$, giving full-system-minus-LLM-only gains of $0.029$, $0.028$, and $0.008$. These conditional input contrasts retain the LLM and raw-text access, and extend the readout intervention to both smaller and full training budgets. LLM-only, the listed probes and readouts, and probability averaging do not overturn the three reported \Agent\ wins on Food101. The small full-budget mean gain qualifies its practical magnitude. Uncertainty reporting follows Appendix~\ref{app:reporting-scope}.

\begin{table}[t]
\centering
\footnotesize
\setlength{\tabcolsep}{4pt}
\caption{\textbf{Simple readout and LLM-only comparisons with \Agent} on CrisisMMD ($20$-shot and $100\%$), Food101 (all three budgets), MELD ($20$-shot), and MetMeme ($20$-shot), where \Agent\ wins in Table~\ref{tab:agent_fm_cls_reg} (accuracy). \textbf{Non-text} and \textbf{Text} are the two single-modality readouts \Agent\ receives, and \textbf{Avg.} averages their class probabilities with no trained parameter. Readouts and average are the mean over five label-subsampling seeds, on the same complete test split as \Agent. Food101 stores the ten highest logits of each readout, and the remaining classes take the tenth. The lower panel compares Food101 with LLM-only, which retains raw-text access and omits both readouts as \mbox{in Table~\ref{tab:agent-controls}.} Differences are five-seed mean contrasts in percentage points.}
\label{tab:agent-score-avg}
\begin{tabular}{@{}l l c c c c c c@{}}
\toprule
\textbf{Dataset} & \textbf{Budget} & \textbf{Non-text} & \textbf{Text} & \textbf{Avg.} & \textbf{\Agent} & \textbf{\Joint} & \textbf{\Decoupled} \\
\midrule
CrisisMMD & 20-shot & 0.235 & 0.090 & 0.221 & \textbf{0.460} & 0.458 & 0.306 \\
CrisisMMD & 100\%   & 0.500 & 0.467 & 0.510 & \textbf{0.683} & 0.668 & 0.538 \\
Food101   & 20-shot & 0.695 & 0.074 & 0.695 & \textbf{0.872} & 0.767 & 0.695 \\
Food101   & 10\%    & 0.736 & 0.074 & 0.737 & \textbf{0.880} & 0.804 & 0.733 \\
Food101   & 100\%   & 0.802 & 0.122 & 0.810 & \textbf{0.906} & 0.862 & 0.808 \\
MELD      & 20-shot & 0.309 & 0.281 & 0.342 & \textbf{0.591} & 0.375 & 0.445 \\
MetMeme   & 20-shot & 0.184 & 0.178 & 0.187 & \textbf{0.290} & 0.252 & 0.208 \\
\bottomrule
\end{tabular}
\par\medskip
\begin{tabular}{@{}l c c c@{}}
\toprule
\multicolumn{4}{c}{\textbf{Food101: removing readouts while retaining the LLM}} \\
\textbf{Budget} & \textbf{Full \Agent} & \textbf{LLM-only} & \textbf{Full $-$ LLM-only (pp)} \\
\midrule
20-shot & 0.872 & 0.843 & $+2.9$ \\
10\% & 0.880 & 0.852 & $+2.8$ \\
100\% & 0.906 & 0.898 & $+0.8$ \\
\bottomrule
\end{tabular}
\end{table}

\paragraph{Same-size controller.}
With SmolLM2-135M-Instruct as controller, at or below the scale of the encoder towers, and the Inline-1 protocol unchanged, \Agent\ never outperforms the better fusion paradigm in the evaluated classification and QA settings, whereas Qwen3-8B does so in some of these settings. Its mean parse-failure rate over these evaluations is near $0.9$, so the $135$M model does not follow the answer protocol.

\subsection{Grouped Routing Comparison}
\label{app:baselines}
The reference TB-Majority assigns the majority winner among outer-training cells sharing task type and budget type. Direct classification and the cascade instead use the diagnostic and metadata features in Appendix~\ref{app:features}. All three are evaluated on the same $144$ cells in $15$ groups, with DrugBank-Desc and DrugBank-Pharma held out together. \mbox{Results average five labelled-set feature runs.}

\paragraph{Additional routing classifiers.}
\label{app:additional-routing-classifiers}
Table~\ref{tab:routing-additional-classifiers} compares six additional configurations: logistic regression with stronger regularization ($C=0.1$), shrinkage LDA, $5$-NN, an RBF-kernel SVM, naive Bayes with missing-value support, and depth-$2$ histogram gradient boosting. Logistic regression achieves the highest accuracy ($0.571$) and lowest relative regret ($0.086$) among these alternatives. The table's intervals concern accuracy differences from TB-Majority and are available for logistic regression and gradient boosting, both spanning zero. Cascade--classifier intervals were not computed.

\begin{table}[t]
\centering\footnotesize
\setlength{\tabcolsep}{4pt}
\caption{\textbf{Additional routing classifiers.} Accuracy, mean number of correctly selected cells, and relative regret. The final column gives accuracy differences from TB-Majority, with $95\%$ intervals where available. Bold marks the best accuracy and regret among these six classifiers.}
\label{tab:routing-additional-classifiers}
\begin{adjustbox}{max width=\textwidth}
\begin{tabular}{@{}lrrrr@{}}
\toprule
Classifier & Accuracy$\uparrow$ & Correct cells & Relative regret$\downarrow$ & $\Delta$Acc. vs.\ TB-Majority \\
\midrule
Logistic ($C=0.1$) & \textbf{0.571} & 82.2 & \textbf{0.086} & $+0.085\;[-0.12,0.30]$ \\
LDA (shrinkage covariance) & 0.443 & 63.8 & 0.157 & $-0.043$ \\
kNN ($k=5$) & 0.436 & 62.8 & 0.104 & $-0.050$ \\
SVM (RBF kernel) & 0.432 & 62.2 & 0.092 & $-0.054$ \\
Naive Bayes (missing-value support) & 0.414 & 59.6 & 0.127 & $-0.072$ \\
HGB (depth 2) & 0.365 & 52.6 & 0.127 & $-0.121\;[-0.32,0.09]$ \\
\bottomrule
\end{tabular}
\end{adjustbox}
\end{table}

\paragraph{Metadata-only reference.}
The metadata-only baseline retains the cascade protocol and removes only diagnostic features (Appendix~\ref{app:metadata-only}). At $\epsilon=0$, it reports $0.542$ accuracy, $78.0$ correctly selected cells averaged over five runs, and $0.0566$ mean relative regret. Relative to this baseline, the full cascade selects $24.8$ more cells correctly on average and lowers mean regret by $0.0283$, approximately $50\%$. Table~\ref{tab:router-per-dataset} provides the per-dataset comparison. At the displayed precision, the cascade has lower regret on eight datasets, equal regret on four, and higher regret on four (CUB, KIBA, MELD, and PDBbind). Grouping the two DrugBank tasks together gives eight improvements, three ties, and four regressions over the $15$ evaluation groups. These are descriptive comparisons of mean regret at zero tolerance. Table~\ref{tab:router-comparison} reports all five tolerances.

\paragraph{Diagnostic feature-group ablations.}
Table~\ref{tab:routing-feature-ablation} complements Full versus metadata-only with single-group additions and removals. The text diagnostic is $\Delta_{\mathrm{zs}}$. None of the single-group additions recovers Full's accuracy or regret. The largest accuracy reductions occur when supervised representation diagnostics or the text diagnostic is removed. Geometry removal produces a smaller mean accuracy reduction, whose interval includes zero, but increases regret. The accuracy interval for text-diagnostic removal also includes zero, whereas the regret intervals for all three removals favour Full. These paired comparisons describe the contribution of each group in the presence of the remaining features.

\begin{table}[t]
\centering\footnotesize
\setlength{\tabcolsep}{3pt}
\caption{\textbf{Routing feature-group ablations at $\epsilon=0$.} Accuracy and relative regret cover all $144$ dataset--budget combinations and average five runs with the text-probe sampling protocol of Appendix~\ref{app:zs-feature}; class identities remain fixed for CUB and OxfordPet. SRD denotes the five supervised representation diagnostics (two ANOVA and three $5$-NN features). ``Metadata +'' adds one diagnostic group to metadata-only routing. ``Full minus'' removes one group from the complete diagnostic set. Differences are Full minus the indicated ablation, with reported paired $95\%$ intervals. Positive accuracy differences and negative regret differences favour Full. Displayed means and differences are rounded independently.}
\label{tab:routing-feature-ablation}
\begin{adjustbox}{max width=\textwidth}
\begin{tabular}{@{}lrrrr@{}}
\toprule
Features & Accuracy & Regret & $\Delta$ accuracy [95\% CI] & $\Delta$ regret [95\% CI] \\
\midrule
Full & 0.714 & 0.028 & --- & --- \\
Metadata only & 0.542 & 0.057 & --- & --- \\
\midrule
Metadata + geometry & 0.549 & 0.113 & $+0.165\;[-0.022,0.340]$ & $-0.084\;[-0.183,-0.016]$ \\
Metadata + SRD & 0.542 & 0.073 & $+0.172\;[0.041,0.304]$ & $-0.045\;[-0.083,-0.012]$ \\
Metadata + text diagnostic & 0.535 & 0.062 & $+0.179\;[0.025,0.316]$ & $-0.034\;[-0.056,-0.013]$ \\
\midrule
Full minus geometry & 0.671 & 0.041 & $+0.043\;[-0.040,0.106]$ & $-0.013\;[-0.028,-0.001]$ \\
Full minus SRD & 0.544 & 0.108 & $+0.170\;[0.004,0.343]$ & $-0.080\;[-0.180,-0.012]$ \\
Full minus text diagnostic & 0.529 & 0.105 & $+0.185\;[-0.001,0.366]$ & $-0.077\;[-0.174,-0.014]$ \\
\bottomrule
\end{tabular}
\end{adjustbox}
\end{table}

\paragraph{Sensitivity to individual evaluation groups.}
\label{app:metadata-group-sensitivity}
For the Full-minus-metadata-only comparison at $\epsilon=0$, paired dataset-group bootstrap resampling draws $15$ groups with replacement, keeping both DrugBank tasks in one group and all budget cells and five seed rows together. Each draw uses the same groups for both methods and computes the cell-weighted difference. We use $4{,}000$ draws with NumPy \texttt{default\_rng(0)} and report the $2.5$th and $97.5$th percentiles. Accuracy is $0.713889$ versus $0.541667$, giving $+0.172222$ with interval $[0.001449,0.345991]$. Relative regret is $0.028338$ versus $0.056601$, giving $-0.028263$ with interval $[-0.054377,-0.002035]$. These group-level intervals condition on the fitted held-out predictions and candidate utilities.

At $\epsilon=0$, we hold the existing five-run outer-held-out predictions fixed and omit each of the $15$ evaluation groups in turn from the aggregate, treating the two DrugBank tasks as one group. We recompute cell-weighted Cascade-minus-metadata-only differences on the remaining $14$ groups. For group $g$ with $n_g$ cells and mean difference $\delta_g$, the omitted-group difference is $\sum_{h\ne g}n_h\delta_h/\sum_{h\ne g}n_h$. Accuracy differences use correct-count increments implied by five runs and the displayed cell counts, consistent with the reported totals $102.8$ and $78.0$.

The full-set accuracy difference is $24.8/144=0.1722$. After omitting any one group, accuracy differences range from $+0.117$ to $+0.204$, and relative-regret differences range from $-0.036$ to $-0.020$. Omitting VQA-RAD gives the smallest accuracy advantage, $(24.8-9)/(144-9)=0.1170$, while omitting KIBA gives the largest, $(24.8+2.6)/(144-10)=0.2045$. The most negative regret difference occurs without MELD, and the least negative without VQA-RAD. Across individual groups, accuracy improves/ties/worsens in $10/3/2$ groups. Regret improves/ties/worsens in $8/3/4$. Thus the direction of the aggregate advantage does not depend on any single evaluation group, although gains vary across tasks.

\subsection{Utility-Based Routing Comparison}
\label{app:utility-routing}
Tables~\ref{tab:router-utility} and~\ref{tab:cascade-prediction} complement the mean results in Table~\ref{tab:router-comparison} with paired differences and pointwise bootstrap intervals. At the reference tolerance, all four intervals exclude zero in favour of the cascade. The lower bound against TB-Majority is close to zero. At the other four tolerances, accuracy intervals against TB-Majority cross zero, indicating uncertainty in the direction of the accuracy difference at those tolerances. The comparison evaluates each complete predictor, including its feature access and decision rule.

Across the five tie bands, mean accuracy/regret is $0.678/0.045$ for the cascade, $0.498/0.088$ for direct classification, and $0.503/0.127$ for TB-Majority, computed from the displayed summaries. Direct classification has lower accuracy than TB-Majority at the first three bands and higher accuracy at the last two, while its regret is lower throughout. The cascade has lower mean regret than direct classification at every reported tolerance. Larger tolerances do not necessarily improve prediction or utility.

\begin{table}[t]
\centering\footnotesize
\caption{\textbf{Stage-level and final routing accuracy.} Results use $\epsilon=0$, $15$ dataset groups, and averages over five labelled-set seeds. Final accuracy covers all $144$ dataset--budget combinations. For all three configurations, Step~2 accuracy is computed on the same subset whose true winner is \Joint\ or \Decoupled. The feature-encoder variant replaces the general-domain feature encoders with the base-tier configuration.}
\label{tab:routing-stage-accuracy}
\begin{tabular}{@{}lccc@{}}
\toprule
Configuration & Step 1 ACC & Step 2 ACC & Final ACC \\
\midrule
Cascade, default features & 0.806 & 0.843 & 0.714 \\
Direct, default features & 0.654 & 0.627 & 0.450 \\
Cascade, base-tier general-domain features & 0.804 & 0.867 & 0.718 \\
\bottomrule
\end{tabular}
\end{table}

\begin{table}[t]
\centering\footnotesize
\caption{\textbf{Cascade minus direct three-way classifier.} Positive accuracy differences and negative relative-regret differences favour the cascade. Differences use higher-precision summaries. Reference-tolerance accuracy is $102.8/144$. Values are rounded independently and may differ by $0.001$ from subtraction of displayed means. Intervals are pointwise paired group-bootstrap percentiles ($15$ groups, $4{,}000$ draws, NumPy \texttt{default\_rng(0)}). The $1\%$ regret difference uses displayed means. Its interval is not reported.}
\label{tab:router-utility}
\begin{tabular}{@{}c cc@{}}\toprule
$\epsilon$ & Accuracy difference [95\% CI] & Relative-regret difference [95\% CI] \\\midrule
0\% & $+0.264\;[0.139,0.399]$ & $-0.063\;[-0.113,-0.024]$ \\
0.2\% & $+0.265\;[0.099,0.452]$ & $-0.072\;[-0.164,-0.016]$ \\
0.5\% & $+0.267\;[0.101,0.452]$ & $-0.066\;[-0.159,-0.011]$ \\
1\% & $+0.024\;[-0.206,0.184]$ & $-0.008\;[\text{--}]$ \\
2\% & $+0.078\;[-0.009,0.149]$ & $-0.009\;[-0.024,0.002]$ \\
\bottomrule\end{tabular}\end{table}

\begin{table}[t]
\centering\footnotesize
\caption{\textbf{Cascade minus TB-Majority.} Positive accuracy differences and negative relative-regret differences favour the cascade. Differences use higher-precision summaries. Reference-tolerance accuracy is $102.8/144$. Values are rounded independently and may differ by $0.001$ from subtraction of displayed means. Intervals are pointwise paired group-bootstrap percentiles ($15$ groups, $4{,}000$ draws, NumPy \texttt{default\_rng(0)}). The $1\%$ regret difference uses displayed means. Its interval is not reported.}
\label{tab:cascade-prediction}
\begin{tabular}{@{}c cc@{}}\toprule
$\epsilon$ & Accuracy difference [95\% CI] & Relative-regret difference [95\% CI] \\\midrule
0\% & $+0.228\;[0.003,0.457]$ & $-0.099\;[-0.199,-0.021]$ \\
0.2\% & $+0.197\;[-0.044,0.448]$ & $-0.091\;[-0.193,-0.010]$ \\
0.5\% & $+0.201\;[-0.031,0.445]$ & $-0.089\;[-0.191,-0.009]$ \\
1\% & $+0.116\;[-0.161,0.348]$ & $-0.084\;[\text{--}]$ \\
2\% & $+0.133\;[-0.072,0.311]$ & $-0.050\;[-0.106,-0.000812]$ \\
\bottomrule\end{tabular}\end{table}

\subsection{Per-Dataset Routing Results}
\label{app:per-dataset-routing}
Table~\ref{tab:router-per-dataset} reports all $16$ tasks at the reference tolerance. The cascade attains perfect winner accuracy on AI2D, both DrugBank tasks, and VQA-RAD, but its accuracy is only $0.275$ on CUB, $0.280$ on ScienceQA, and $0.320$ on Memotion. Every dataset has nonzero accuracy after averaging the five runs. Relative regret is largest on MELD ($0.1189$), Memotion ($0.1111$), and ScienceQA ($0.1107$). TB-Majority is better on both metrics for CUB, MELD, and KIBA, has lower regret but lower accuracy on IEMOCAP, and higher accuracy but higher regret on Memotion. The aggregate advantage thus coexists with task-specific reversals. On ScienceQA, the displayed regret is identical for all three methods despite different winner accuracy, further illustrating that these metrics capture different aspects of selection.
\begin{table}[t]
\centering\footnotesize
\setlength{\tabcolsep}{4pt}
\caption{\textbf{Per-dataset routing at $\epsilon=0$.} Each method reports winner accuracy / mean relative regret over the dataset's budget cells and five labelled-set seeds. The aggregate weights each dataset by its number of budget settings. DrugBank tasks are reported separately but are held out together in outer and inner evaluation. Displayed dataset scores are rounded. Aggregate accuracies correspond to $102.8$ and $78.0$ correctly selected cells for Cascade and Metadata-only, respectively, averaged over five seeds.}
\label{tab:router-per-dataset}
\begin{adjustbox}{max width=\textwidth}\begin{tabular}{@{}lrcccc@{}}\toprule
Dataset & Cells & Cascade & Direct & TB-Majority & Metadata-only \\\midrule
AI2D & 5 & 1.000 / 0.0000 & 1.000 / 0.0000 & 0.000 / 0.3679 & 1.000 / 0.0000 \\
CrisisMMD & 10 & 0.580 / 0.0108 & 0.540 / 0.0074 & 0.300 / 0.1179 & 0.300 / 0.0942 \\
CUB & 8 & 0.275 / 0.0284 & 0.250 / 0.0299 & 0.875 / 0.0053 & 0.625 / 0.0123 \\
DrugBank-Desc & 5 & 1.000 / 0.0000 & 0.000 / 0.6553 & 0.000 / 0.6553 & 1.000 / 0.0000 \\
DrugBank-Pharma & 5 & 1.000 / 0.0000 & 1.000 / 0.0000 & 0.000 / 0.5507 & 1.000 / 0.0000 \\
Fracture & 14 & 0.714 / 0.0063 & 0.286 / 0.0666 & 0.143 / 0.2447 & 0.357 / 0.0609 \\
IEMOCAP & 10 & 0.840 / 0.0042 & 0.700 / 0.0130 & 0.800 / 0.0023 & 0.300 / 0.0403 \\
KIBA & 10 & 0.640 / 0.0103 & 0.100 / 0.1213 & 0.900 / 0.0001 & 0.900 / 0.0001 \\
MELD & 10 & 0.640 / 0.1189 & 0.580 / 0.1582 & 1.000 / 0.0000 & 0.600 / 0.0414 \\
Memotion & 10 & 0.320 / 0.1111 & 0.000 / 0.2426 & 0.500 / 0.2035 & 0.000 / 0.2301 \\
MetMeme & 10 & 0.660 / 0.0360 & 0.360 / 0.0985 & 0.400 / 0.1010 & 0.600 / 0.0500 \\
OxfordPet & 13 & 0.923 / 0.0000 & 0.923 / 0.0000 & 0.846 / 0.0067 & 0.923 / 0.0025 \\
PDBbind & 10 & 0.820 / 0.0018 & 0.740 / 0.0037 & 0.500 / 0.0084 & 0.800 / 0.0014 \\
ScienceQA & 5 & 0.280 / 0.1107 & 0.200 / 0.1107 & 0.200 / 0.1107 & 0.200 / 0.1107 \\
Food101 & 10 & 0.820 / 0.0280 & 0.560 / 0.0331 & 0.500 / 0.0703 & 0.500 / 0.0703 \\
VQA-RAD & 9 & 1.000 / 0.0000 & 0.000 / 0.1484 & 0.000 / 0.1484 & 0.000 / 0.1484 \\
\midrule
Cell-weighted aggregate & 144 & 0.714 / 0.0283 & 0.450 / 0.0911 & 0.486 / 0.1274 & 0.542 / 0.0566 \\
\bottomrule\end{tabular}\end{adjustbox}
\end{table}

\paragraph{Stage-specific failure patterns.}
The fold-level diagnostic inspection identifies different failure modes on CUB, KIBA, and MELD. On CUB, errors arise at the Agent/Fusion decision: the router continues to favor \Agent\ at larger budgets where \Decoupled\ is preferred. The text diagnostic $\Delta_{\mathrm{zs}}$ remains positive across budgets (approximately $0.04$--$0.19$), but this proxy is not a measured Agent--fusion utility difference. The five-run aggregate relative regret is $0.0284$, illustrating that frequent winner errors can have a modest average utility cost. On KIBA, the Agent/Fusion decision correctly rejects \Agent. The reported errors instead concern Joint/Decoupled choices in the k-shot regime. Small fusion-utility differences in most of these cells limit the aggregate regret to $0.0103$.

On MELD, the diagnostic inspection identifies costly k-shot errors in which the router selects \Decoupled\ despite a substantial \Agent\ advantage. Its text diagnostic is near zero in the reported cells (approximately $-0.04$ to $0.13$). A possible explanation is a mismatch between text--label matching in the fixed BGE probe and the controller's ability to interpret transcripts. Under the reported five-run aggregation, the full router has slightly higher accuracy than metadata-only ($0.640$ versus $0.600$), but higher regret ($0.1189$ versus $0.0414$). Its mistakes therefore incur greater utility loss on average.

In the inspected CUB, KIBA, and MELD folds, inner selection excludes the explicit budget block from both stages of the full router, while the metadata-only reference retains budget descriptors. The full router consequently lacks a direct budget covariate in these folds, although its diagnostic features can still vary with the labelled subset and thus indirectly convey budget information. This selection pattern is consistent with difficulty tracking budget-dependent winner changes. Together, these cases distinguish errors at the Agent/Fusion boundary from errors between fusion systems, and show why similar winner accuracy can conceal very different utility losses.

\subsection{Regression Output Parsing}
\label{app:regression-parsing}
Table~\ref{tab:regression-parse} reports default-AGENT parsing failures. Rates are $0.1$--$0.2\%$ on IEMOCAP, $0.3$--$2.9\%$ on KIBA, and $2.8$--$3.3\%$ on PDBbind. Under the zero-substitution rule, the reported MAE includes both prediction error and output-format failures. The MAE contribution of each parsing failure depends on the distance between zero and the target value. These are end-to-end system results. Conditional MAE on parsed \mbox{outputs cannot be reconstructed from failure percentages alone.}
\begin{table}[t]
\centering\footnotesize
\caption{\textbf{Unparseable outputs of the default AGENT on regression tasks.} Entries are the observed failure percentages at the three main-table budgets. Unparseable outputs are replaced by numerical zero before MAE evaluation. These rates do not quantify the corresponding MAE contribution \mbox{or the conditional error on successfully parsed outputs.}}
\label{tab:regression-parse}
\begin{tabular}{@{}l r r r@{}}
\toprule
Dataset & $20$-shot & $10\%$ & $100\%$ \\
\midrule
PDBbind & $3.3\%$ & $2.8\%$ & $2.8\%$ \\
KIBA & $0.3\%$ & $1.0\%$ & $2.9\%$ \\
IEMOCAP & $0.2\%$ & $0.1\%$ & $0.2\%$ \\
\bottomrule\end{tabular}
\end{table}

\subsection{Sensitivity of the Descriptive Joint--Decoupled Association}
\label{app:jd-descriptive}
Figure~\ref{fig:jd-conditions}A counts Joint--Decoupled preferences across every budget of each QA dataset, regardless of Agent\textquotesingle s rank. Here $n$ denotes the number of budgets, not independent datasets.

Here $u$ is the raw \texttt{knn5\_mod\_max} diagnostic: the larger of the two modalities' five-seed mean leave-one-out kNN-5 accuracies (Eq.~\eqref{eq:feat-knn5}), before any size residualization. We first average within each modality, then take the maximum. The margin uses direction-adjusted utilities from Eq.~\eqref{eq:utility}:
\[
\begin{aligned}
 u(d,K)&=\max_{m\in\{a,b\}}\left\{\frac15\sum_{s=0}^4 a_{\mathrm{kNN5},m}^{(s)}(d,K)\right\},
 \\
 m_J(d,K)&=\frac{U(\Joint;d,K)-U(\Decoupled;d,K)}{|U^\star(d,K)|}.
\end{aligned}
\]
where $U^\star=\max\{U(\Agent),U(\Joint),U(\Decoupled)\}$. Thus the denominator includes Agent: it scales the Joint--Decoupled difference by the best of all three configurations. It does not change which fusion configuration wins, but can affect cross-dataset margin magnitudes and ranks. The plotted margin averages cell-wise ratios over budgets.

For Figure~\ref{fig:jd-conditions}B, each of the eight classification datasets contributes one point, with equal weight. We average $u$ and the Joint--Decoupled margin over identical cells with available $u$: $84$ of $85$ classification cells. Memotion's one-shot cell lacks $u$ and is excluded from both means. The matched-cell Spearman correlation is $-0.667$. The leave-one-dataset-out sensitivity range is $[-0.750,-0.571]$, obtained by recomputing the descriptive correlation on each remaining set of seven datasets. If the margin mean instead includes every budget cell, the correlation becomes $-0.738$. Only Memotion changes, reversing its margin rank relative to Food101. Using chance-normalized separability $(u-1/C)/(1-1/C)$ with matched cells gives $-0.619$, for class counts $C$ of $7$, $10$, $5$, $8$, $7$, $101$, $200$, and $37$ on MetMeme, Fracture, Memotion, CrisisMMD, MELD, Food101, CUB, and OxfordPet, respectively. Across these checks, the association remains negative, while its magnitude varies with the ranks of the eight datasets.

\clearpage
\section{Datasets and Encoder Matching}
\label{app:datasets}

Section~\ref{app:dataset-overview} inventories the datasets. Sections~\ref{app:matching} and~\ref{app:encoder-families} define matching and list encoder families. Section~\ref{app:preprocessing} specifies preprocessing and splits, and Section~\ref{app:domain-ablation} describes the domain-pretraining comparison.

\subsection{Dataset Overview}
\label{app:dataset-overview}

We study 16 datasets spanning five domains (general vision--language, audio, medical, molecule, protein) and three task types. Table~\ref{tab:dataset_full} gives the complete inventory with modality pairs, task type, evaluation metric, and train, development, and test sizes. Table~\ref{tab:label-counts} \mbox{gives the labelled-set size of every budget cell.}

\begin{table}[t]
\centering
\footnotesize
\setlength{\tabcolsep}{4pt}
\caption{\textbf{Complete dataset inventory: $16$ datasets in five domains} ($8$ classification, $3$ regression, $5$ QA). The three reported budgets are 20-shot, 10\%, and 100\% for CLS/REG, and 10\%, 50\%, and 100\% for QA. Train is the pool used for labelled-set sampling. Test size is the number of examples scored: all systems use the complete test split, fixed across budgets and seeds. Dev is disjoint from the training pool and test split. For datasets with a training holdout, Train is the remaining pool after removing Dev. $100\%$ uses that pool only. Development labels are additional to \mbox{the reported training budget (Appendix~\ref{app:datasets}).} The eight control datasets use these same complete test splits (Table~\ref{tab:agent-controls}). Dataset-level timing estimates and their coverage are listed in Table~\ref{tab:cost}, panel B.}
\label{tab:dataset_full}
\begin{adjustbox}{max width=\textwidth,center}
\begin{tabular}{llllrrrl}
\toprule
\textbf{Dataset} & \textbf{Modality A} & \textbf{Modality B} & \textbf{Task} & \textbf{Train Size} & \textbf{Dev Size} & \textbf{Test Size} & \textbf{Metric} \\
\midrule
\rowcolor{gray!10}\multicolumn{8}{c}{\textbf{\textit{General vision--language (8 datasets)}}} \\
\midrule
CUB-200-2011 & image & caption & classification & 8,244 & 1,777 & 1,767 & accuracy ($\uparrow$) \\
CrisisMMD & image & tweet text & classification & 13,608 & 2,237 & 2,237 & accuracy ($\uparrow$) \\
Food101 & image & web-page text & classification & 58,106 & 6,403 & 21,504 & accuracy ($\uparrow$) \\
Memotion & image & OCR text & classification & 5,593 & 699 & 700 & accuracy ($\uparrow$) \\
MetMeme & image & meme text (English) & classification & 2,394 & 798 & 800 & accuracy ($\uparrow$) \\
OxfordPet & image & generated caption & classification & 2,944 & 736 & 3,669 & accuracy ($\uparrow$) \\
AI2D & diagram image & question & MCQ & 5,969 & 746 & 747 & accuracy ($\uparrow$) \\
ScienceQA & image & question & MCQ & 6,218 & 2,097 & 2,017 & accuracy ($\uparrow$) \\
\midrule
\rowcolor{gray!10}\multicolumn{8}{c}{\textbf{\textit{Audio (2 datasets)}}} \\
\midrule
MELD & audio & transcript & classification & 9,988 & 961 & 2,610 & accuracy ($\uparrow$) \\
IEMOCAP & audio & transcript & regression (valence) & 6,425 & 1,606 & 2,008 & MAE ($\downarrow$) \\
\midrule
\rowcolor{gray!10}\multicolumn{8}{c}{\textbf{\textit{Medical (2 datasets)}}} \\
\midrule
Fracture & X-ray image & patient complaint & classification & 790 & 113 & 226 & accuracy ($\uparrow$) \\
VQA-RAD & radiology image & question & VQA & 1,435 & 358 & 451 & accuracy ($\uparrow$) \\
\midrule
\rowcolor{gray!10}\multicolumn{8}{c}{\textbf{\textit{Molecule (2 datasets)}}} \\
\midrule
DrugBank-Desc & SMILES & description & MCQ & 923 & 115 & 116 & accuracy ($\uparrow$) \\
DrugBank-Pharma & SMILES & pharmacodynamics & MCQ & 804 & 100 & 101 & accuracy ($\uparrow$) \\
\midrule
\rowcolor{gray!10}\multicolumn{8}{c}{\textbf{\textit{Protein (2 datasets)}}} \\
\midrule
KIBA & protein sequence & ligand SMILES & regression & 94,077 & 5,000 & 11,761 & MAE ($\downarrow$) \\
PDBbind & protein sequence & ligand SMILES & regression & 10,903 & 2,370 & 5,218 & MAE ($\downarrow$) \\
\bottomrule
\end{tabular}
\end{adjustbox}
\end{table}

\subsection{Encoder Matching Protocol}
\label{app:matching}

We match encoder towers by architecture and scale according to the following definition.

\begin{definition}[Encoder matching]
\label{def:matching}
For dataset $d$, let $F_J$ be the joint FM with towers $(F_J^a, F_J^b)$ and let $(E_a, E_b)$ be the single-modality encoders used by \Decoupled and, as readouts, by \Agent. The triple $(F_J, E_a, E_b)$ is \emph{matched} if:
\begin{enumerate}[label=(\roman*),leftmargin=2em]
    \item $E_a$ has the same architecture family, depth, width, and parameter scale as $F_J^a$, and the same patch size wherever a patch-matched checkpoint is released at that architecture, \mbox{and likewise $E_b$ for $F_J^b$.}
    \item $E_a$ and $E_b$ were never aligned to each other: no pretraining stage trained the \mbox{pair $(E_a, E_b)$ jointly.}
    \item $F_J$, $E_a$, and $E_b$ remain frozen during the downstream adaptation evaluated here. Only fusion heads or unimodal readouts are fitted on $\calS_{d,K}$.
\end{enumerate}
\end{definition}

Condition (ii) distinguishes the \Joint\ and \Decoupled\ encoder pools, and condition (iii) specifies downstream adaptation. Results compare complete public checkpoints, including their pretraining objectives, data, and prior task-related supervision. The \Decoupled\ vision tower is a non-contrastive self-supervised encoder at the same architecture and scale, DINOv2~\citep{oquab2024dinov2} where one is released at that size and MAE~\citep{He_2022_CVPR} otherwise. DINOv2 is released at patch $14$ only, so B/16 \Joint\ towers are matched at patch $14$.

\paragraph{Default pairs and reporting.} Each domain has one default encoder pair per fusion paradigm, chosen so that the \Joint\ and \Decoupled\ towers match most closely: CLIP-ViT-L/14 against DINOv2-ViT-L/14 with GPT-2 in the general domain, CLAP against HTS-AT with RoBERTa-base in audio, BiomedCLIP against DINOv2-ViT-B/14 with PubMedBERT in the medical domain, MoleculeSTM against GraphMVP with SciBERT for molecules, and BALM against ESM-2 with ChemBERTa-2 for proteins. For every dataset, budget, and seed, we select one of eight fusion heads on development utility and average the selected models' test utilities across seeds (Eq.~\eqref{eq:head-selection}). The other encoders listed below serve the scaling and domain-pretraining ablations. Development data govern both within-head choices and head identity, with the same development split shared \mbox{across heads and seeds (Appendix~\ref{app:training-details}).}

\paragraph{A strong unaligned pair.} The general-domain \Decoupled\ pool also contains the vision tower of CLIP-ViT-L/14 paired with GPT-2. That vision tower was trained contrastively against CLIP's own text tower but never against GPT-2, so the pair satisfies condition (ii). It provides a comparison in which the vision tower is shared while the text tower and cross-modal pretraining differ.

\subsection{Encoder Table by Modality Family}
\label{app:encoder-families}

\paragraph{General vision--language (CUB, CrisisMMD, Food101, Memotion, MetMeme, OxfordPet, AI2D, ScienceQA).}
Table~\ref{tab:encoders_general} lists the pools by scale tier. For AI2D and ScienceQA, all three paradigms use the same image, question and candidate answers. The fusion models pair the image embedding with a text embedding of the question concatenated with each candidate answer, and \Agent\ receives the question and candidates together with its readout outputs. The \Agent\ readouts are fit on the same unimodal encoders that \Decoupled\ uses (Appendix~\ref{app:tool-schema}).

\begin{table}[t]
\centering
\footnotesize
\setlength{\tabcolsep}{4pt}
\caption{\textbf{Encoder pools for the general domain, by scale tier.} ``Joint FM'' lists the paired models. The vision and text encoders are the matched unimodal towers used by \Decoupled\ and \Agent\ (Definition~\ref{def:matching}). The large-tier pair is the default. SigLIP-SO400M/14 has no matched self-supervised counterpart and is grouped with the large tier. $\ddagger$: the vision tower of CLIP-ViT-L/\mbox{14, never aligned to GPT-2.}}
\label{tab:encoders_general}
\adjustbox{max width=\textwidth}{%
\begin{tabular}{llll}
\toprule
\textbf{Scale tier} & \textbf{Joint FM} & \textbf{Vision encoder} & \textbf{Text encoder} \\
\midrule
Base  & SigLIP-ViT-B/16, CLIP-ViT-B/16 & MAE-ViT-B/16, DINOv2-ViT-B/14 & BERT-base \\
Large & CLIP-ViT-L/14, EVA02-CLIP-L/14, SigLIP-SO400M/14 & DINOv2-ViT-L/14, CLIP-ViT-L/14 vision tower$^{\ddagger}$ & GPT-2 \\
Huge  & OpenCLIP-ViT-H/14 & MAE-ViT-H/14 & GPT-2-medium \\
\bottomrule
\end{tabular}
}
\end{table}

\paragraph{Audio--language (MELD, IEMOCAP).}
The joint FM is CLAP \citep{wu2024largescaleclap} (HTS-AT audio tower, RoBERTa-base text tower) in its general and music variants (\texttt{laion/larger\_clap\_general}, \texttt{laion/larger\_clap\_music}). The \Decoupled\ pool pairs RoBERTa-base with three audio encoders: HTS-AT \citep{chen2022htsat}, AST \citep{gong2021ast}, and wav2vec~2.0-base \citep{baevski2020wav2vec2}. The HTS-AT encoder is the standalone AudioSet-pretrained checkpoint, that is, CLAP's audio tower \emph{before} contrastive language--audio training rather than the tower extracted from CLAP. The comparison measures the combined effect of CLAP's additional paired data, objective, and optimization on this shared base. That pair is the default. AST and wav2vec~2.0 serve the domain-pretraining ablation.

\paragraph{Medical vision--language (Fracture, VQA-RAD).}
The \Joint\ pool contains general CLIP (ViT-B/16, ViT-B/32) and three biomedical paired models (BiomedCLIP \citep{zhang2023biomedclip}, PubMedCLIP \citep{eslami2023pubmedclip}, PLIP \citep{huang2023plip}). The \Decoupled\ pool pairs PubMedBERT \citep{gu2021pubmedbert} with DINOv2-ViT-B/14, MAE-ViT-B/16, Phikon \citep{filiot2023phikon}, or Rad-DINO \citep{perezgarcia2025raddino} (domain ablation in Appendix~\ref{app:domain-ablation}). The default pair is BiomedCLIP against DINOv2-ViT-B/14 \mbox{with PubMedBERT, since BiomedCLIP's text tower is PubMedBERT.}

\paragraph{Molecule (DrugBank-Desc, DrugBank-Pharma).}
The joint FM is MoleculeSTM \citep{liu2023moleculestm}, which aligns molecular graphs (GIN encoder) with textual chemical descriptions (SciBERT \citep{beltagy2019scibert}). The \Decoupled\ encoders are GraphMVP \citep{liu2022graphmvp} (molecular graph) and SciBERT (text).

\paragraph{Protein (KIBA, PDBbind).}
The joint FM is BALM \citep{gorantla2025balm}, which fine-tunes ESM-2 \citep{lin2023esm2} and ChemBERTa-2 \citep{ahmad2022chemberta2} with low-rank adapters and aligns protein and ligand embeddings by cosine similarity in a shared space, in its BindingDB and LeakyPDB variants. The \Decoupled\ encoders use the same base families without BALM\textquotesingle s paired adaptation: ESM-2 ($150$M, protein sequence) and ChemBERTa-2 ($77$M, ligand SMILES). The comparison therefore includes both paired adaptation and alignment.

\paragraph{BALM supervision and held-out evaluation.}
BALM's paired training uses affinity supervision. We address its relationship to the downstream evaluation through the dataset-specific procedures below: KIBA excludes the $60$ pairs overlapping BALM's training data before splitting, while PDBbind assigns BALM's training and validation PDB identifiers to the corresponding downstream splits and reserves the remaining identifiers for testing. These procedures separate held-out evaluation from prior training exposure. The budget $K$ counts labels used for subsequent head or readout fitting. It does not count the supervision used to produce an existing checkpoint. The PDBbind results therefore measure budget-limited downstream adaptation of a checkpoint with prior affinity supervision, evaluated on held-out identifiers.

\subsection{Preprocessing and Splits}
\label{app:preprocessing}

Every dataset has a training, a development and a test split that do not overlap, and the labelled sets $\calS_{d,K}$ are drawn from the training split only. Classification and regression use $k$-shot and percentage budgets, while QA uses percentage budgets. Sampling procedures are specified in Appendix~\ref{app:budget-subsampling}. Seeds $0$ to $4$ give five labelled sets per budget. The main tables report the five-seed mean, and Appendix~\ref{app:perseed} the per-seed standard deviations. The development split is the official one where it exists, a $10\%$ per-class holdout of the training split for Food101, and a random $20\%$ of the training split for OxfordPet, IEMOCAP and VQA-RAD. The development holdouts contain $6{,}403$ examples for Food101, $736$ for OxfordPet, $1{,}606$ for IEMOCAP, and $358$ for VQA-RAD. These examples do not overlap the reported training pools. The $100\%$ budget uses only the remaining pool. Counts after filtering and split construction are reported explicitly \mbox{in Table~\ref{tab:dataset_full}.} Within each dataset, all three paradigms use the same complete test split, fixed across label budgets and seeds. \mbox{The datasets are prepared as follows.}
\begin{itemize}[leftmargin=1.2em,itemsep=1pt]
  \item \textbf{CUB-200-2011.} Image and the first of its ten captions, $200$ classes. All $11{,}788$ images are re-split per class into about $70/15/15$, \mbox{so every class appears in every split.}
  \item \textbf{CrisisMMD.} Tweet image and text of the humanitarian task of CrisisMMD v2.0, $8$ classes, with the text annotation as label. The official train, dev and test splits of the full set are used, which \mbox{keep pairs whose image and text annotations differ.}
  \item \textbf{Food101 (UPMC Food-101).} Image and web-page text truncated to $2{,}000$ characters, $101$ classes. Pairs with no text or under $40$ characters are removed. \mbox{Official train and test splits.}
  \item \textbf{Memotion.} Meme image and \mbox{its OCR text, five sentiment levels, official splits.}
  \item \textbf{MetMeme.} The English memes of MET-Meme only, image and meme text, seven \mbox{sentiment categories, official splits restricted to English.}
  \item \textbf{OxfordPet.} Image and a generated caption, which sometimes names the breed, \mbox{$37$ breeds. Official test split.}
  \item \textbf{AI2D.} Diagram and a question with four options, from the AI2D subset of The Cauldron. Multi-turn records are flattened to one question per row, rows whose options or answer cannot be parsed are dropped, and all questions associated with a diagram are assigned to the same split. Diagram images do \mbox{not cross the training, development, and test splits.}
  \item \textbf{ScienceQA.} Questions that come with an image, each with its two to five options. Lecture and hint fields are not used. Official splits.
  \item \textbf{MELD.} Utterance audio (mono, $16$\,kHz) and transcript, seven emotions. Utterances whose audio could not be extracted are dropped. Official splits.
  \item \textbf{IEMOCAP.} Utterance audio and transcript, with the continuous valence rating \mbox{as target. Official test split.}
  \item \textbf{Fracture.} X-ray image and the patient's chief complaint, which is written \mbox{in several languages, ten fracture types, official splits.}
  \item \textbf{VQA-RAD.} Radiology image and question, evaluated using a shared vocabulary of $500$ candidate answers constructed from the training split alone. The fusion models pair the image embedding with a text embedding of the question concatenated with each candidate answer (Appendix~\ref{app:qa-scoring}). The \Agent\ controller also receives the question. Both open-ended and yes/no questions are retained. Reference answers outside the vocabulary count as incorrect. \mbox{We use the official test split.}
  \item \textbf{DrugBank-Desc and DrugBank-Pharma.} The DrugBank subsets released with MoleculeSTM that exclude PubChem overlap, SMILES and the description or pharmacodynamics text. Each item pairs a molecule with four candidate texts, the correct one and three drawn from other rows of the same split and shuffled using item seed $\mathrm{base\_seed}\times10000+i$, with default MCQ seed $42$. Fusion encodes the fixed question ``Which text describes this molecule?'' together with each candidate text (Appendix~\ref{app:qa-scoring}). Official splits. Following the distinct Description and Pharmacodynamics benchmarks in MoleculeSTM, these are separate prediction tasks with different target texts: general drug descriptions versus pharmacodynamic effects. The routing evaluation groups both DrugBank tasks together in every outer and inner split, \mbox{retaining their distinct task labels and budget cells.}
  \item \textbf{KIBA.} Protein sequence and drug SMILES from the TDC release, with the KIBA score as target. The $60$ pairs that appear in the training data of BALM are removed, the rest is split at random $80/10/10$, and the development split is capped at $5{,}000$ pairs.
  \item \textbf{PDBbind.} Protein sequence and ligand SMILES, with $\mathrm{p}K_d = -\log_{10}$ of the affinity in molar units as target, pooling $K_d$, $K_i$ and IC$_{50}$ measurements. Rows with missing fields, duplicate pairs and $\mathrm{p}K_d$ outside $[3, 12]$ are removed. The split follows the PDB identifiers of BALM's train and validation sets, \mbox{and all other identifiers form the test split.}
\end{itemize}

Each frozen encoder uses the input format and preprocessing supplied with its released checkpoint. Text is tokenized with the model-specific tokenizer (BERT tokenizer for BERT-tower models, GPT-2 tokenizer for GPT-2-tower models). Images are resized to $224 \times 224$ and normalised with the statistics shipped with each checkpoint. Audio is encoder-specific in the same way. Each clip is downmixed to mono and resampled to the encoder's native rate, clips longer than $10$ s are split into $10$ s windows that are encoded independently, and the window embeddings are mean-pooled and $L_2$-normalised into one vector per clip. Molecular graphs are built by the featurizer released with the encoder that consumes them, and ligand SMILES and protein sequences are tokenized by \mbox{the tokenizers released with ChemBERTa and ESM-2 respectively.}

\subsection{Domain-Pretraining Ablation}
\label{app:domain-ablation}

The encoder-configuration ablation swaps the \Joint\ model or one \Decoupled\ tower of the medical and audio pools above, retaining the downstream training and evaluation protocol. On Fracture, \Joint\ is run with the general CLIP-ViT-B/16 and with each of the three biomedical paired models, and \Decoupled\ with the general DINOv2-ViT-B/14 and with the radiology and pathology encoders Rad-DINO and Phikon, all with PubMedBERT as the text tower. PubMedCLIP uses a ViT-B/32 vision tower and CLIP's causal text Transformer. Against the default \Decoupled\ pair, this changes both visual patch size ($32$ versus $14$) and text backbone (CLIP versus PubMedBERT), alongside pretraining. On MELD, the general-audio CLAP is compared with its music variant as a wrong-domain control, and \Decoupled\ is run with each of the three audio encoders. Results are in Appendix~\ref{app:domain-pretraining-ablation}.

\FloatBarrier
\section{Training and Model Selection Details}
\label{app:training-details}

Sections~\ref{app:training-objective}, \ref{app:optimizer}, and~\ref{app:fusion-operators} specify fusion training. Sections~\ref{app:agent-inference-details}, \ref{app:budget-subsampling}, and~\ref{app:implementation} give Agent inference, budget sampling, and implementation details. Grouped evaluation and the routing configurations are in Sections~\ref{app:eval-protocol}--\ref{app:metadata-only}.

\subsection{Shared Training Objective}
\label{app:training-objective}

Both fusion paradigms (\Joint\ and \Decoupled) optimise the same supervised objective:
\[
    \min_\theta
    \frac{1}{|\calS_{d,K}|}
    \sum_{(x,y)\in\calS_{d,K}}
    \ell_d(f_\theta(x),y) ,
\]
where $\ell_d$ is cross-entropy for classification and MCQ tasks, mean-squared error for regression (KIBA, PDBbind, IEMOCAP valence). Encoders are frozen throughout, and only the downstream head parameters $\theta$ are updated. Regularisation is the decoupled weight decay $\lambda$ of AdamW (Table~\ref{tab:training_hparams}) rather than an $\ell_2$ term added to the loss.

\subsection{Optimizer and Hyperparameters}
\label{app:optimizer}

\begin{table}[t]
\centering
\footnotesize
\setlength{\tabcolsep}{4pt}
\caption{\textbf{Classification and regression fusion-training hyperparameter grid.} QA fusion uses the fixed configuration in Appendix~\ref{app:qa-scoring}. Selection is nested within each seed: development utility selects hyperparameters and checkpoints within each head, then selects head identity. The selection rule is given in Eq.~\eqref{eq:head-selection}.}
\label{tab:training_hparams}
\adjustbox{max width=\textwidth}{%
\begin{tabular}{ll}
\toprule
\textbf{Item} & \textbf{Values} \\
\midrule
Optimizer & AdamW ($\beta_1=0.9$, $\beta_2=0.999$, $\epsilon=10^{-8}$) \\
Learning rate & $\{10^{-5}, 3\times10^{-5}, 10^{-4}, 3\times10^{-4}, 10^{-3}\}$ \\
Weight decay $\lambda$ & $\{0, 10^{-4}, 10^{-3}, 10^{-2}\}$ \\
Dropout & $\{0.0, 0.1, 0.3\}$ applied to the head input \\
Batch size & $\{32, 64, 128\}$. Smaller batches for very low budgets \\
Maximum epochs & 200 with early stopping (patience = 20 epochs) \\
Stopping criterion & development utility $U_{\calV_d}(f)$, Eq.~\eqref{eq:utility} \\
Seeds & 5 labelled-set seeds. Report the mean of seed-specific selected models \\
\bottomrule
\end{tabular}
}
\end{table}

Table~\ref{tab:training_hparams} lists the classification and regression fusion-training grid. QA fusion uses the separate fixed configuration in Section~\ref{app:qa-scoring}. For the 20-shot setting, we additionally restrict the learning rate search to $\{10^{-4}, 3\times10^{-4}, 10^{-3}\}$ and cap the batch size at $\min(32, |\calS_{d,K}|)$.

\subsection{Head Architectures}
\label{app:fusion-operators}

For each fusion configuration, dataset, budget, and seed, we fit all eight heads on the same labelled set and select their hyperparameters, checkpoints, and final head identity by development utility (Eq.~\eqref{eq:head-selection}). Classification and regression use the grid in Table~\ref{tab:training_hparams}, with at most $200$ epochs and early-stopping patience $20$. QA uses at most $100$ epochs under the configuration below. Reported means aggregate the seed-specific selections. Separate per-head test grids describe head sensitivity.

\paragraph{\Joint head (deployed).}
The joint FM exposes a paired two-tower output, pooled to $(u_J, v_J) = F_J(x^{(a)}, x^{(b)})$ and retained as token sequences for the one head that needs them, over which \Joint fits the same eight heads as \Decoupled. For fixed-class classification, the simplest of them is concat-then-linear, $h_\theta^J([u_J; v_J]) = W [u_J; v_J] + b$ with $W \in \mathbb{R}^{|\calY_d| \times (r_a + r_b)}$. For regression, the output is a scalar with no activation. Tables~\ref{tab:agent_fm_cls_reg} and~\ref{tab:agent_fm_qa} report the mean of seed-specific development-selected \Joint\ models. Table~\ref{tab:fusion_joint} \mbox{is a separate per-head diagnostic grid.}

\paragraph{\Decoupled fusion heads (deployed).}
We search over eight fusion families, summarised \mbox{in Table~\ref{tab:fusion_heads}.} Seven of them consume the pooled per-side outputs $(u, v) = (E_a(x^{(a)}), E_b(x^{(b)}))$. The \texttt{attention} family instead reads the token sequences $(Z^A, Z^B)$ that pooling collapses:

\begin{table}[t]
\centering
\footnotesize
\setlength{\tabcolsep}{4pt}
\caption{\textbf{Fusion head family descriptions.} Fixed-class output layers below describe classification. In QA, the second input is a candidate-text embedding and the output is the shared scalar scoring head in Appendix~\ref{app:qa-scoring}, rather than a fixed class-specific output layer. Citations identify architectural inspirations. The formulas specify our implementations, not full reproductions of the cited systems. $u \in \mathbb{R}^{r_a}$, $v \in \mathbb{R}^{r_b}$ are the pooled per-side embeddings and $d_h$ is a hidden dimension selected from $\{256, 512\}$ where applicable; the \texttt{attention} head uses a fixed width of $128$. The \texttt{attention} head is the one family that does not consume the pooled vectors: it reads the per-side token sequences $Z^A, Z^B$ that pooling would otherwise collapse, and pools after attending. The resulting vector width depends on the fusion operator.}
\label{tab:fusion_heads}
\adjustbox{max width=\textwidth}{%
\begin{tabular}{lll}
\toprule
\textbf{Name} & \textbf{Formula} & \textbf{Parameters} \\
\midrule
\texttt{concat} & $[u; v]$ then linear & $W \in \mathbb{R}^{(r_a+r_b)\times|\calY|}$ \\
\texttt{add} & $W_a u + W_b v$ (shared $d_h$) then linear & $W_a, W_b \in \mathbb{R}^{d_h\times r_a/r_b}$ \\
\texttt{gated}~\citep{arevalo2017gated} & $g \odot (W_a u) + (1-g) \odot (W_b v)$, $g=\sigma(W_g[u;v])$ & gate $W_g \in \mathbb{R}^{d_h\times(r_a+r_b)}$ \\
\texttt{bilinear}~\citep{kim2017hadamard} & $u^\top W_c v$, a rank-$32$ bilinear map, then linear & $W_c \in \mathbb{R}^{r_a\times 32\times r_b}$ \\
\texttt{attention}~\citep{tsai-etal-2019-multimodal,li2023blip2} & learned queries $Q$ cross-attend to both token sequences, $\mathrm{Pool}\big(\mathrm{Attn}(Q, K{=}V{=}[Z^A; Z^B])\big)$, then linear & 2-head, $d_h=128$ \\
\texttt{film}~\citep{Perez_Strub_deVries_Dumoulin_Courville_2018} & $(\gamma(v) \odot W u + \beta(v))$ FiLM layer & $\gamma, \beta: \mathbb{R}^{r_b}\to\mathbb{R}^{d_h}$ \\
\texttt{mlp} & 2-layer MLP on $[u;v]$ with ReLU & hidden $d_h$ \\
\texttt{logit\_sum} & readout logits summed: $h_a(u) + h_b(v)$ & two linear heads \\
\bottomrule
\end{tabular}
}
\end{table}

\paragraph{QA candidate scoring.}
\label{app:qa-scoring}
For \Joint\ and \Decoupled, each fusion operator takes an image or molecular embedding $u_i$ and one candidate-text embedding $v_{ij}$. Candidates use the configuration's frozen text encoder. For AI2D, ScienceQA, and VQA-RAD, its input concatenates the question with each candidate answer. All candidate positions share the scalar head
\[
 s_{ij}=\mathrm{Linear}_{256\to1}\!\left(\mathrm{Dropout}_{0.1}\!\left(\mathrm{ReLU}\!\left(\mathrm{Linear}_{d_\psi\to256}(\psi_\omega(u_i,v_{ij}))\right)\right)\right).
\]
The non-text embedding is repeated across candidates, yielding a $B\times M_{\max}$ score matrix. Padding is masked with $-\infty$ before cross-entropy over the correct option. QA fusion uses AdamW (learning rate $10^{-3}$, weight decay $10^{-4}$), cosine scheduling, batch size $64$, and at most $100$ epochs, with development-accuracy early stopping and head selection. VQA-RAD uses the same scorer over $500$ shared answers. Each VQA-RAD candidate embedding includes the question and the candidate answer. DrugBank candidates encode the fixed question together with each candidate text (Appendix~\ref{app:datasets}).

\subsection{Agent Inference Details}
\label{app:agent-inference-details}

The \Agent controller and encoders have frozen weights. For each dataset, budget, and seed $s$, the trainable single-modality readouts are fitted independently on $\calS_{d,K}^{(s)}$. Each trainable readout selects its own checkpoint by maximizing development utility on the shared disjoint split $\calV_d$ (accuracy for classification/QA, negative MAE for regression). For image MCQ, the image readout uses bilinear scores $s_j=o_j^\top Wv$ on frozen \Decoupled\ image and option embeddings, trained by cross-entropy on the training split. Weight decay and training duration are selected on the development split. The text readout directly computes unscaled cosine similarity between sample-text and option embeddings in the frozen \Decoupled\ text embedding space and has no trained head. The reported \Agent\ result averages the five seed-specific systems, each using its corresponding selected readouts and labelled-set demonstrations. We use Qwen3-32B as the instruction-tuned language model controller. The key inference settings \mbox{are listed in Table~\ref{tab:agent_inference}:}

\begin{table}[t]
\centering
\footnotesize
\setlength{\tabcolsep}{4pt}
\caption{\textbf{\Agent inference configuration.}}
\label{tab:agent_inference}
\adjustbox{max width=\textwidth}{%
\begin{tabular}{ll}
\toprule
\textbf{Item} & \textbf{Value} \\
\midrule
LLM backbone & Qwen3-32B (instruction-tuned, frozen) \\
Temperature & 0.0 (greedy decoding for reproducibility) \\
LLM turns & $1$ in Inline-1 (both readouts pre-filled), $2$ in Multi-2 (one tool call, then the answer, Section~\ref{sec:agent-variants}) \\
Maximum output tokens & 4096 \\
Tool schema format & OpenAI-function-call JSON \\
Parser & task-specific regex with fallback to string match \\
Unparseable answer & scored as wrong (a regression answer that cannot be parsed is taken as $0$) \\
In-context examples & up to $N_{\mathrm{ICL}}{=}10$, drawn class-balanced from $\calS_{d,K}$ \\
\midrule
Evaluation examples & the same complete test split as \Joint\ and \Decoupled\ within each dataset \\
\bottomrule
\end{tabular}
}
\end{table}

The agent tools return structured records as described in Appendix~\ref{app:agent-prompt}. In the deployed pre-filled mode both records are written into the user message before the single turn. In the multi-turn variant of Section~\ref{sec:agent-variants} they are returned as ``tool'' role \mbox{messages following the Hermes-2 format used by Qwen3.}

\paragraph{Shared test examples.} Within each dataset, all three paradigms are evaluated on the same complete test split, fixed across label budgets and seeds. Table~\ref{tab:dataset_full} reports the actual scored counts for all $16$ datasets. The control experiments use these same complete splits.

\subsection{Label Budget Subsampling}
\label{app:budget-subsampling}

The reported budget counts labelled training examples only. The separate development split used for hyperparameter selection and early stopping is held fixed across budgets and is excluded from these counts. For each dataset $d$ and budget $K$, the labelled set $\calS_{d,K}$ is constructed as follows. At a $k$-shot budget, classification tasks target $k$ examples per class, sampled without replacement; the achieved counts can be lower than $k|\mathcal{Y}_d|$ (Table~\ref{tab:label-counts}). Regression tasks use $k$ examples per quantile bucket over five equal-frequency buckets of the target ($100$ in total at $20$-shot). At a percentage budget, classification tasks keep that share of every class, and regression and QA tasks sample uniformly. \Joint\ and \Decoupled\ use the same labelled set at each seed. The relative sizes of the $20$-shot and $10\%$ budgets depend on the dataset and its class count, so budget trends should be read against the actual number of labelled examples, listed for each cell \mbox{in Table~\ref{tab:label-counts}.}

\begin{table}[t]
\centering
\footnotesize
\setlength{\tabcolsep}{4pt}
\caption{\textbf{Number of labelled training examples behind every budget cell} (seed $0$). Shots are per class for classification and per quantile bucket (five buckets) for regression, so a $20$-shot cell holds $20\,|\mathcal{Y}_d|$ or $100$ labels and can exceed the $10\%$ cell on datasets with many classes (CUB, OxfordPet) or a small training split (Fracture). QA datasets are run at percentage budgets only. \mbox{Counts exclude the separate development split.}}
\label{tab:label-counts}
\adjustbox{max width=\textwidth}{%
\begin{tabular}{@{}lcccccccccccccc@{}}
\toprule
\textbf{Dataset} & $k{=}$1 & $k{=}$5 & $k{=}$20 & $k{=}$50 & $k{=}$100 & 10\% & 20\% & 30\% & 40\% & 50\% & 60\% & 70\% & 80\% & 100\% \\
\midrule
CUB & 200 & 1,000 & 4,000 & -- & -- & 795 & -- & 2,518 & -- & 4,106 & -- & 5,722 & -- & 8,244 \\
CrisisMMD & 8 & 40 & 160 & 375 & 675 & 1,361 & -- & 4,083 & -- & 6,803 & -- & 9,525 & -- & 13,608 \\
Food101 & 101 & 505 & 2,020 & 5,050 & 10,100 & 5,810 & -- & 17,431 & -- & 29,053 & -- & 40,674 & -- & 58,106 \\
Memotion & 5 & 25 & 100 & 250 & 500 & 559 & -- & 1,679 & -- & 2,796 & -- & 3,914 & -- & 5,593 \\
MetMeme & 7 & 35 & 140 & 350 & 673 & 239 & -- & 718 & -- & 1,197 & -- & 1,675 & -- & 2,394 \\
OxfordPet & 37 & 185 & 740 & 1,850 & -- & 294 & 590 & 882 & 1,177 & 1,472 & 1,767 & 2,061 & 2,354 & 2,944 \\
MELD & 7 & 35 & 140 & 350 & 700 & 998 & -- & 2,997 & -- & 4,994 & -- & 6,992 & -- & 9,988 \\
Fracture & 10 & 50 & 200 & 500 & 778 & 79 & 159 & 237 & 316 & 394 & 474 & 553 & 631 & 790 \\
IEMOCAP & 5 & 25 & 100 & 250 & 500 & 642 & -- & 1,928 & -- & 3,212 & -- & 4,498 & -- & 6,425 \\
KIBA & 5 & 25 & 100 & 250 & 500 & 9,408 & -- & 28,223 & -- & 47,038 & -- & 65,854 & -- & 94,077 \\
PDBbind & 5 & 25 & 100 & 250 & 500 & 1,090 & -- & 3,271 & -- & 5,452 & -- & 7,632 & -- & 10,903 \\
AI2D & -- & -- & -- & -- & -- & 597 & -- & 1,791 & -- & 2,984 & -- & 4,178 & -- & 5,969 \\
ScienceQA & -- & -- & -- & -- & -- & 622 & -- & 1,865 & -- & 3,109 & -- & 4,353 & -- & 6,218 \\
VQA-RAD & -- & -- & -- & -- & -- & 144 & 287 & 430 & 574 & 718 & 861 & 1,004 & 1,148 & 1,435 \\
DrugBank-Desc & -- & -- & -- & -- & -- & 92 & -- & 277 & -- & 462 & -- & 646 & -- & 923 \\
DrugBank-Pharma & -- & -- & -- & -- & -- & 80 & -- & 241 & -- & 402 & -- & 563 & -- & 804 \\
\bottomrule
\end{tabular}}
\end{table}

\subsection{Implementation Notes}
\label{app:implementation}

Feature extraction uses the HuggingFace \texttt{transformers} and \texttt{open\_clip} libraries. All encoders are loaded from their official checkpoints with no modification. Final head training is performed with PyTorch 2.1 on NVIDIA A100 (80 GB) or H100 (80 GB) GPUs.

\subsection{Grouped Routing Evaluation}
\label{app:eval-protocol}
The cascade, direct classifier, and TB-Majority are evaluated on the same $144$ dataset--budget cells. The $16$ datasets define $15$ groups: DrugBank-Desc and DrugBank-Pharma form one group and every other dataset forms its own. An outer fold holds out every budget and seed of one group. All inner leave-one-group-out operations use the same group identities, so the DrugBank tasks never straddle an inner or outer split. They remain distinct prediction \mbox{tasks. Their observations or labels are not merged.}

Each held-out cell reads the features of one labelled set per run, while training cells use the five-set feature mean. Reported accuracy and relative regret average five runs and weight cells equally. The text probe uses run-specific example sets. CUB and OxfordPet retain the same $20$ classes, and the minimum-budget or reused pools constrain example variation (Appendix~\ref{app:zs-feature}). Candidate utilities, and therefore the tie-aware target labels, are based on five-seed mean test scores. Labels are re-derived at every tolerance. Training-set features and metadata are used to predict these targets, without \mbox{reading held-out candidate utilities during fitting or selection.}

\paragraph{Paired group bootstrap.}
For each cell and seed, compute cascade correctness minus comparator correctness, or cascade relative regret minus comparator relative regret. For each of $4{,}000$ replicates, sample $15$ groups with replacement, keeping all budgets and five seed rows of every sampled group, including repeated groups with their multiplicities. Divide the sum of paired row differences by the sampled row count. Each cell receives equal weight in this statistic. The $2.5$th and $97.5$th percentiles give each pointwise $95\%$ interval. Sampling uses NumPy \texttt{default\_rng(0)}. The two DrugBank tasks remain together in every replicate. Utilities and winner labels are held fixed, so these intervals quantify variation across observed groups conditional on those utilities. They do not resample test \mbox{examples or establish simultaneous coverage across tolerances.}

\subsection{Cascade Configuration}
\label{app:cascade-config}
The reported cascade uses two L2-regularized logistic regressions with inverse regularization strength $C=1$. Step~1 is fit on all outer-training cells with the Agent/Fusion targets. Step~2 is fit only on Fusion-labelled training cells with Joint/Decoupled targets. Imputation, missingness-indicator construction, standardization, and classifier fitting use only the training partition of each individual inner or outer fit. Missingness indicators are appended before standardization. No outer-held-out group rows are used to fit preprocessing, \mbox{rank features, choose configurations, or determine thresholds.}

\paragraph{Step labels}
\label{app:step-labels}
Step~1 maps the subset $\phi_1(d,K)$ to an \Agent\ score with a classifier $f_1$ trained on the labels
\begin{equation}
    y_1(d,K;\epsilon) \;=\;
    \begin{cases}
        \Agent, & U(\Agent;d,K) \;\geq\; \displaystyle\max_{p \in \{\Joint,\Decoupled\}} U(p;d,K) \;-\; \epsilon_d, \\[6pt]
        \mathrm{Fusion},  & \text{otherwise.}
    \end{cases}
\end{equation}
Step~2 maps the subset $\phi_2(d,K)$ to a \Joint\ score with a classifier $f_2$ trained on the cells whose Step-1 label is $\mathrm{Fusion}$, with the labels
\begin{equation}
    y_2(d,K;\epsilon) \;=\;
    \begin{cases}
        \Joint,    & U(\Joint;d,K) \;>\; U(\Decoupled;d,K) + \epsilon_d, \\[4pt]
        \Decoupled,& \text{otherwise.}
    \end{cases}
    \label{eq:jd-label}
\end{equation}
On Fusion-labelled cells these labels implement Eq.~\eqref{eq:tie-set}. If an \Agent\ cell is incorrectly passed to Step~2, Eq.~\eqref{eq:jd-label} defines its fallback fusion choice.

\paragraph{Budget-specific features and size adjustment.}
Mutual-kNN overlap uses the default encoder pair on the full training pool, capped at $8{,}000$ paired observations. All budget cells of a dataset share this label-free feature. Other embedding diagnostics use the current budget-specific subset $\calS_{d,K}^{(s)}$. The five supervised representation diagnostics are residualized as $\tilde f=f-(a+b\log_{10}|\calS_{d,K}|)$, using linear fits on outer-training cells separately for regression and other tasks. The covariate is the number of labelled examples in that cell, not the complete training-pool size used in the budget metadata. \mbox{The text diagnostic is not residualized.}

\paragraph{Outer-fold preprocessing and inner validation.}
Residualization and Laplacian ranking are fitted once on the outer-training datasets, using diagnostic features and budget information without paradigm-winner labels. Inner validation groups contribute to these transformations, which remain fixed during grouped selection of the feature width and budget block. Imputation, missingness indicators, standardization, and the classifier are refitted on each individual fit's training partition. The outer evaluation measures the complete selection procedure on a group excluded from transformation fitting, configuration selection, and threshold estimation. Its diagnostics are computed from the permitted target-training inputs and budget-subset labels, then processed by the fitted transformations.

\paragraph{Fixed features and candidates.}
Step~1 fixes the regression and QA indicators. Step~2 fixes QA. The symbolic-sequence indicator is an ordinary ranked candidate in both steps. Other eligible diagnostics enter the same pool. No preliminary screening is applied. The text diagnostic $\Delta_{\mathrm{zs}}$ participates only in Step~1 ranking and is appended if absent from the chosen width. It is therefore always included in the final Step~1 predictor. It is a label-dependent proxy for text-side task information.

\paragraph{Laplacian ranking.}
Form one row per training dataset by averaging its features across budgets. Standardize these rows and construct a $5$-nearest-neighbour graph with weights $W_{ij}=\exp(-\|u_i-u_j\|^2/t)$, then symmetrize the graph. The bandwidth is the squared median Euclidean distance. With $D=\operatorname{diag}(W\mathbf{1})$ and $L=D-W$, define
\[
\tilde f=f-\frac{f^\top D\mathbf{1}}{\mathbf{1}^\top D\mathbf{1}}\mathbf{1},\qquad
\mathrm{LS}(f)=\frac{\tilde f^\top L\tilde f}{\tilde f^\top D\tilde f}.
\]
Rank eligible features in increasing Laplacian score \citep{he2005laplacian}. The ranking uses no paradigm-winner labels. Dataset-level rows and evaluation groups are distinct: the two DrugBank rows may both occur in a training fold but are always \mbox{excluded together when their group is held out.}

\paragraph{Width and budget block.}
Candidate widths are $\{1,2,3,4,6,8,10,12,16,21\}$ and the candidate-pool size. Each configuration combines the leading ranked features, the step's fixed features, and one budget block: none, all five descriptors, or all except the budget fraction. For every inner held-out group $h$, evaluate a configuration by pooled leave-one-group-out AUROC on the remaining outer-training groups. Average these AUROCs over $h$ and choose the maximizing configuration. Selection is performed separately for the two steps.

\paragraph{Prior-matched thresholds.}
With the feature configuration selected, obtain grouped out-of-fold logistic decision scores on each step's training cells, conditional on the fixed outer-fold residualization and ranking. Let $\pi_t$ be that table's fraction of positive labels. Set $\tau_t$ to the empirical $(1-\pi_t)$ quantile of its out-of-fold scores. For Step~2, both the prior and scores are restricted to Fusion-labelled training cells. Fit each final classifier on its outer-training table and apply Eq.~\eqref{eq:framework}. Scores at the threshold are assigned to the positive class. Quantile interpolation and tied scores make prior matching approximate. Threshold selection uses the training-class proportion rather than three-way accuracy.

\subsection{Direct Three-Way Classifier}
\label{app:direct-config}
The direct comparator is a single three-class L2-regularized logistic regression ($C=1$), predicting Agent, Joint, or Decoupled by the largest class probability. It accesses the same diagnostic and task-metadata information and uses the same missingness representation, imputation, standardization, and grouped feature-selection framework as the cascade. Its feature rules match Cascade Step~1: regression and QA indicators are fixed. The symbolic-sequence indicator is an ordinary ranked candidate. $\Delta_{\mathrm{zs}}$ participates in ranking and is appended if absent from the selected width. Preprocessing follows the same outer-fold and fit-specific boundaries as the cascade. Targets follow the same tie convention at each tolerance. Selection uses pooled macro one-vs-rest three-class AUROC. The comparison therefore evaluates the full predictors, including their different decision rules, under shared diagnostic access.

\subsection{Metadata-Only Ablation}
\label{app:metadata-only}
The metadata-only ablation retains the cascade's two-stage structure, task and budget metadata, grouped tuning procedure, and prior-matched thresholds. It removes representation geometry, supervised representation diagnostics, and the task-specific text diagnostic, measuring their combined contribution beyond metadata. Table~\ref{tab:routing-feature-ablation} also reports additions and removals of individual diagnostic groups.

\FloatBarrier
\section{Cascade Feature Inventory}
\label{app:decision-rule}

This appendix defines the base features (Section~\ref{app:features}), text diagnostic (Section~\ref{app:zs-feature}), and metadata and missing-value handling (Section~\ref{app:metadata-missingness}) used by the cascade in Section~\ref{sec:hier-selection}.

\begin{table}[t!]
\centering
\footnotesize
\setlength{\tabcolsep}{4pt}
\caption{\textbf{The $26$ base features of $\phi(d,K)$}, grouped by extraction angle. Geometry and kNN features use $E_a,E_b$. The text--label probe additionally uses a fixed sentence encoder. Each per-side base function is evaluated on both modalities and kept as its minimum and maximum over the two sides. $n_{\mathrm{train}}$ is the size of the training split and $|\calY_d|$ the number of classes (five quantile buckets for regression, undefined for MCQ and imputed). CKA and mutual-kNN require paired observations. Labelled kNN on $[A;B]$ requires paired labelled observations. Fold-defined missingness indicators are appended before standardization. The two ANOVA and three labelled kNN features form the supervised representation diagnostics group used in the ablations. Per-feature definitions are in Appendix~\ref{app:features}.}
\label{tab:feature-pool-summary}
\renewcommand{\arraystretch}{1.1}
\begin{tabularx}{\textwidth}{@{}>{\raggedright\arraybackslash}p{2.9cm} c >{\raggedright\arraybackslash}p{4.6cm} >{\raggedright\arraybackslash}X@{}}
\toprule
\textbf{Extraction angle} & \textbf{\#} & \textbf{Base functions} & \textbf{What it captures} \\
\midrule
\rowcolor{gray!10}\multicolumn{4}{c}{\textbf{\textit{Label-free geometry ($12$ features)}}} \\
Per-modality geometry & $10$ & $\mathrm{unif}$, $\mathrm{srank}$, $\mathrm{decay}$, $\mathrm{cond}$, $\mathrm{iso}$ & shape of each side's embedding cloud \\
\cmidrule(lr){1-4}
Cross-modal alignment & $2$ & $\mathrm{CKA}$, $\mathrm{mknn}_{10}$ & agreement between the two spaces \\
\midrule
\rowcolor{gray!10}\multicolumn{4}{c}{\textbf{\textit{Label, budget, and task features ($14$ features)}}} \\
ANOVA separability & $2$ & $f_{\mathrm F}$ & class separability of each side \\
\cmidrule(lr){1-4}
Frozen-embedding task signal & $3$ & $\mathrm{knn5}$ & label signal a $k$-NN recovers, per side and on $[A;B]$ \\
\cmidrule(lr){1-4}
Text diagnostic & $1$ & $\Delta_{\mathrm{zs}}$ (always in Step 1) & task-type-specific text signal for routing \\
\cmidrule(lr){1-4}
Budget descriptors & $5$ & $\mathrm{budget\_frac}$, $\mathrm{is\_kshot}$, $\log_{10} n_{\mathrm{train}}$, $\log_{10}(n_{\mathrm{train}}/|\calY_d|)$, $\log_{10}(|\calS_{d,K}|/|\calY_d|)$ & supervision available at this budget \\
Task metadata & $3$ & $r_d$, $q_d$, $\sigma_d$ & task format and symbolic-sequence inputs \\
\bottomrule
\end{tabularx}
\end{table}

\begin{table}[t]
\centering\footnotesize
\caption{\textbf{Data required by routing features.} The complete pipeline operates on paired target data and selects from these feature groups at each step. The table distinguishes features that use sample correspondence \mbox{from those computed independently within each modality.}}
\label{tab:feature-availability}
\begin{tabularx}{\textwidth}{@{}>{\raggedright\arraybackslash}p{3.8cm}>{\raggedright\arraybackslash}X>{\raggedright\arraybackslash}X@{}}
\toprule
Feature group & Observations required & Labels required \\
\midrule
Per-side uniformity, rank, decay, condition, isotropy & Samples from each modality, with no cross-modal correspondence & None \\
Per-side ANOVA and kNN score & Modality-specific samples, with no cross-modal correspondence & Compatible task labels on each side \\
CKA and mutual-kNN overlap & Corresponding observations of both modalities & None \\
kNN score on $[A;B]$ & Paired observations & Labels for those paired observations \\
$\Delta_{\mathrm{zs}}$ (Step 1) & Text inputs and fixed sentence-encoder probe, with no cross-modal correspondence & Text-side labels and classification class names \\
Budget descriptors & Training-pool and labelled-set sizes & Task/class metadata, with a separate size convention if the two pools differ \\
Task indicators & Task format and modality types & Metadata only \\
Missingness indicators & Feature availability in the training fold & No additional labels \\
\bottomrule
\end{tabularx}
\end{table}

\subsection{Feature Inventory}
\label{app:features}

The feature vector $\phi(d,K)$ contains $26$ base features plus fold-defined missingness indicators, grouped by extraction type of Section~\ref{sec:geometry-summary}. We first define the seven \emph{per-side base functions} (five label-free geometric functions, ANOVA separability, and kNN task signal), then the two \emph{cross-modal base functions} on paired embeddings, and the five \emph{budget descriptors}. The PI aggregators that turn per-side values into symmetric features are defined in Section~\ref{app:features-aggregators}.

Throughout, $Z = (z_1, \ldots, z_n)^\top \in \mathbb{R}^{n \times r_m}$ is the embedding matrix returned by encoder $E_m:\calX_m\to\mathbb{R}^{r_m}$ on the observations used by the feature (the current budget subset $\calS_{d,K}^{(s)}$, except for full-training-pool mutual-kNN), $\hat z_i = z_i / \|z_i\|_2$ is the unit-norm row, $\bar z = \tfrac{1}{n}\sum_i z_i$ is the mean, and $Z_c = Z - \mathbf{1}\bar z^\top$ is the centred matrix with singular values $\sigma_1 \ge \cdots \ge \sigma_q > 0$. All expectations over pairs $i \neq j$ are estimated by uniform Monte Carlo with $50{,}000$ samples and a fixed seed.

\paragraph{Per-side geometric base functions.}

\begin{equation}
\mathrm{unif}(Z) \;=\; \log \mathbb{E}_{i \neq j}\!\left[\exp\!\big(-2\,\|\hat z_i - \hat z_j\|_2^2\big)\right] ,
\label{eq:feat-unif}
\end{equation}
the Wang--Isola hypersphere uniformity \citep{wang2020understanding}. Smaller (more negative) values indicate \mbox{more uniformly spread embeddings on the unit sphere.}

\begin{equation}
\mathrm{srank}(Z) \;=\; \frac{\sum_{i=1}^{q} \sigma_i^2}{\sigma_1^2}
\;=\; \frac{\|Z_c\|_F^2}{\|Z_c\|_2^2} ,
\label{eq:feat-srank}
\end{equation}
the stable rank of the centred embedding matrix (\mbox{a robust effective dimensionality).}

\begin{equation}
\mathrm{decay}(Z) \;=\; \mathrm{slope}_i\!\left(\log \sigma_i\,\big|\, i = 1, \ldots, q\right),
\label{eq:feat-decay}
\end{equation}
the least-squares slope of the log singular values (more negative means faster decay).

\begin{equation}
\mathrm{cond}(Z) \;=\; \frac{\sigma_{\max}}{\sigma_{\min}} ,
\label{eq:feat-cond}
\end{equation}
the spectral \mbox{condition number over the strictly-positive singular values.}

\begin{equation}
\mathrm{iso}(Z) \;=\; \mathbb{E}_{i \neq j}\!\left[\langle \hat z_i, \hat z_j\rangle\right] ,
\label{eq:feat-iso}
\end{equation}
the mean cosine similarity of random pairs. Larger values indicate greater anisotropy.

\begin{equation}
\mathrm{F}(Z,y)=
\frac{\sum_c n_c\|\mu_c-\bar z\|_2^2/(C-1)}
{\sum_c\sum_{i:y_i=c}\|z_i-\mu_c\|_2^2/(N-C)},
\qquad f_{\mathrm F}(Z,y)=\log_{10}\mathrm{F}(Z,y).
\label{eq:feat-fisher}
\end{equation}
This ANOVA-style statistic normalizes between- and within-class scatter by their degrees of freedom, removing the explicit $(C-1)/(N-C)$ factor of the unnormalized ratio. Its null expectation depends on the data distribution and coordinate dependence. Computation uses at most $8{,}000$ rows of the labelled subset and omits singleton classes. Here $N$, $C$, $n_c$, and the means describe the retained observations and represented classes. Regression targets use bins $\lfloor10y\rfloor$, of width $0.1$ in the target units. These differ from \mbox{the five quantile buckets used for budget sampling.}

\paragraph{Per-side task-signal base function.}

\begin{equation}
\mathrm{knn5}(Z, y) \;=\; \frac{1}{n}\sum_{i=1}^{n} \mathbf{1}\!\left[
  \mathrm{majority}\!\left(\{y_j : j \in \mathcal{N}_5^{\cos}(\hat z_i)\}\right) \,=\, y_i
\right] ,
\label{eq:feat-knn5}
\end{equation}
the leave-one-out $k$-nearest-neighbour accuracy with cosine distance and $k=5$, where $\mathcal{N}_5^{\cos}(\hat z_i) = \arg\mathrm{top}_{5\,j \neq i}\, \langle \hat z_i, \hat z_j\rangle$. The query examples, neighbour candidates and labels are restricted to $\calS_{d,K}$. For classification this is plain accuracy. For regression we use the Pearson $r$ of the kNN mean prediction against the target. The combined value applies the same rule to the concatenated unit-norm embeddings $[\hat z_i^A; \hat z_i^B]$, which requires paired labelled examples. The two per-side values can instead be computed \mbox{from separate labelled pools with compatible target definitions.}

\paragraph{Cross-modal base functions.}
For paired matrices $Z^A\in\mathbb{R}^{n\times r_a}$ and $Z^B\in\mathbb{R}^{n\times r_b}$:

\begin{equation}
\mathrm{CKA}(Z^A, Z^B) \;=\; \frac{\big\|(Z^A_c)^\top Z^B_c\big\|_F^2}
                                    {\big\|(Z^A_c)^\top Z^A_c\big\|_F \cdot \big\|(Z^B_c)^\top Z^B_c\big\|_F} ,
\label{eq:feat-cka}
\end{equation}
the linear centred-kernel-alignment between the two embedding spaces \citep{kornblith2019similarity}.

\begin{equation}
\mathrm{mknn}_{10}(Z^A, Z^B) \;=\; \frac{1}{n}\sum_{i=1}^{n} \frac{\big| \mathcal{N}_{10}^{\cos}(\hat z_i^A) \cap \mathcal{N}_{10}^{\cos}(\hat z_i^B) \big|}{10} ,
\label{eq:feat-mknn}
\end{equation}
the mutual $k$-nearest-neighbour overlap at $k=10$. It is computed once per dataset on the default encoder pair's full training pool, using at most $8{,}000$ paired rows, and shared by all budget cells. For $n<11$, the implementation returns NaN without reducing $k$. This branch is not triggered here: the smallest pool has $790$ observations (Fracture), and all $144$ cell entries are nonmissing. Both functions require matching row identities across modalities. Their label-free computation uses this row correspondence.

\paragraph{Budget descriptors.}
For a row at split tag $s$ with full training pool $n_{\mathrm{train}}$ and label set $\calY_d$:

\begin{equation}
\mathrm{budget\_frac}(d,K) \;=\; \frac{|\calS_{d,K}|}{n_{\mathrm{train}}}, \qquad
\mathrm{is\_kshot}(s) \;=\; \mathbf{1}[s = \texttt{k=}N] ,
\label{eq:feat-budget}
\end{equation}
together with $\log_{10} n_{\mathrm{train}}$, $\log_{10}\!\big(n_{\mathrm{train}} / |\calY_d|\big)$ and the number of labels per class $\log_{10}\!\big(|\calS_{d,K}| / |\calY_d|\big)$, where $|\calS_{d,K}|$ is the actual number of labelled examples of the cell (Table~\ref{tab:label-counts}): $20$ per class for a $20$-shot classification cell and $20$ per quantile bucket for a regression cell. For regression $|\calY_d|$ is the five quantile buckets, and for MCQ tasks, which have no fixed class set, the per-class counts are missing and imputed inside the fold. The full budget block includes all five descriptors. \mbox{The reduced block omits only the budget fraction.}

\subsubsection*{Permutation-invariant aggregators}
\label{app:features-aggregators}
The cascade aggregates the per-side values $f^A,f^B\in\mathbb{R}$ as

\begin{equation}
f_{\min} = \min(f^A, f^B), \qquad
f_{\max} = \max(f^A, f^B).
\end{equation}

With two sides, the mean and the absolute difference are linear in these two and are not kept. The modality-symmetric inventory is enumerated \mbox{in Table~\ref{tab:feature-inventory}.} The rows marked with a $\dagger$ (ANOVA, kNN-5, budget) need labels or budget metadata. The ANOVA and kNN-5 rows read only the labels of $\calS_{d,K}$. The remaining $12$ angle-A and angle-B aggregates are label-free geometry. Each step selects its feature subset within the outer fold using the procedure in Section~\ref{sec:hier-selection} and Appendix~\ref{app:cascade-config}.

\begin{table}[t]
\centering
\footnotesize
\setlength{\tabcolsep}{4pt}
\caption{\textbf{Feature inventory of $\phi(d,K)$ used by the cascade.} Each row lists a base function (with its defining equation) and the modality-symmetric aggregators retained in $\phi$. The text-specific $\Delta_{\mathrm{zs}}$ enters directly. ``Side'' is per modality before aggregation. ``PI'' is the resulting symmetric feature count. Rows marked $\dagger$ need labels or budget metadata, and the labels they read are those of $\calS_{d,K}$.}
\label{tab:feature-inventory}
\adjustbox{max width=\textwidth}{%
\begin{tabular}{l l c c}
\toprule
\textbf{Angle} & \textbf{Base function} & \textbf{Side count} & \textbf{PI aggregators (PI count)} \\
\midrule
A (per-side geometry) & $\mathrm{unif}$ (Eq.~\eqref{eq:feat-unif})           & A, B & $\min, \max$ \;(2) \\
                      & $\mathrm{srank}$ (Eq.~\eqref{eq:feat-srank})         & A, B & $\min, \max$ \;(2) \\
                      & $\mathrm{decay}$ (Eq.~\eqref{eq:feat-decay})         & A, B & $\min, \max$ \;(2) \\
                      & $\mathrm{cond}$ (Eq.~\eqref{eq:feat-cond})           & A, B & $\min, \max$ \;(2) \\
                      & $\mathrm{iso}$ (Eq.~\eqref{eq:feat-iso})             & A, B & $\min, \max$ \;(2) \\
                      & $f_{\mathrm F}$\;$\dagger$ (Eq.~\eqref{eq:feat-fisher}) & A, B & $\min, \max$ \;(2) \\
\midrule
B (cross-modal)       & $\mathrm{CKA}$ (Eq.~\eqref{eq:feat-cka})             & paired & symmetric scalar \;(1) \\
                      & $\mathrm{mknn}_{10}$ (Eq.~\eqref{eq:feat-mknn})      & paired & symmetric scalar \;(1) \\
\midrule
C (task signal $\dagger$) & $\mathrm{knn5}$ (Eq.~\eqref{eq:feat-knn5})       & A, B, $[A;B]$ & $\min, \max$ over $(A,B)$, plus combined \;(3) \\
\midrule
D (budget $\dagger$)  & $\mathrm{budget\_frac}, \mathrm{is\_kshot}$ (Eq.~\eqref{eq:feat-budget}) & --- & symmetric scalar each \;(2) \\
                      & $\log_{10} n_{\mathrm{train}}$, $\log_{10}\!\frac{n_{\mathrm{train}}}{|\calY_d|}$ & --- & symmetric scalar each \;(2) \\
                      & $\log_{10}\!\frac{|\calS_{d,K}|}{|\calY_d|}$  & --- & symmetric scalar \;(1) \\
C (text diagnostic $\dagger$) & $\Delta_{\mathrm{zs}}$ (Appendix~\ref{app:zs-feature}) & Text & direct scalar, fixed in Step 1 (1) \\
Task metadata & $r_d$, $q_d$, $\sigma_d$ & --- & binary indicators (3) \\
\bottomrule
\end{tabular}
}
\end{table}

\paragraph{Notes on implementation choices.}
\begin{itemize}[leftmargin=2em,itemsep=2pt]
  \item \textbf{Encoder pair behind $\phi$.} The embedding diagnostics use the default unaligned encoder pair of the dataset's domain, the same pair that \Decoupled\ and \Agent\ \mbox{are deployed with (Appendix~\ref{app:datasets}).} The text--label term of $\Delta_{\mathrm{zs}}$ additionally uses the fixed probe specified below.
  \item \textbf{Which labels the task-signal features read.} The $12$ geometry aggregates use no labels at all. The five label-dependent features, two $f_{\mathrm F}$ aggregates and three $\mathrm{knn5}$ values, are computed on the labelled set $\calS_{d,K}$ of the cell and on nothing else, so they change with the budget. Each budget has five labelled sets, one per subsampling seed. A held-out cell reads the features of one set, and the reported results average five runs, one per set. Training cells read the mean over the five sets. Singleton classes are omitted from the ANOVA calculation. The kNN score requires at least six examples. Undefined values are represented as missing and imputed inside the fold. The supervised representation diagnostics receive residuals $\tilde f=f-(a+b\log_{10}|\calS_{d,K}|)$, with separate regression and non-regression fits using only outer-training cells (Section~\ref{sec:hier-selection}).
  \item \textbf{Excluded cross-modal scalars.} The modality-gap centroid distance \citep{liang2022mind} requires a common coordinate space. Different tower widths preclude a direct subtraction. Equal width alone also does not establish a shared basis for independently trained encoders. The alignment loss $\mathbb{E}\|\hat z^A_i - \hat z^B_i\|_2^2$, cross-modal retrieval and the mean paired cosine also depend on a shared coordinate basis. We therefore use relational cross-modal features that accommodate different coordinate bases.
\end{itemize}

\subsection{Task-Type-Specific Text Diagnostic}
\label{app:zs-feature}
For $a_{\mathrm{TL}}$, all datasets use BAAI/bge-m3 as a fixed probe, taking its [CLS] vector with L2 normalization and a maximum input length of $512$ tokens. The checkpoint is pinned to revision \texttt{5617a9f61b028005a4858fdac845db406aefb181}. Inference uses \texttt{AutoModel} (XLMRobertaModel), the last-layer [CLS] representation, and CPU execution with Transformers 5.12.0 and PyTorch 2.6.0. The cached safetensors revision \texttt{9a0624b896d81da7492a910ffa53731274b6cf3d} (\texttt{refs/pr/130}) was also checked against the pinned checkpoint and the runtime model: all $391$ parameter tensors were elementwise identical, with maximum absolute difference zero. The probe is used solely for routing feature extraction. Texts are classified by cosine similarity against task class names, lowercased with underscores replaced by spaces. Food101 uses the official \texttt{classes.json}. OxfordPet uses the complete list of $37$ \mbox{breeds, including those absent from the example pool.}

The reported routing results average five runs with separately sampled text-probe sets. Each set contains up to $20$ examples from the permitted labelled subset $\calS_{d,K}$, with the available sample count limiting smaller budgets. Each run computes the text diagnostic from its sampled examples before routing. Class-balanced sampling takes examples round-robin from class buckets, using at most $20$ classes when $C>20$. This probe resampling is distinct from the per-seed selection of Agent in-context demonstrations from $\calS_{d,K}^{(s)}$. For $C\leq20$, represented classes are balanced as far as available examples permit. At $k=1$ each class contributes one example. For $C>20$, the pool contains one example from each of $20$ selected classes. CUB and OxfordPet use the same $20$-class subset across their cells and seeds. Food101's $50$ pools contain $26$ distinct class subsets whose union covers all $101$ classes. Thus, resampling for CUB and OxfordPet varies examples within fixed classes and permitted budget subsets. At $k=1$, each selected class has only one available example. Food101\textquotesingle s class coverage summarizes the collected pools. Classification always chooses among all $C$ class names, so chance normalization uses $1/C$ even when the probe covers fewer classes.

Empty texts are excluded: seven MetMeme entries are removed, leaving some pools with $5$ or $19$ examples. Fracture and OxfordPet lack dedicated pools at $20/40/60/80\%$. Their verified nested percentage splits permit reuse of a smaller-budget pool wholly contained in the current labelled set. This applies to $40$ cell--seed rows.

The text-neighbour term uses the same cosine leave-one-out $5$-NN function as the other task-signal features, on the cell's labelled set with a cap of $8{,}000$ rows. Text denotes captions or other sample text for image--text tasks, transcriptions for audio, descriptions for DrugBank, and SMILES for KIBA and PDBbind. Its evaluation set can therefore be larger than the text--label probe pool. For continuous regression targets, the diagnostic uses the task-specific definition $-r$. The zero assigned to QA \mbox{marks an inactive diagnostic, not equal system utility.}

The regression implementation, \texttt{embedding\_quality.\_knn\_regression}, returns NaN for $n<6$. Each regression $k=1$ cell has five observations, giving $15$ missing rows across three datasets and five seeds. Classification $5$-NN follows the same minimum-size rule. Its only affected cell is Memotion at $k=1$. Missing $\Delta_{\mathrm{zs}}$ values use the training fold's median. When either regression targets or leave-one-out predictions have standard deviation below $10^{-12}$, the implementation returns $r=0$, hence $\Delta_{\mathrm{zs}}=0$. No row took this constant-input branch. None of the $150$ regression rows had an exactly zero reported correlation. A held-out cell reads its single feature-seed value. Training cells read their five-seed mean. Unlike the ANOVA and kNN label diagnostics, $\Delta_{\mathrm{zs}}$ \mbox{receives no size residual adjustment.}

Step~1 always retains this diagnostic, appending it when it falls outside the selected Laplacian-ranked width. Step~2 does not use it. Table~\ref{tab:routing-feature-ablation} reports the corresponding feature-group ablations.

\subsection{Task Metadata and Missingness}
\label{app:metadata-missingness}
The indicators $r_d=\mathbf{1}[\mathrm{regression}]$, $q_d=\mathbf{1}[\mathrm{QA}]$, and $\sigma_d=\mathbf{1}[\mathrm{symbolic\ sequence\ modality}]$ are known before fitting candidate systems. The symbolic indicator distinguishes protein-sequence and SMILES inputs from image/audio inputs. The deployed text LLM receives non-text inputs through readouts. Step~1 fixes $r_d$ and $q_d$. Step~2 fixes $q_d$. The symbolic indicator is a ranked candidate in both steps. At $\epsilon=0$, it is selected in $75/75$ outer-fold--seed fits of Step~1 and $65/75$ fits of Step~2.

For a feature $x_j$ with missing values, append $m_j=\mathbf{1}[x_j\text{ is missing}]$ and impute $x_j$ with the training median before standardization. QA supervised representation diagnostics are unavailable under this feature convention. Leave-one-out $5$-NN requires at least six observations. One example per class does not by itself make it undefined when there are enough classes. In contrast, singleton classes do not supply within-class scatter for the ANOVA feature. The missingness indicators let the predictor distinguish unavailable \mbox{features from features observed at the median value.}

\FloatBarrier
\section{Agent Tool Schema and Prompt Templates}
\label{app:agent-prompt}

The \Agent\ combines unimodal readouts and available text through an LLM controller. Unless otherwise stated, it uses Qwen3-32B, Inline-1, and ten in-context examples. The ablations vary the controller, dispatch mode, or demonstration setting. It receives a task description, the sample text when one modality is text, $N_{\mathrm{ICL}}$ in-context examples (written ``K-shot examples'' in the prompt text below), and the outputs of the two readout tools. This appendix gives the complete tool schema (Section~\ref{app:tool-schema}), system prompt template (Section~\ref{app:system-prompt}), user prompt template (Section~\ref{app:user-prompt}), and pointers to worked examples (Section~\ref{app:worked-examples}).

\subsection{Tool Schema}
\label{app:tool-schema}

The agent's tool surface is $\mathcal{G}_d = \{g_a, g_b\}$: two single-modality readouts $g_a = \mathrm{readout}_a \circ E_a$ and $g_b = \mathrm{readout}_b \circ E_b$ computed from the frozen unimodal encoders that \Decoupled uses. Each readout returns class scores for classification, a scalar prediction for regression, option scores for MCQ, or answer-vocabulary scores for VQA-RAD. Trainable readouts are fitted separately on $\calS_{d,K}$ and selected using development utility. For image MCQ, let $v\in\mathbb R^{r_v}$ be the frozen \Decoupled\ image embedding and $o_j\in\mathbb R^{r_t}$ the option embedding from its text encoder. The image readout scores $s_j=o_j^\top Wv$, with $W\in\mathbb R^{r_t\times r_v}$ trained by cross-entropy over candidate answers. Development utility determines weight decay and training duration. The text readout instead computes unscaled cosine similarity between the current sample text embedding $t$ and each option embedding $o_j$, $s_j^{\mathrm{text}}=t^\top o_j/(\|t\|_2\|o_j\|_2)$, in the frozen \Decoupled\ text embedding space, without fitting a supervised head. Candidate answers condition the image scorer, but the image readout does not consume the sample's question text. For VQA-RAD, the readouts score a fixed answer vocabulary constructed from the training split. The LLM receives the resulting per-candidate scores alongside the available sample text.

\paragraph{Tool return format.} Each classification/QA tool returns:
\begin{itemize}[leftmargin=2em,itemsep=2pt]
    \item \textbf{prob}: softmax distribution over all classes (sums to $\approx 1$).
    \item \textbf{conf}: the sigmoid applied separately to each raw class logit. These scores are not normalised \mbox{across classes or calibrated as probabilities of correctness.}
    \item summary statistics: \mbox{top-1 softmax margin, maximum sigmoid, normalized entropy.}
\end{itemize}
For regression tasks, each tool returns a \mbox{single scalar prediction in the dataset's native unit.}

The sigmoid transform preserves within-sample logit ordering, but its values change under a common logit shift that leaves softmax probabilities unchanged. We use softmax margin and entropy to describe readout concentration. The experimental prompts also use the uncalibrated \texttt{conf} scores shown below. Their removal and sensitivity to logit shifts were not evaluated.

The OpenAI-function-call JSON schemas are:

\begin{schemabox}
{
  "name": "get_vision_prediction",
  "description": "Return per-class probability distribution from the
    vision / modality-A encoder's linear classifier for the current
    sample. No arguments.",
  "parameters": {"type": "object", "properties": {}, "required": []}
}

{
  "name": "get_text_prediction",
  "description": "Return per-class probability distribution from the
    text / modality-B encoder's linear classifier for the current
    sample. No arguments.",
  "parameters": {"type": "object", "properties": {}, "required": []}
}

\end{schemabox}

\subsection{System Prompt Template}
\label{app:system-prompt}

The system prompt is assembled at runtime from modular blocks. The core structure is:

\begin{promptbox}[System Prompt --- Core Structure]
You are a multimodal fusion agent for \placeholder{domain}-domain tasks.

Your job is to reason about the available evidence (tool predictions and your own analysis of the sample) and \mbox{output the correct answer for the current sample.}

\smallskip
\textbf{\#\# Inputs per sample}\\[2pt]
Each sample combines TEXT with another modality (image, audio clip, biological sequence, or chemical SMILES). You receive:
\begin{itemize}[leftmargin=1.5em,itemsep=1pt,topsep=2pt]
  \item \texttt{sample\_text}: \mbox{your direct view of the text modality.}
  \item The non-text modality is \mbox{NOT directly visible; access it only via tools.}
  \item K-shot examples (when \mbox{shown): labeled training samples to gauge modality contribution.}
  \item \placeholder{N} classifier-head tools (\mbox{relative quality is task-dependent).}
\end{itemize}

\smallskip
\textbf{\#\# Tools you may call}\\
\placeholder{tool listing generated per task}

\boxsep

\textbf{\#\# Reasoning protocol}

\smallskip
Before each tool call, write a brief analysis (1--2 sentences): \mbox{which tool \mbox{you intend to call and why.}}

\textbf{STEP 0 (Prior anchor):} If K-shot examples are provided, read them carefully. Commit YOUR OWN best guess using examples + \texttt{sample\_text} + the class list. \mbox{Write: \texttt{my\_prior = <class>}.}

\smallskip
Reading tool output:
\begin{itemize}[leftmargin=1.5em,itemsep=1pt,topsep=2pt]
  \item \textbf{prob} (softmax): relative distribution. \textbf{conf} (sigmoid): per-class \mbox{independent score. Read BOTH.}
  \item High sigmoid + large prob margin = peaked endorsement. Trust it.
  \item Flat distribution / low max sigmoid = \mbox{weak signal; rely on prior.}
  \item Multiple classes with high sigmoid = confident but ambiguous; \mbox{cross-check with a second tool.}
\end{itemize}

\boxsep

\textbf{\#\# Output format}\\
\texttt{ANSWER: <your final answer>}

ANSWER must be one concrete choice from the candidate set. Do NOT output ``Cannot determine'', ``Uncertain'', or similar.
\end{promptbox}

\paragraph{Task-specific reasoning hints.} For Memotion and Fracture the system prompt adds a ``\#\# Task-specific reasoning notes'' block. For meme sentiment classification (Memotion), the hint reads:

\begin{promptbox}[Task-specific hint --- Memotion]
Two reminders when judging meme sentiment:
\begin{enumerate}[leftmargin=1.8em,itemsep=2pt,topsep=3pt]
  \item \textbf{Joint evidence.} Sentiment is the emotional affect of the (image, caption) pair taken together. Neither channel alone \mbox{is the sentiment; read them as one communication.}
  \item \textbf{Intensity modifier.} \texttt{very\_} is an intensity modifier, not a confidence modifier. Pick \texttt{very\_positive} / \texttt{very\_negative} only when the joint emotional affect is strong; pick plain \texttt{positive} / \texttt{negative} when mild; pick \texttt{neutral} only \mbox{when the pair is genuinely flat or descriptive.}
\end{enumerate}
\end{promptbox}

\subsection{User Prompt Templates by Task Type}
\label{app:user-prompt}

\paragraph{Classification tasks.}

\begin{promptbox}[User Prompt --- Classification]
\textbf{\#\# Task}\\
\placeholder{task\_hint}

\smallskip
\textbf{\#\# Label meanings}
\begin{itemize}[leftmargin=1.5em,itemsep=1pt,topsep=2pt]
  \item \placeholder{class\_1}: \placeholder{description\_1}
  \item \placeholder{class\_2}: \placeholder{description\_2}
  \item \ldots
\end{itemize}

\smallskip
\textbf{\#\# Training examples (sample text $\to$ class)}\\[2pt]
These are a few labeled samples from this task's training set. Use them to gauge how predictive \texttt{sample\_text} is of the class.
\begin{itemize}[leftmargin=1.5em,itemsep=1pt,topsep=2pt]
  \item ``\placeholder{example\_text\_1}'' $\to$ \placeholder{class\_1}
  \item ``\placeholder{example\_text\_2}'' $\to$ \placeholder{class\_2}
  \item \ldots (10 stratified examples)
\end{itemize}

\smallskip
\textbf{\#\# Sample text}\\
``\placeholder{test\_sample\_text}''

\boxsep

\textbf{\#\# Procedure}
\begin{enumerate}[leftmargin=1.8em,itemsep=2pt,topsep=3pt]
  \item Commit \texttt{my\_prior = <class>} FIRST using K-shot + \texttt{sample\_text}.
  \item Call the \mbox{more informative tool; read BOTH prob and conf.}
  \item Reconcile \mbox{with your prior per the system reasoning protocol.}
  \item ANSWER must \mbox{be one of the listed class names verbatim.}
\end{enumerate}
\end{promptbox}

\paragraph{MCQ tasks (AI2D, ScienceQA, DrugBank).}

\begin{promptbox}[User Prompt --- MCQ]
\textbf{\#\# Task}\\
\placeholder{task\_hint}

\smallskip
\textbf{\#\# Question}\\
\placeholder{question\_text}

\smallskip
\textbf{\#\# Options}
\begin{itemize}[leftmargin=1.8em,itemsep=1pt,topsep=2pt,label={}]
  \item \textbf{A)} \placeholder{option\_A}
  \item \textbf{B)} \placeholder{option\_B}
  \item \textbf{C)} \placeholder{option\_C}
  \item \textbf{D)} \placeholder{option\_D}
\end{itemize}

\smallskip
\textbf{\#\# Sample text}\\
``\placeholder{context\_text}''

\boxsep

\textbf{\#\# Procedure}
\begin{enumerate}[leftmargin=1.8em,itemsep=2pt,topsep=3pt]
  \item Read the question and options. Note a tentative \mbox{answer from domain knowledge; do not commit yet.}
  \item Call a tool and read the per-option scores. A \mbox{top-1 with clear margin is a peaked endorsement.}
  \item Combine reasoning with tool evidence. If they agree, commit. \mbox{If they disagree, weigh which is more reliable.}
  \item Commit to exactly one option letter. \mbox{ANSWER is a single letter.}
\end{enumerate}
\end{promptbox}

\paragraph{VQA tasks (VQA-RAD).}

\begin{promptbox}[User Prompt --- VQA (Open-Ended)]
\textbf{\#\# Task}\\
\placeholder{task\_hint}

\smallskip
\textbf{\#\# Question}\\
\placeholder{question\_text}

\smallskip
\textbf{\#\# Sample text}\\
``\placeholder{clinical\_or\_scene\_text}''

\boxsep

\textbf{\#\# Procedure}
\begin{enumerate}[leftmargin=1.8em,itemsep=2pt,topsep=3pt]
  \item Form a tentative short \mbox{answer from domain knowledge; do not commit yet.}
  \item Call a tool. High sigmoid + clear margin = peaked endorsement. Flat distribution = \mbox{weak signal; rely on your own reasoning.}
  \item For questions answerable by language alone (yes/no, definitions), your knowledge tends to be more reliable. For questions requiring the medical image, \mbox{the tool's encoder evidence tends to be stronger.}
  \item End with \texttt{ANSWER: <short answer>}.
\end{enumerate}
\end{promptbox}

\paragraph{Regression tasks (KIBA, PDBbind, IEMOCAP).}
The template below shows the fields for a dataset with a text modality (IEMOCAP). For KIBA and PDBbind, where neither modality is text, the sample-text fields are omitted, so the sample \mbox{reaches the LLM only through the readout outputs.}

\begin{promptbox}[User Prompt --- Regression]
\textbf{\#\# Task}\\
\placeholder{task\_hint}

\smallskip
\textbf{\#\# Target}\\
\placeholder{continuous\_variable\_description}

\smallskip
\textbf{\#\# Training examples (sample text $\to$ target value)}
\begin{itemize}[leftmargin=1.5em,itemsep=1pt,topsep=2pt]
  \item ``\placeholder{example\_text\_1}'' $\to$ \placeholder{value\_1}
  \item ``\placeholder{example\_text\_2}'' $\to$ \placeholder{value\_2}
  \item \ldots
\end{itemize}

\smallskip
\textbf{\#\# Sample text}\\
``\placeholder{test\_sample\_text}''

\boxsep

\textbf{\#\# Procedure}
\begin{enumerate}[leftmargin=1.8em,itemsep=2pt,topsep=3pt]
  \item Inspect K-shot to see how \texttt{sample\_text} and tool scalars map to target values. \mbox{Form a rough internal estimate.}
  \item Call a tool and read its scalar output. A value within the range \mbox{implied by K-shot similar samples is strong evidence.}
  \item When tools diverge, do NOT average blindly; pick the tool whose scalar best fits the K-shot pattern. Fall back to a weighted \mbox{average only when both tools look equally plausible.}
  \item Commit to a single number. End with \texttt{ANSWER: <float>}.
\end{enumerate}
\end{promptbox}

\subsection{Worked Examples}
\label{app:worked-examples}

Appendix~\ref{app:case-studies} presents two complete trajectories in Inline-1 mode, with both readouts included in the user message. The CrisisMMD example contrasts a peaked readout with a flat one (Appendix~\ref{app:cs-crisismmd}). The Food101 example shows agreement between the sample text and vision readout (Appendix~\ref{app:cs-food101}).

\section{Case Studies: Agent Outputs and Rationales}
\label{app:case-studies}

We present two Inline-1 outputs alongside their inputs and generated rationales, highlighting how the reported decisions relate to the text and readouts.

\subsection{Case Study 1: CrisisMMD (Concentrated Vision and Diffuse Text Readouts)}
\label{app:cs-crisismmd}

CrisisMMD classifies \mbox{humanitarian content of disaster tweets into 8 categories.}

\begin{promptbox}[Case Study 1 --- CrisisMMD (Inline-1)]
\textbf{Task}: \mbox{Classify the humanitarian content of a disaster-related tweet.}

\smallskip
\textbf{Sample text}: ``Harvey forces National Weather Service to add new color to its rainfall map''

\smallskip
\textbf{In-context examples (10):}\\[2pt]
``Dramatic video shows school children rescued from rubble'' $\to$ missing\_or\_found\_people\\
``Puerto Rico (2001) My beautiful little enchanted island'' $\to$ not\_humanitarian\\
``A friend's text message saved Sarasota man from deadly wildfire'' $\to$ rescue\_volunteering\\
``RT @HuffPost: Puerto Rico's official death toll hits 39'' $\to$ injured\_or\_dead\_people\\
\ldots\ (6 more)

\boxsep

\textbf{Tool output --- get\_vision\_prediction} (DINOv2-ViT-L/14):
\smallskip

\begin{tabular}{@{}lrr@{}}
\toprule
Class & prob (softmax) & conf (sigmoid) \\
\midrule
other\_relevant\_information  & 0.7043 & 0.9281 \\
not\_humanitarian             & 0.2057 & 0.7904 \\
rescue\_volunteering\ldots    & 0.0376 & 0.4080 \\
\bottomrule
\end{tabular}

\smallskip
\textbf{Tool output --- get\_text\_prediction} (GPT-2):
\smallskip

\begin{tabular}{@{}lrr@{}}
\toprule
Class & prob (softmax) & conf (sigmoid) \\
\midrule
other\_relevant\_information  & 0.1872 & 0.5844 \\
not\_humanitarian             & 0.1861 & 0.5830 \\
rescue\_volunteering\ldots    & 0.1686 & 0.5588 \\
\bottomrule
\end{tabular}

\smallskip
Both blocks show the top three of eight classes. The tools also return the top-1 softmax margin, maximum sigmoid, and normalised entropy.

\boxsep

\rolelabel{Agent, Final}
The vision readout is strongly peaked at other\_relevant\_information (prob 0.704, conf 0.928). The text readout is nearly flat (top margin $\approx$ 0.001). The caption describes a technical map update --- relevant to Hurricane Harvey but not a direct \mbox{humanitarian event. I commit.}

\smallskip
\textbf{ANSWER: other\_relevant\_information}
\end{promptbox}

\textit{Analysis.} Both readouts rank the reported answer first, but with different margins: $0.499$ for vision and about $0.001$ for text. The generated rationale cites the concentrated vision prediction alongside the tweet's content.

\subsection{Case Study 2: Food101 Classification (Text Prior Agrees with Vision Readout)}
\label{app:cs-food101}

Food101 is a 101-class food classification task. In this example, the accompanying recipe-page text explicitly names the food.

\begin{promptbox}[Case Study 2 --- Food101 Classification (Inline-1)]
\textbf{Task}: Classify a food image into one of 101 categories. Inputs: \mbox{image + noisy web-scraped caption.}

\smallskip
\textbf{Sample text} (truncated): ``CakeJournal.com $\cdots$ \textbf{8 Easy Apple Pie Recipes} November 26, 2013 $\cdots$ Apple pie is one of my favorite fall desserts $\cdots$ These Apple Pie Bars are just oozing with everything I love about apple pie $\cdots$''

\smallskip
\textbf{In-context examples (10):}\\[2pt]
``$\cdots$ Deviled Egg Potato Salad $\cdots$'' $\to$ deviled\_eggs\\
``$\cdots$ Bacon Fried Rice $\cdots$'' $\to$ fried\_rice\\
``$\cdots$ Chorizo Breakfast Burrito $\cdots$'' $\to$ breakfast\_burrito\\
\ldots\ (7 more --- all show food name in page text $\to$ label)

\boxsep

\textbf{Tool output --- get\_vision\_prediction} (DINOv2-ViT-L/14):
\smallskip

\begin{tabular}{@{}lrr@{}}
\toprule
Class & prob (softmax) & conf (sigmoid) \\
\midrule
\textbf{apple\_pie}  & \textbf{0.8795} & \textbf{0.9989} \\
bread\_pudding       & 0.0306          & 0.9702          \\
carrot\_cake         & 0.0238          & 0.9619          \\
baklava              & 0.0193          & 0.9535          \\
\bottomrule
\end{tabular}

\smallskip
Top-1 prob margin = 0.849. Max sigmoid = 0.999.

\smallskip
\textbf{Tool output --- get\_text\_prediction} (GPT-2):
\smallskip

\begin{tabular}{@{}lrr@{}}
\toprule
Class & prob (softmax) & conf (sigmoid) \\
\midrule
chocolate\_mousse    & 0.1160 & 0.5926 \\
macaroni\_and\_cheese & 0.1158 & 0.5922 \\
cup\_cakes           & 0.1036 & 0.5652 \\
\ldots               &        &        \\
\bottomrule
\end{tabular}

\smallskip
Top-two softmax margin $\approx 0.0002$ (from the displayed probabilities). Max sigmoid = 0.593.

\boxsep

\rolelabel{Agent, Final}
The K-shot pattern shows food names appear in the page text. The current sample contains ``Apple Pie'' at least 15 times. The LLM prior strongly suggests \texttt{apple\_pie}. The vision readout confirms with an extreme signal: prob 0.880, sigmoid 0.999, margin 0.849. The text readout (GPT-2) is flat --- expected, as GPT-2 features do not parse recipe page semantics. Both \mbox{the LLM prior and the vision readout agree.}

\smallskip
\textbf{ANSWER: apple\_pie}
\end{promptbox}

\textit{Analysis.} The page text and vision readout both support \texttt{apple\_pie}, matching the final answer. The text readout instead ranks other classes highest and barely separates its top two predictions.

\FloatBarrier
\section{Computational Cost}
\label{app:cost}

Table~\ref{tab:cost} reports inference, preparation, and routing costs for the general-domain default configurations, with LLM-only as a reference. It distinguishes measurements from estimates.

\begingroup
\footnotesize
\setlength{\tabcolsep}{4pt}
\renewcommand{\arraystretch}{1.15}
\begin{longtable}{@{}p{0.19\textwidth}p{0.18\textwidth}p{0.18\textwidth}p{0.18\textwidth}p{0.18\textwidth}@{}}
\caption{\textbf{Computational costs and deployment requirements.} A: per-sample inference and deployment for the general-domain default configurations. B: dataset-level preparation and evaluation on one A100-80GB. C: one-time routing diagnostics per cell. [M] denotes measured values, [E] estimates, and --- an unreported measurement. The panels have different accounting units and must not be summed.}\label{tab:cost}\\
\toprule
\endfirsthead
\multicolumn{5}{l}{\textbf{Table \thetable\ continued: computational costs and deployment requirements}}\\
\toprule
\endhead
\midrule
\multicolumn{5}{r}{Continued on next page}\\
\endfoot
\bottomrule
\endlastfoot
\multicolumn{5}{l}{\textbf{A. System inference and deployment}}\\
\midrule
\textbf{Item} & \textbf{\Joint} & \textbf{\Decoupled} & \textbf{\Agent} & \textbf{LLM-only (reference)}\\
\midrule
Models run & CLIP-ViT-L/14 + head & DINOv2-L/14 + GPT-2 + head & Qwen3-32B + two readout pipelines & Qwen3-32B\\
Parameters & $0.43$B & $0.43$B & $\approx32.43$B & $32$B\\
LLM tokens/sample & Not applicable & Not applicable & $1.0$k--$8.5$k & $1.0$k assumption for estimate\\
FLOPs/sample [E] & $1.8\times10^{11}$ & $1.8\times10^{11}$ & $6.4\times10^{13}$ to $5.4\times10^{14}$ & $\approx6.4\times10^{13}$ at assumed lower token count\\
FLOPs relative to fusion [E] & $1\times$ & $1\times$ & $\sim360$--$3{,}000\times$ & $\sim360\times$ at assumed lower token count\\
Batched time/sample [M] & --- & --- & $0.60$ s ($0.27$--$0.95$) & ---\\
BF16 weight storage [E] & $\approx0.9$ GB & $\approx0.9$ GB & $\approx64.9$ GB & $\approx64$ GB\\
Additional runtime memory & Activations and runtime buffers & Activations and runtime buffers & KV cache, activations and runtime buffers & KV cache, activations and runtime buffers\\
Deployment options & Local GPU/CPU & Local GPU/CPU & Self-hosted LLM or external model service & Self-hosted LLM or external model service\\
LLM dependency & No & No & Yes & Yes\\
Task-specific fitting & Fusion head & Fusion head & Two unimodal readouts & None\\
\midrule
\multicolumn{5}{@{}p{0.97\textwidth}@{}}{\textit{Scope.} Approximate parameter counts use $32$B for the LLM and $0.43$B for the readout encoders and heads. BF16 weight storage is two bytes per parameter, excluding KV caches and runtime buffers. Encoder FLOPs use the $2PT$ approximation with $257$ image tokens and $77$ text tokens. Agent FLOPs add encoder computation to $2(32\times10^9)T$ for $T=1{,}000$--$8{,}500$ LLM tokens. Ratios use the fusion estimate before rounding. Agent timing is $0.60$ s per sample ($0.27$--$0.95$) under batched vLLM serving on A100-80GB GPUs. External model service is a deployment option, not a measured configuration.}\\
\midrule
\multicolumn{5}{l}{\textbf{B. Dataset-level preparation and evaluation}}\\
\midrule
\textbf{Dataset} & \textbf{Images / vision extraction} & \textbf{Fusion-head sweep} & \textbf{Cached-head full test} & \textbf{Test examples}\\
\midrule
CUB & $11{,}788$ / $\sim3.4$ min [M] & $7.5$ min [M] & $<1$ min & $1{,}767$\\
CrisisMMD & $18{,}082$ / $\sim5.3$ min [M] & $\sim3$ min [E] & $<1$ min & $2{,}237$\\
Food101 & $86{,}013$ / $\sim25$ min [M] & $\sim4$ min [E] & $<1$ min & $21{,}504$\\
Memotion & $6{,}992$ / $\sim2.0$ min [M] & $\sim2$ min [E] & $<1$ min & $700$\\
MetMeme & $3{,}992$ / $\sim1.15$ min [E] & $\sim2$ min [E] & $<1$ min & $800$\\
OxfordPet & $7{,}349$ / $\sim2.1$ min [M] & $\sim2$ min [E] & $<1$ min & $3{,}669$\\
\midrule
\multicolumn{5}{@{}p{0.97\textwidth}@{}}{\textit{Scope.} Vision extraction is reusable preprocessing. Each head-training time covers one paradigm's $8$-head $\times$ $5$-seed sweep for one cell. Full-test head evaluation uses cached embeddings and excludes encoder computation. MetMeme extraction time is scaled proportionally as $2.9\times3{,}992/10{,}045\approx1.15$ minutes. Agent inference timing is reported once in panel A.}\\
\multicolumn{5}{@{}p{0.97\textwidth}@{}}{\textit{Pretraining example.} CLIP ViT-L/14 was trained for $12$ days on $256$ V100 GPUs \citep{pmlr-v139-radford21a}, equivalent to $73{,}728$ V100 GPU-hours. This historical pretraining cost is separate from downstream adaptation and excludes the subsequent higher-resolution training epoch. It is an example rather than a total pretraining-cost comparison across systems.}\\
\midrule
\multicolumn{5}{l}{\textbf{C. One-time routing diagnostics per cell [E]}}\\
\midrule
\textbf{Component} & \multicolumn{2}{l}{\textbf{Input / scope}} & \multicolumn{2}{l}{\textbf{Estimated FLOPs}}\\
\midrule
BGE text probe & \multicolumn{2}{p{0.38\textwidth}}{At most $20$ examples plus $C$ class names, BGE-M3 ($0.57$B), up to $512$ tokens each} & \multicolumn{2}{p{0.38\textwidth}@{}}{Up to $1.3\times10^{14}$ for CUB ($C=200$)}\\
Representation geometry & \multicolumn{2}{p{0.38\textwidth}}{Cached budget-subset embeddings, with mutual-kNN using the training pool, at most $8{,}000$ rows} & \multicolumn{2}{p{0.38\textwidth}@{}}{Order of $10^{11}$}\\
Label diagnostics & \multicolumn{2}{p{0.38\textwidth}}{kNN and ANOVA-style statistics on embeddings of $\calS_{d,K}$} & \multicolumn{2}{p{0.38\textwidth}@{}}{About $10^{11}$ or less}\\
Logistic routing & \multicolumn{2}{p{0.38\textwidth}}{Approximately $20$ selected features} & \multicolumn{2}{p{0.38\textwidth}@{}}{Negligible relative to feature extraction}\\
\midrule
\multicolumn{5}{@{}p{0.97\textwidth}@{}}{\textit{Scope.} These analytical estimates exclude embedding extraction and candidate adaptation. The BGE calculation is $2(0.57\times10^9)(20+200)512\approx1.3\times10^{14}$ FLOPs. It is an upper estimate under $2PT$, not an implementation-independent bound. No end-to-end routing latency has been measured.}\\
\end{longtable}
\endgroup

The router selects for task utility, with ties favoring lower cross-modal training requirements. Deployment decisions additionally depend on inference cost and hardware availability.

\paragraph{Magnitude of AGENT's accuracy advantage.}
From the rounded main-table means, \Agent\ exceeds the better fusion configuration in $14$ of the $39$ classification/QA cells. The gain exceeds $1$, $2$, and $5$ absolute percentage points in $13$, $12$, and $8$ cells, respectively (Table~\ref{tab:agent-gain-thresholds}). \Agent\ wins none of the main-table regression settings. Appendix~\ref{app:perseed} reports seed variation. The two smallest positive accuracy gains are $0.002$ and $0.015$ on CrisisMMD at $20$-shot and full budget. Deployment decisions can weigh these margins against the costs in Table~\ref{tab:cost}.

\begin{table}[t]
\centering\footnotesize
\caption{\textbf{AGENT gain over the better fusion configuration.} Counts are computed from the rounded main-table mean accuracies on $39$ classification/QA cells. Thresholds are strict and expressed in absolute percentage points (pp). They are descriptive, not significance tests or universal deployment requirements. The $9$ \mbox{regression cells use MAE and are excluded.}}
\label{tab:agent-gain-thresholds}
\begin{tabular}{@{}lcc@{}}
\toprule
Accuracy gain threshold & Cells exceeding threshold & Share of $14$ positive-gain cells \\
\midrule
$>0$ pp & $14/39$ & $14/14$ \\
$>1$ pp & $13/39$ & $13/14$ \\
$>2$ pp & $12/39$ & $12/14$ \\
$>5$ pp & $8/39$ & $8/14$ \\
\bottomrule
\end{tabular}
\end{table}

\paragraph{Small relative gains over the full budget grid.}
Across all $144$ routing cells, $\Agent$ strictly exceeds the best FM in $46$ cells. Table~\ref{tab:agent-small-margins} reports cumulative small-margin counts using the same utility normalization as the tie tolerance, $\Delta/|U^\star|$. The four cells below $1\%$ are OxfordPet at $5$-shot, IEMOCAP at $5$-shot, and CrisisMMD at $20$-shot and $70\%$. In these cells, choosing the best FM sacrifices less than $1\%$ of the best utility magnitude and can be attractive when inference cost matters. The choice depends on the application's utility--cost trade-off. Unlike Table~\ref{tab:agent-gain-thresholds}, this analysis covers the full budget grid and includes regression through normalized utility.

\begin{table}[t]
\centering
\footnotesize
\caption{\textbf{Small AGENT advantages across the full routing grid.} Of $144$ cells, $46$ have strictly positive $\Delta=U_A-\max(U_J,U_D)$ at $\epsilon=0$. The relative margin is $\Delta/|U^\star|$, using the same denominator as the tie tolerance. Here $U^\star=U_A$. Counts are cumulative within these $46$ cells. Accuracy and negative MAE enter through utility, not pooled raw metric differences.}
\label{tab:agent-small-margins}
\begin{tabular}{@{}lc@{}}
\toprule
\textbf{Relative advantage over the best FM} & \textbf{Number of cells} \\
\midrule
$<0.5\%$ & $3$ \\
$<1\%$ & $4$ \\
$<2\%$ & $5$ \\
$<5\%$ & $8$ \\
\bottomrule
\end{tabular}
\end{table}

\paragraph{Cost of selection before adaptation.}
At a new task, the router uses frozen embeddings, training-data diagnostics, the BGE text probe, and fitted logistic decision scores. Embeddings and class-name encodings can be reused when their inputs are unchanged. Label-dependent diagnostics follow the corresponding labelled set. Table~\ref{tab:cost}, panel B, reports vision extraction, while panel C estimates diagnostic computation after embeddings are available. These costs precede adaptation and inference of the selected system. End-to-end routing speedup has not been measured.

\section{Reproducibility Statement}
\label{app:reproducibility}

We will release, upon publication, the feature-extraction code with the exact checkpoint identifiers and preprocessing for every encoder, the head-training code, the complete \Agent\ implementation, the per-dataset split indices with the subsampling seed of each label budget, and scripts that regenerate the benchmark tables and \mbox{figures from the released caches and head weights.}

The planned cascade release includes the dataset-budget utilities and feature values for all labelled-set seeds, together with the fold-specific preprocessing, feature choices, thresholds, predicted configurations, logistic scores, and regrets. The planned head-selection manifest links each reported utility to its development-selected seed model. Table~\ref{tab:router-per-dataset} reports per-dataset aggregate accuracy and regret. Appendix~\ref{app:reporting-scope} describes the scope of the matched controls and routing evidence.

Randomness enters through labelled-set sampling and model initialization. Label-budget subsampling uses seeds $\{0,\dots,4\}$, five per cell, and head initialisation takes the subsampling seed. \Agent\ decodes at temperature $0$, and its reported utility is averaged over seeds $\{0,\dots,4\}$, corresponding to the five \Decoupled\ seeds. For each seed $s$, the in-context examples of \Agent\ are drawn stratified by class from that seed's labelled set $\calS_{d,K}^{(s)}$, so they vary with the seed and never use labels outside the budget. For each held-out cell, the cascade is deterministic given its labelled-set features. Unparseable classification and QA outputs count as incorrect. For regression, the implementation substitutes a prediction of $0$ before computing MAE (Table~\ref{tab:agent_inference}). Table~\ref{tab:regression-parse} gives failure percentages at the main-table budgets. The resulting MAE combines numerical prediction error with parsing failures. The release will include exact failure counts and seed-resolved records.

Per-sample predictions, trained head weights, and the full feature caches ($\sim$200\,GB) will be hosted alongside the code.

\section{Scope of the Reported Controls and Routing Evidence}
\label{app:reporting-scope}

\paragraph{System comparison and causal attribution.}
The benchmark compares three pretrained configurations on paired target tasks under shared splits, training-label budgets, metrics, and development-based fusion-head selection. Architecture and scale matching constrain some differences, while pretraining data, objectives, optimization, and representation quality vary with the candidate systems. Reported utilities and routing results therefore characterize these complete configurations.

\paragraph{Question access in VQA-RAD.}
All three paradigms receive the VQA-RAD question. Fusion encodes it together with each of the $500$ candidate answers and combines that text embedding with the image embedding. The \Agent\ controller receives the question as direct text. Question-removal controls change the controller input while retaining its readout outputs.

\paragraph{Three-paradigm routing and component controls.}
The routing objective is to predict the preferred member of $\mathcal P=\{\Joint,\Decoupled,\Agent\}$, the three assembly strategies instantiated in Section~\ref{sec:three-alignments}. Winner accuracy and relative regret use this same candidate set throughout. Component controls address a complementary question: what utility is added by the readouts, direct text access, or a second modality? On AI2D at full budget, removing the readouts improves accuracy from $0.778$ to $0.830$. On CUB, the visual probe nearly matches full-budget \Decoupled\ ($0.916$ versus $0.917$). On Memotion at $20$-shot, an alternative BERT-only probe reaches $0.426$, above the best primary configuration at $0.420$. Conversely, Food101's LLM-only, listed single-modality, and averaging controls remain below \Agent\ at all three reported budgets. These findings qualify the contribution of individual components while the routing analysis measures predictability of the three-paradigm preference. Regret uses the best member of $\mathcal P$ as its reference.

\paragraph{Training-label budgets with a fixed development resource.}
The disjoint development set $\calV_d$ remains fixed as the training set $\calS_{d,K}$ varies, providing common conditions for selecting fusion heads, hyperparameters, and checkpoints. Development and test examples remain outside the training pool. Training budgets count $\lvert\calS_{d,K}\rvert$ labels, with development labels reported separately. This design measures training-budget sensitivity, while applications can allocate their total labels between training and development.

\paragraph{What the AGENT controls establish.}
All conditions in Table~\ref{tab:agent-controls} use each dataset's complete test split and the same five-seed aggregation. The Deployed reference is the corresponding main-table full-budget result. The table caption specifies the retained and removed inputs for each intervention, and Appendix~\ref{app:agent-controls} analyzes their effects. LLM-only tests the value of readouts with raw-text access retained, while shuffled readouts test sample--readout correspondence. The question-removal and LLM-plus-readouts conditions distinguish removing the question body from removing direct text content. Each contrast measures an input change conditional on the other inputs retained by that control.

\paragraph{Coverage and uncertainty of the input evidence.}
Table~\ref{tab:agent-controls} covers eight datasets at full budget, supplemented by Food101 readout-removal controls at three budgets in Table~\ref{tab:agent-score-avg}. These controls report five-seed mean differences on fixed test examples. Seed variation describes fitted-system variability, while routing group-bootstrap intervals describe aggregate routing differences. Paired test-example intervals are not available for the input controls.

\paragraph{Paired alignment and target-task setting.}
The alignment and paradigm-selection problem studied here assumes paired multimodal target data. \Joint\ also uses a checkpoint aligned on paired data during pretraining. Pretraining pairs and target-task pairs serve different stages of the pipeline and need not be the same observations. The trainable \Agent\ readouts are fitted separately within this paired target-task setting. The image-MCQ text readout instead uses frozen embedding similarity. Per-side geometry and labelled readout diagnostics can be computed independently, whereas CKA, mutual-kNN overlap, and concatenated kNN use target-sample correspondence. Thus the complete cascade, including Step~1, operates in the paired setting defined in Section~\ref{sec:three-alignments}.

\paragraph{Grouped routing evidence.}
The $144$ cells are evaluated in $15$ groups, with both DrugBank tasks grouped throughout outer evaluation, inner classifier fitting, threshold estimation, and bootstrap resampling. An outer test group is excluded from all fitted transformations. Inner validation excludes its group from classifier fitting and fit-specific preprocessing, while retaining residualization and feature ranking fitted on the outer-training data. Group-bootstrap intervals describe cell-weighted differences conditional on the candidate utilities and labels. Table~\ref{tab:router-per-dataset} reports task-level variation.

\paragraph{Utility and deployment cost.}
Task-utility selection excludes deployment cost. Appendix~\ref{app:cost} reports configuration-specific FLOPs estimates and measured timing separately. A cost-aware analysis must state its compute boundary, token lengths, hardware, and units. Pareto domination within a cell requires no lower utility and no greater cost, with one strict improvement. A scalar extension $U_\lambda=U-\lambda C$ also needs a stated trade-off scale. Accuracy points cannot be pooled with MAE differences. Such cost-aware objectives extend the utility-based selection evaluated here.

\end{document}